\documentclass[10pt]{article} %

\usepackage[preprint]{rlj} %

\usepackage{amssymb}            %
\usepackage{mathtools}          %
\usepackage{mathrsfs}           %
\usepackage{graphicx}           %
\usepackage{subcaption}         %
\usepackage[space]{grffile}     %
\usepackage{url}                %

\newcommand{\mdp}{\mathcal{M}}
\newcommand{\statesp}{\mathcal{S}}
\newcommand{\actsp}{\mathcal{A}}
\newcommand{\rewsp}{r}
\newcommand{\dyn}{p}
\newcommand{\initdist}{\mathbb{I}}
\newcommand{\discount}{\gamma}

\newcommand{\pol}{\pi}

\newcommand{\stateset}{\mathbb{S}}
\newcommand{\actset}{\mathbb{A}}
\newcommand{\accset}{\mathbb{C}}
\newcommand{\termset}{\mathbb{T}}
\newcommand{\aleaset}{\mathbb{W}}
\newcommand{\dimS}{n_{\mathcal{S}}}
\newcommand{\rvstate}{S}
\newcommand{\state}{s}

\newcommand{\rvmodstate}{\hat{S}}

\newcommand{\rvact}{A}
\newcommand{\act}{a}
\newcommand{\rvrew}{R}
\newcommand{\rew}{r}
\newcommand{\rewfcn}{r}
\newcommand{\rvalea}{W}
\newcommand{\alea}{w}
\newcommand{\rvmodalea}{\hat{W}}

\newcommand{\params}{\theta}

\newcommand{\predmean}{z}
\newcommand{\predact}{u}
\newcommand{\predvar}{P}
\newcommand{\prepredvar}{Q}
\newcommand{\predchol}{M}
\newcommand{\unitvec}{q}
\newcommand{\unitball}{\mathbb{Q}}
\newcommand{\sysmat}{A}
\newcommand{\inmat}{B}
\newcommand{\cloopmat}{F}
\newcommand{\contgain}{K}
\newcommand{\kalgain}{G}

\newcommand{\prs}{\mathcal{R}}
\newcommand{\errball}{\mathcal{B}}

\newcommand{\ellscale}{\epsilon}
\newcommand{\linerr}{\mathcal{E}}

\newcommand{\globid}{t}
\newcommand{\predid}{n}
\newcommand{\predhorizon}{N}
\newcommand{\epiid}{j}
\newcommand{\modid}{e}
\newcommand{\nummods}{E}

\newcommand{\meanenv}{\mu}
\newcommand{\varenv}{\Sigma}
\newcommand{\cholenv}{L}

\newcommand{\meanmod}{\mu}
\newcommand{\varmod}{\Sigma}
\newcommand{\cholmod}{L}

\newcommand{\mlmeanmod}{\bar{\mu}}
\newcommand{\mlvarmod}{\bar{\Sigma}}

\newcommand{\epivarmod}{\hat{\Sigma}}

\newcommand{\initpol}{\kappa}
\newcommand{\buffer}{\mathcal{D}}

\newcommand{\gauss}{\mathcal{N}}

\newcommand{\E}{\mathbb{E}}

\newcommand{\lipschitz}{\ell_{\nabla \mu}}
\newcommand{\lipschitzmod}{\ell_G}

\newcommand{\lipschitzvarchol}{\ell_L}
\newcommand{\hess}{\mathcal{H}}
\newcommand{\grad}{\nabla}
\newcommand{\diff}{\delta}

\newcommand{\tube}{\mathcal{P}}
\newcommand{\contrset}{\mathcal{K}}
\newcommand{\pre}{\mathrm{Pre}}

\newcommand{\algo}{UPSi\xspace}

\usepackage{amsthm}
\usepackage{xspace}

\theoremstyle{plain}
\newtheorem{theorem}{Theorem}
\newtheorem{lemma}[theorem]{Lemma}

\theoremstyle{definition}
\newtheorem{definition}{Definition}
\newtheorem{assumption}{Assumption}

\usepackage{ifthen}
\usepackage{soul}

\usepackage{algorithm}
\usepackage{algpseudocode}

\title{Uncertainty-Aware Predictive Safety Filters for \\ Probabilistic Neural Network Dynamics}

\setrunningtitle{UPSi: Uncertainty-Aware Predictive Safety Filters for Probabilistic Neural Network Dynamics}

\author{Bernd Frauenknecht\textsuperscript{1,$\star$}, Lukas Kesper\textsuperscript{1,2,$\star$}, Daniel Mayfrank\textsuperscript{2}, Henrik Hose\textsuperscript{1}, Sebastian Trimpe\textsuperscript{1}}

\emails{\{bernd.frauenknecht; lukas.kesper; trimpe\}@dsme.rwth-aachen.de}

\affiliations{
$^{1}$\textbf{Institute for Data Science in Mechanical Engineering (DSME), RWTH Aachen University}\\
$^{2}$\textbf{Institute of Climate and Energy Systems (ICE), Energy Systems Engineering (ICE-1), Forschungszentrum Jülich GmbH}\\
\par %
$^\star$ Equal contribution
}

\contribution{
    We present a rigorous mechanism for robust reachable set computation using probabilistic ensemble models, building on recent results in uncertainty quantification for these models.
    }
    {
    Prior rigorous approaches to robust reachable set computation rely on first-principles models or learning-based models that scale poorly with data, such as Gaussian processes. Further, prior approaches to reachable set computation with neural network models do not necessarily provide an overapproximation of environment outcomes and are therefore not rigorous.
    }
\contribution{ Based on these reachable sets, we introduce an explicit certainty constraint in the predictive safety filter formulation to avoid model exploitation.
    }
    {
    Despite their desirable representational capacity, neural network dynamics models are prone to model exploitation, i.e., out-of-distribution predictions become arbitrarily poor. To leverage the probabilistic ensemble model for safety filtering, we need to enforce in-distribution evaluation.
    }
\contribution{
We empirically validate the reachable set overapproximation and evaluate our filter mechanism integrated into MBPO with practical simplifications on common safe exploration benchmarks. We report substantial improvements in exploration safety compared to a prior neural network-based PSF while yielding comparable performance to standard MBPO.
}
{
None.
}


\summary{Predictive safety filters (PSFs) leverage model predictive control to enforce constraint satisfaction during deep reinforcement learning (RL) exploration, yet their reliance on first-principle models or Gaussian processes limits scalability and broader applicability. Meanwhile, model-based RL (MBRL) methods routinely employ probabilistic ensemble (PE) neural networks to capture complex, high-dimensional dynamics from data with minimal prior knowledge. However, existing attempts to integrate PEs into PSFs lack rigorous uncertainty quantification. We introduce the \emph{Uncertainty-Aware Predictive Safety Filter}~(\algo), a PSF that provides rigorous safety verification using PE dynamics models by formulating future outcomes as reachable sets. \algo~introduces an explicit certainty constraint that prevents model exploitation and integrates seamlessly into common MBRL frameworks. We evaluate \algo under practical simplifications within Dyna-style MBRL on standard safe RL benchmarks and report substantial improvements in exploration safety over prior neural network PSFs while maintaining performance on par with standard MBRL. \algo~bridges the gap between the scalability and generality of modern MBRL and the safety guarantees of predictive safety filters.}

\begin{document}

\maketitle  %
\begin{center}
{\itshape Reinforcement Learning Journal, 2026. \\ Reinforcement Learning Conference (RLC), Montr\'eal, August 2026.}
\vspace{2em}
\end{center}
\begin{abstract}
Predictive safety filters (PSFs) leverage model predictive control to enforce constraint satisfaction during deep reinforcement learning (RL) exploration, yet their reliance on first-principles models or Gaussian processes limits scalability and broader applicability. Meanwhile, model-based RL (MBRL) methods routinely employ probabilistic ensemble (PE) neural networks to capture complex, high-dimensional dynamics from data with minimal prior knowledge. However, existing attempts to integrate PEs into PSFs lack rigorous uncertainty quantification. We introduce the \emph{Uncertainty-Aware Predictive Safety Filter}~(\algo), a PSF that provides rigorous safety verification using PE dynamics models by formulating future outcomes as reachable sets. \algo~introduces an explicit certainty constraint that prevents model exploitation and integrates seamlessly into common MBRL frameworks. We evaluate \algo under practical simplifications within Dyna-style MBRL on standard safe RL benchmarks and report substantial improvements in exploration safety over prior neural network PSFs while maintaining performance on par with standard MBRL. \algo~bridges the gap between the scalability and generality of modern MBRL and the safety guarantees of predictive safety filters.

\end{abstract}

\section{Introduction}

Exploration safety is a major concern in deep reinforcement learning (RL). Predictive safety filters (PSFs) \citep{wabersich_linear_2018} use concepts from model predictive control (MPC) \citep{rawlings_model_2024} to provide rigorous safety guarantees. During environment interaction, a model of the environment predicts all possible outcomes of the current action over a finite horizon. The PSF aims to find an action sequence such that constraints are satisfied and adapts the RL action if necessary.
Despite real-world successes~\citep{bejarano_safety_2025}, PSFs typically rely on model classes common in MPC literature to predict future outcomes, such as first-principles engineering models~\citep{bejarano_safety_2025} or Gaussian processes (GPs)~\citep{koller_learning-based_2018}, that either demand substantial prior knowledge or scale poorly with data, limiting the applicability of PSFs.
In contrast, state-of-the-art model-based reinforcement learning (MBRL) approaches~\citep{janner_when_2019-1, frauenknecht_trust_2024, frauenknecht_rollouts_2025} use Bayesian
neural network architectures, such as probabilistic ensembles~(PEs)~\citep{lakshminarayanan_simple_2017}, to model high-dimensional, nonlinear dynamics from large datasets with minimal prior knowledge. While PEs hold the potential to extend PSF methods to a substantially broader class of systems,
existing approaches that incorporate PEs into PSFs~\citep{gronauer_reinforcement_2024,
guzelkaya_ensemble_2024} lack rigorous uncertainty handling and therefore remain heuristic.

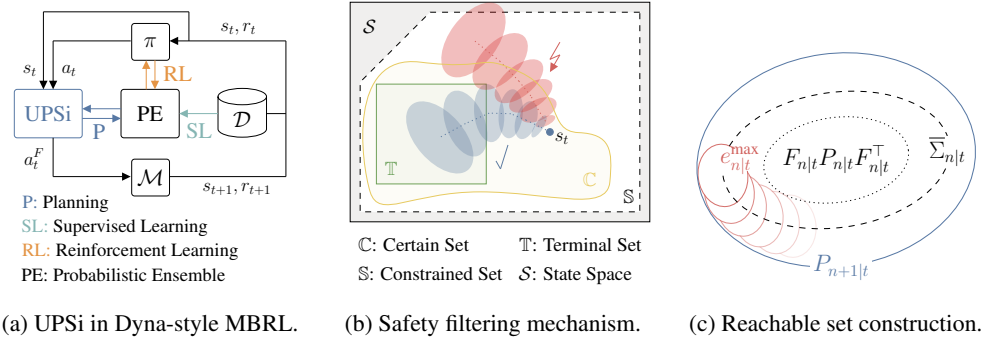
\begin{figure}[htbp]
    \centering
    \vspace{-0.5em}
    \begin{subfigure}[b]{0.32\textwidth}
        \centering
        \tikzset{every picture/.style={line width=0.75pt}} %

\resizebox{\linewidth}{!}{

\begin{tikzpicture}[x=0.75pt,y=0.75pt,yscale=-1,xscale=1]
\clip (0,0) rectangle (340, 288); %

\draw   (151,162) .. controls (151,160.34) and (152.34,159) .. (154,159) -- (187,159) .. controls (188.66,159) and (190,160.34) .. (190,162) -- (190,194) .. controls (190,195.66) and (188.66,197) .. (187,197) -- (154,197) .. controls (152.34,197) and (151,195.66) .. (151,194) -- cycle ;
\draw   (151,22) .. controls (151,20.34) and (152.34,19) .. (154,19) -- (187,19) .. controls (188.66,19) and (190,20.34) .. (190,22) -- (190,54) .. controls (190,55.66) and (188.66,57) .. (187,57) -- (154,57) .. controls (152.34,57) and (151,55.66) .. (151,54) -- cycle ;
\draw  [color={rgb, 255:red, 87; green, 120; blue, 164 }  ,draw opacity=1 ] (30,90) .. controls (30,88.34) and (31.34,87) .. (33,87) -- (97,87) .. controls (98.66,87) and (100,88.34) .. (100,90) -- (100,134) .. controls (100,135.66) and (98.66,137) .. (97,137) -- (33,137) .. controls (31.34,137) and (30,135.66) .. (30,134) -- cycle ;
\draw  [color={rgb, 255:red, 255; green, 255; blue, 255 }  ,draw opacity=1 ] (20,1) -- (320,1) -- (320,291) -- (20,291) -- cycle ;
\draw   (290,94.44) -- (290,127.91) .. controls (290,132.57) and (278.81,136.34) .. (265,136.34) .. controls (251.19,136.34) and (240,132.57) .. (240,127.91) -- (240,94.44) .. controls (240,89.78) and (251.19,86) .. (265,86) .. controls (278.81,86) and (290,89.78) .. (290,94.44) .. controls (290,99.1) and (278.81,102.88) .. (265,102.88) .. controls (251.19,102.88) and (240,99.1) .. (240,94.44) ;
\draw   (140,90) .. controls (140,88.34) and (141.34,87) .. (143,87) -- (197,87) .. controls (198.66,87) and (200,88.34) .. (200,90) -- (200,134) .. controls (200,135.66) and (198.66,137) .. (197,137) -- (143,137) .. controls (141.34,137) and (140,135.66) .. (140,134) -- cycle ;
\draw    (190,177) -- (310,177) ;
\draw    (70,177) -- (147,177) ;
\draw [shift={(150,177)}, rotate = 180] [fill={rgb, 255:red, 0; green, 0; blue, 0 }  ][line width=0.08]  [draw opacity=0] (8.93,-4.29) -- (0,0) -- (8.93,4.29) -- cycle    ;
\draw    (70,177) -- (70,137) ;
\draw    (70,84) -- (70,37) ;
\draw [shift={(70,87)}, rotate = 270] [fill={rgb, 255:red, 0; green, 0; blue, 0 }  ][line width=0.08]  [draw opacity=0] (8.93,-4.29) -- (0,0) -- (8.93,4.29) -- cycle    ;
\draw    (70,37) -- (150,37) ;
\draw    (60,7) -- (210,7) ;
\draw    (210,37) -- (210,7) ;
\draw    (60,7) -- (60,84) ;
\draw [shift={(60,87)}, rotate = 270] [fill={rgb, 255:red, 0; green, 0; blue, 0 }  ][line width=0.08]  [draw opacity=0] (8.93,-4.29) -- (0,0) -- (8.93,4.29) -- cycle    ;
\draw    (193,37) -- (310,37) ;
\draw [shift={(190,37)}, rotate = 0] [fill={rgb, 255:red, 0; green, 0; blue, 0 }  ][line width=0.08]  [draw opacity=0] (8.93,-4.29) -- (0,0) -- (8.93,4.29) -- cycle    ;
\draw    (310,37) -- (310,177) ;
\draw [color={rgb, 255:red, 0; green, 0; blue, 0 }  ,draw opacity=1 ]   (289.5,112) -- (310,112) ;
\draw [color={rgb, 255:red, 228; green, 148; blue, 68 }  ,draw opacity=1 ]   (175,57) -- (175,84) ;
\draw [shift={(175,87)}, rotate = 270] [fill={rgb, 255:red, 228; green, 148; blue, 68 }  ,fill opacity=1 ][line width=0.08]  [draw opacity=0] (8.93,-4.29) -- (0,0) -- (8.93,4.29) -- cycle    ;
\draw [color={rgb, 255:red, 228; green, 148; blue, 68 }  ,draw opacity=1 ]   (166,60) -- (166,87) ;
\draw [shift={(166,57)}, rotate = 90] [fill={rgb, 255:red, 228; green, 148; blue, 68 }  ,fill opacity=1 ][line width=0.08]  [draw opacity=0] (8.93,-4.29) -- (0,0) -- (8.93,4.29) -- cycle    ;
\draw [color={rgb, 255:red, 133; green, 182; blue, 178 }  ,draw opacity=1 ]   (240,112) -- (203,112) ;
\draw [shift={(200,112)}, rotate = 360] [fill={rgb, 255:red, 133; green, 182; blue, 178 }  ,fill opacity=1 ][line width=0.08]  [draw opacity=0] (8.93,-4.29) -- (0,0) -- (8.93,4.29) -- cycle    ;
\draw [color={rgb, 255:red, 87; green, 120; blue, 164 }  ,draw opacity=1 ]   (100,117) -- (137,117) ;
\draw [shift={(140,117)}, rotate = 180] [fill={rgb, 255:red, 87; green, 120; blue, 164 }  ,fill opacity=1 ][line width=0.08]  [draw opacity=0] (8.93,-4.29) -- (0,0) -- (8.93,4.29) -- cycle    ;
\draw [color={rgb, 255:red, 87; green, 120; blue, 164 }  ,draw opacity=1 ]   (103,107) -- (140,107) ;
\draw [shift={(100,107)}, rotate = 0] [fill={rgb, 255:red, 87; green, 120; blue, 164 }  ,fill opacity=1 ][line width=0.08]  [draw opacity=0] (8.93,-4.29) -- (0,0) -- (8.93,4.29) -- cycle    ;

\draw (261,24.5) node  [font=\Large] [align=left] {$\displaystyle s_{t} ,r_{t}$};
\draw (259.5,189.5) node  [font=\Large] [align=left] {$\displaystyle s_{t+1} ,r_{t+1}$};
\draw (220.5,129) node  [font=\LARGE,color={rgb, 255:red, 133; green, 182; blue, 178 }  ,opacity=1 ] [align=left] {\textcolor[rgb]{0.52,0.71,0.7}{SL}};
\draw (65,112) node  [font=\LARGE,color={rgb, 255:red, 87; green, 120; blue, 164 }  ,opacity=1 ] [align=left] {\textcolor[rgb]{0.34,0.47,0.64}{UPSi}};
\draw (170.5,178) node  [font=\LARGE] [align=left] {$\displaystyle \mathcal{M}$};
\draw (170.5,38) node  [font=\LARGE] [align=left] {$\displaystyle \pi $};
\draw (170,112) node  [font=\LARGE] [align=left] {PE};
\draw (265,118) node  [font=\LARGE] [align=left] {$\displaystyle \mathcal{D}$};
\draw (149.5,255) node  [font=\Large] [align=left] {\textcolor[rgb]{0.89,0.58,0.27}{RL:} Reinforcement Learning};
\draw (83,205) node  [font=\Large] [align=left] {\textcolor[rgb]{0.34,0.47,0.64}{P:} Planning};
\draw (117.5,131) node  [font=\LARGE,color={rgb, 255:red, 87; green, 120; blue, 164 }  ,opacity=1 ] [align=left] {P};
\draw (198.5,71) node  [font=\LARGE,color={rgb, 255:red, 228; green, 148; blue, 68 }  ,opacity=1 ] [align=left] {RL};
\draw (46,67.5) node  [font=\Large] [align=left] {$\displaystyle s_{t}$};
\draw (86.5,67.5) node  [font=\Large] [align=left] {$\displaystyle a_{t}$};
\draw (50.5,158.5) node  [font=\Large] [align=left] {$\displaystyle a_{t}^{F}$};
\draw (133.5,230) node  [font=\Large] [align=left] {\textcolor[rgb]{0.52,0.71,0.7}{SL:} Supervised Learning};
\draw (142.5,278) node  [font=\Large] [align=left] {PE: Probabilistic Ensemble};

\end{tikzpicture}

}
        \vspace{-1em}
        \caption{\algo in Dyna-style MBRL.}
        \label{fig:upsi_block}
    \end{subfigure}
    \begin{subfigure}[b]{0.32\textwidth}
        \centering
        \tikzset{every picture/.style={line width=0.75pt}} %

\resizebox{\linewidth}{!}{

\begin{tikzpicture}[x=0.75pt,y=0.75pt,yscale=-1,xscale=1]
\clip (0,0) rectangle (330, 293); %

\draw  [fill={rgb, 255:red, 184; green, 176; blue, 172 }  ,fill opacity=0.2 ] (22,10.5) -- (322,10.5) -- (322,230.5) -- (22,230.5) -- cycle ;
\draw  [fill={rgb, 255:red, 255; green, 255; blue, 255 }  ,fill opacity=1 ][dash pattern={on 4.5pt off 4.5pt}] (312,20.5) -- (102,20.5) -- (32,90.5) -- (32,220.5) -- (312,220.5) -- cycle ;
\draw  [color={rgb, 255:red, 255; green, 255; blue, 255 }  ,draw opacity=1 ] (20,8) -- (324,8) -- (324,300.5) -- (20,300.5) -- cycle ;
\draw  [color={rgb, 255:red, 106; green, 159; blue, 88 }  ,draw opacity=1 ][fill={rgb, 255:red, 106; green, 159; blue, 88 }  ,fill opacity=0.05 ] (48.26,92.54) -- (158.26,92.54) -- (158.26,192.54) -- (48.26,192.54) -- cycle ;
\draw  [draw opacity=0][fill={rgb, 255:red, 87; green, 120; blue, 164 }  ,fill opacity=0.3 ] (83.3,113.49) .. controls (93.32,104.73) and (114.16,112.15) .. (129.85,130.08) .. controls (145.54,148.01) and (150.14,169.65) .. (140.12,178.42) .. controls (130.1,187.18) and (109.26,179.76) .. (93.57,161.83) .. controls (77.88,143.9) and (73.28,122.26) .. (83.3,113.49) -- cycle ;
\draw  [draw opacity=0][fill={rgb, 255:red, 87; green, 120; blue, 164 }  ,fill opacity=0.3 ] (124.69,99.82) .. controls (134.69,95.89) and (148.47,107.15) .. (155.48,124.96) .. controls (162.49,142.78) and (160.07,160.41) .. (150.07,164.34) .. controls (140.07,168.27) and (126.29,157.02) .. (119.28,139.2) .. controls (112.27,121.38) and (114.7,103.75) .. (124.69,99.82) -- cycle ;
\draw  [draw opacity=0][fill={rgb, 255:red, 87; green, 120; blue, 164 }  ,fill opacity=0.3 ] (158.26,92.54) .. controls (166.49,91.72) and (174.45,104.07) .. (176.04,120.11) .. controls (177.63,136.16) and (172.26,149.83) .. (164.03,150.65) .. controls (155.8,151.46) and (147.84,139.12) .. (146.25,123.07) .. controls (144.66,107.03) and (150.03,93.36) .. (158.26,92.54) -- cycle ;
\draw  [draw opacity=0][fill={rgb, 255:red, 87; green, 120; blue, 164 }  ,fill opacity=0.3 ] (185.25,97.69) .. controls (191.89,98.36) and (196.16,109.89) .. (194.79,123.44) .. controls (193.41,136.99) and (186.92,147.43) .. (180.28,146.76) .. controls (173.64,146.08) and (169.37,134.55) .. (170.75,121.01) .. controls (172.12,107.46) and (178.62,97.02) .. (185.25,97.69) -- cycle ;
\draw  [draw opacity=0][fill={rgb, 255:red, 87; green, 120; blue, 164 }  ,fill opacity=0.3 ] (204.19,109.88) .. controls (209.69,111.63) and (211.49,121.44) .. (208.21,131.79) .. controls (204.92,142.14) and (197.8,149.12) .. (192.29,147.38) .. controls (186.79,145.63) and (184.99,135.82) .. (188.28,125.47) .. controls (191.56,115.11) and (198.69,108.13) .. (204.19,109.88) -- cycle ;
\draw  [draw opacity=0][fill={rgb, 255:red, 214; green, 40; blue, 40 }  ,fill opacity=0.3 ] (228.1,109.76) .. controls (230.63,114.96) and (224.76,123.02) .. (215,127.78) .. controls (205.23,132.53) and (195.26,132.18) .. (192.73,126.99) .. controls (190.21,121.8) and (196.07,113.73) .. (205.84,108.98) .. controls (215.6,104.22) and (225.57,104.57) .. (228.1,109.76) -- cycle ;
\draw  [draw opacity=0][fill={rgb, 255:red, 214; green, 40; blue, 40 }  ,fill opacity=0.3 ] (223.43,90.76) .. controls (227.09,96.35) and (220.81,106.92) .. (209.42,114.38) .. controls (198.02,121.84) and (185.82,123.36) .. (182.17,117.78) .. controls (178.52,112.19) and (184.79,101.62) .. (196.18,94.16) .. controls (207.58,86.7) and (219.78,85.18) .. (223.43,90.76) -- cycle ;
\draw  [draw opacity=0][fill={rgb, 255:red, 214; green, 40; blue, 40 }  ,fill opacity=0.3 ] (213.36,70.88) .. controls (218.67,77.22) and (212.97,90.75) .. (200.61,101.12) .. controls (188.26,111.48) and (173.94,114.75) .. (168.63,108.42) .. controls (163.31,102.09) and (169.02,88.55) .. (181.37,78.19) .. controls (193.73,67.82) and (208.05,64.55) .. (213.36,70.88) -- cycle ;
\draw  [draw opacity=0][fill={rgb, 255:red, 214; green, 40; blue, 40 }  ,fill opacity=0.3 ] (196.05,44.71) .. controls (203.83,52.11) and (199.44,69.36) .. (186.23,83.23) .. controls (173.03,97.09) and (156.03,102.33) .. (148.25,94.93) .. controls (140.47,87.52) and (144.86,70.27) .. (158.06,56.41) .. controls (171.26,42.54) and (188.27,37.3) .. (196.05,44.71) -- cycle ;
\draw  [draw opacity=0][fill={rgb, 255:red, 214; green, 40; blue, 40 }  ,fill opacity=0.3 ] (171.19,13.69) .. controls (181.88,21.63) and (179.01,43.57) .. (164.79,62.69) .. controls (150.57,81.8) and (130.38,90.85) .. (119.7,82.9) .. controls (109.01,74.95) and (111.88,53.02) .. (126.1,33.9) .. controls (140.32,14.79) and (160.51,5.74) .. (171.19,13.69) -- cycle ;
\draw  [color={rgb, 255:red, 214; green, 40; blue, 40 }  ,draw opacity=0.3 ][fill={rgb, 255:red, 214; green, 40; blue, 40 }  ,fill opacity=0.3 ] (226.6,125.35) .. controls (227.55,127.3) and (224.26,130.85) .. (219.26,133.29) .. controls (214.25,135.73) and (209.43,136.13) .. (208.48,134.18) .. controls (207.53,132.23) and (210.81,128.67) .. (215.82,126.23) .. controls (220.83,123.79) and (225.65,123.4) .. (226.6,125.35) -- cycle ;
\draw  [draw opacity=0][fill={rgb, 255:red, 87; green, 120; blue, 164 }  ,fill opacity=0.3 ] (217.84,124.13) .. controls (219.86,125.39) and (218.47,131.25) .. (214.74,137.21) .. controls (211,143.18) and (206.34,146.99) .. (204.32,145.72) .. controls (202.3,144.46) and (203.69,138.6) .. (207.42,132.63) .. controls (211.15,126.67) and (215.82,122.86) .. (217.84,124.13) -- cycle ;
\draw [color={rgb, 255:red, 87; green, 120; blue, 164 }  ,draw opacity=1 ] [dash pattern={on 0.84pt off 2.51pt}]  (222,140.5) -- (211.08,134.92) ;
\draw [shift={(222,140.5)}, rotate = 207.05] [color={rgb, 255:red, 87; green, 120; blue, 164 }  ,draw opacity=1 ][fill={rgb, 255:red, 87; green, 120; blue, 164 }  ,fill opacity=1 ][line width=0.75]      (0, 0) circle [x radius= 3.35, y radius= 3.35]   ;
\draw [color={rgb, 255:red, 87; green, 120; blue, 164 }  ,draw opacity=1 ] [dash pattern={on 0.84pt off 2.51pt}]  (211.08,134.92) -- (198.24,128.63) ;
\draw [color={rgb, 255:red, 87; green, 120; blue, 164 }  ,draw opacity=1 ] [dash pattern={on 0.84pt off 2.51pt}]  (198.24,128.63) -- (182.77,122.22) ;
\draw [color={rgb, 255:red, 87; green, 120; blue, 164 }  ,draw opacity=1 ] [dash pattern={on 0.84pt off 2.51pt}]  (182.77,122.22) -- (161.15,121.59) ;
\draw [color={rgb, 255:red, 87; green, 120; blue, 164 }  ,draw opacity=1 ] [dash pattern={on 0.84pt off 2.51pt}]  (161.15,121.59) -- (137.38,132.08) ;
\draw [color={rgb, 255:red, 87; green, 120; blue, 164 }  ,draw opacity=1 ] [dash pattern={on 0.84pt off 2.51pt}]  (137.38,132.08) -- (111.71,145.95) ;
\draw [draw opacity=0] [dash pattern={on 0.84pt off 2.51pt}]  (222,140.5) -- (217.54,129.76) ;
\draw [color={rgb, 255:red, 209; green, 97; blue, 93 }  ,draw opacity=1 ] [dash pattern={on 0.84pt off 2.51pt}]  (217.54,129.76) -- (210.42,118.38) ;
\draw [color={rgb, 255:red, 209; green, 97; blue, 93 }  ,draw opacity=1 ] [dash pattern={on 0.84pt off 2.51pt}]  (210.42,118.38) -- (202.8,104.27) ;
\draw [color={rgb, 255:red, 209; green, 97; blue, 93 }  ,draw opacity=1 ] [dash pattern={on 0.84pt off 2.51pt}]  (202.8,104.27) -- (190.99,89.65) ;
\draw [color={rgb, 255:red, 209; green, 97; blue, 93 }  ,draw opacity=1 ] [dash pattern={on 0.84pt off 2.51pt}]  (190.99,89.65) -- (172.15,69.82) ;
\draw [color={rgb, 255:red, 209; green, 97; blue, 93 }  ,draw opacity=1 ] [dash pattern={on 0.84pt off 2.51pt}]  (172.15,69.82) -- (145.44,48.29) ;
\draw [color={rgb, 255:red, 209; green, 97; blue, 93 }  ,draw opacity=1 ][fill={rgb, 255:red, 209; green, 97; blue, 93 }  ,fill opacity=1 ]   (232,50.5) -- (222,67.75) ;
\draw [color={rgb, 255:red, 209; green, 97; blue, 93 }  ,draw opacity=1 ][fill={rgb, 255:red, 209; green, 97; blue, 93 }  ,fill opacity=1 ]   (232,63.25) -- (223.5,77.9) ;
\draw [shift={(222,80.5)}, rotate = 300.1] [fill={rgb, 255:red, 209; green, 97; blue, 93 }  ,fill opacity=1 ][line width=0.08]  [draw opacity=0] (8.93,-4.29) -- (0,0) -- (8.93,4.29) -- cycle    ;
\draw [color={rgb, 255:red, 209; green, 97; blue, 93 }  ,draw opacity=1 ][fill={rgb, 255:red, 209; green, 97; blue, 93 }  ,fill opacity=1 ]   (232,63.25) -- (222,67.75) ;

\draw [color={rgb, 255:red, 87; green, 120; blue, 164 }  ,draw opacity=1 ]   (168,169.5) -- (172.5,177.5) ;
\draw [color={rgb, 255:red, 87; green, 120; blue, 164 }  ,draw opacity=1 ]   (172.5,177.5) -- (180,153.5) ;

\draw  [color={rgb, 255:red, 231; green, 202; blue, 96 }  ,draw opacity=1 ][fill={rgb, 255:red, 231; green, 202; blue, 96 }  ,fill opacity=0.05 ] (92,70.5) .. controls (165,79.5) and (147.61,60.12) .. (192,80.5) .. controls (236.39,100.88) and (218.23,139.74) .. (250.61,141.12) .. controls (283,142.5) and (285,158.5) .. (282,190.5) .. controls (279,222.5) and (221,200.5) .. (152,200.5) .. controls (83,200.5) and (51,229.5) .. (42,200.5) .. controls (33,171.5) and (19,61.5) .. (92,70.5) -- cycle ;

\draw (63,177) node  [font=\Large,color={rgb, 255:red, 106; green, 159; blue, 88 }  ,opacity=1 ] [align=left] {$\displaystyle \mathbb{T}$};
\draw (301,205) node  [font=\Large] [align=left] {$\displaystyle \mathbb{S}$};
\draw (262,188) node  [font=\Large,color={rgb, 255:red, 231; green, 202; blue, 96 }  ,opacity=1 ] [align=left] {$\displaystyle \mathbb{C}$};
\draw (85.5,253) node  [font=\Large] [align=left] {$\displaystyle \mathbb{C}$: Certain Set};
\draw (252.5,253) node  [font=\Large] [align=left] {$\displaystyle \mathbb{T}$: Terminal Set};
\draw (95.9,282) node  [font=\Large] [align=left] {$\displaystyle \mathbb{S}$: Constraint Set};
\draw (41.65,35.16) node  [font=\Large] [align=left] {$\displaystyle \mathcal{S}$};
\draw (234,146.5) node  [font=\Large] [align=left] {$\displaystyle s_{t}$};
\draw (247.5,284) node  [font=\Large] [align=left] {$\displaystyle \mathcal{S}$: State Space};

\end{tikzpicture}

}
        \vspace{-1em}
        \caption{Safety filtering mechanism.}
        \label{fig:upsi_pred}
    \end{subfigure}
    \begin{subfigure}[b]{0.32\textwidth}
        \centering
        \tikzset{every picture/.style={line width=0.75pt}} %

\resizebox{\linewidth}{!}{
\begin{tikzpicture}[x=0.75pt,y=0.75pt,yscale=-1,xscale=1]
\clip (0,0) rectangle (320, 268);

\draw  [color={rgb, 255:red, 209; green, 97; blue, 93 }  ,draw opacity=0.15 ][fill={rgb, 255:red, 255; green, 255; blue, 255 }  ,fill opacity=1 ] (140.56,209.35) .. controls (142.41,222.89) and (133.23,235.33) .. (120.04,237.13) .. controls (106.86,238.93) and (94.67,229.42) .. (92.82,215.88) .. controls (90.97,202.33) and (100.15,189.89) .. (113.34,188.09) .. controls (126.52,186.29) and (138.71,195.8) .. (140.56,209.35) -- cycle ;
\draw  [color={rgb, 255:red, 209; green, 97; blue, 93 }  ,draw opacity=0.3 ][fill={rgb, 255:red, 255; green, 255; blue, 255 }  ,fill opacity=1 ] (124.37,205.37) .. controls (126.24,218.99) and (117.06,231.49) .. (103.88,233.29) .. controls (90.69,235.09) and (78.5,225.52) .. (76.64,211.9) .. controls (74.77,198.29) and (83.95,185.79) .. (97.13,183.98) .. controls (110.31,182.18) and (122.51,191.75) .. (124.37,205.37) -- cycle ;
\draw  [color={rgb, 255:red, 209; green, 97; blue, 93 }  ,draw opacity=0.45 ][fill={rgb, 255:red, 255; green, 255; blue, 255 }  ,fill opacity=1 ] (108.47,199.6) .. controls (110.3,213.01) and (101.1,225.34) .. (87.92,227.15) .. controls (74.74,228.95) and (62.56,219.54) .. (60.73,206.13) .. controls (58.89,192.72) and (68.09,180.39) .. (81.28,178.58) .. controls (94.46,176.78) and (106.63,186.19) .. (108.47,199.6) -- cycle ;
\draw  [color={rgb, 255:red, 209; green, 97; blue, 93 }  ,draw opacity=0.6 ][fill={rgb, 255:red, 255; green, 255; blue, 255 }  ,fill opacity=1 ] (92.64,191.33) .. controls (94.42,204.33) and (85.18,216.33) .. (71.99,218.14) .. controls (58.81,219.94) and (46.68,210.86) .. (44.9,197.86) .. controls (43.13,184.87) and (52.37,172.86) .. (65.55,171.06) .. controls (78.74,169.26) and (90.86,178.33) .. (92.64,191.33) -- cycle ;
\draw  [color={rgb, 255:red, 209; green, 97; blue, 93 }  ,draw opacity=0.75 ][fill={rgb, 255:red, 255; green, 255; blue, 255 }  ,fill opacity=1 ] (80.74,177.66) .. controls (82.54,190.84) and (73.32,202.98) .. (60.13,204.79) .. controls (46.95,206.59) and (34.8,197.37) .. (33,184.19) .. controls (31.2,171) and (40.42,158.86) .. (53.6,157.05) .. controls (66.79,155.25) and (78.94,164.47) .. (80.74,177.66) -- cycle ;
\draw  [color={rgb, 255:red, 209; green, 97; blue, 93 }  ,draw opacity=0.9 ][fill={rgb, 255:red, 255; green, 255; blue, 255 }  ,fill opacity=1 ] (73.18,158.99) .. controls (74.97,172.14) and (65.75,184.27) .. (52.56,186.07) .. controls (39.38,187.88) and (27.24,178.68) .. (25.44,165.52) .. controls (23.64,152.37) and (32.86,140.24) .. (46.05,138.44) .. controls (59.23,136.63) and (71.38,145.84) .. (73.18,158.99) -- cycle ;
\draw  [color={rgb, 255:red, 0; green, 0; blue, 0 }  ,draw opacity=1 ][dash pattern={on 4.5pt off 4.5pt}] (266.23,132.72) .. controls (271.16,168.7) and (226.81,204.48) .. (167.18,212.63) .. controls (107.56,220.79) and (55.23,198.24) .. (50.31,162.26) .. controls (45.38,126.27) and (89.73,90.49) .. (149.36,82.34) .. controls (208.98,74.18) and (261.31,96.73) .. (266.23,132.72) -- cycle ;
\draw  [color={rgb, 255:red, 0; green, 0; blue, 0 }  ,draw opacity=1 ] (294.05,130.16) .. controls (300.83,179.72) and (246.08,228.14) .. (171.76,238.31) .. controls (97.45,248.47) and (31.71,216.54) .. (24.93,166.98) .. controls (18.15,117.42) and (72.89,69) .. (147.21,58.83) .. controls (221.52,48.67) and (287.27,80.6) .. (294.05,130.16) -- cycle ;
\draw  [draw opacity=0][fill={rgb, 255:red, 255; green, 255; blue, 255 }  ,fill opacity=1 ] (135.3,232.71) -- (195.3,232.71) -- (195.3,262.71) -- (135.3,262.71) -- cycle ;
\draw  [draw opacity=0][fill={rgb, 255:red, 255; green, 255; blue, 255 }  ,fill opacity=1 ] (31.89,145.41) -- (64,145.41) -- (64,178) -- (31.89,178) -- cycle ;
\draw  [color={rgb, 255:red, 31; green, 119; blue, 180 }  ,draw opacity=0.17 ][fill={rgb, 255:red, 255; green, 255; blue, 255 }  ,fill opacity=1 ] (189.21,106.03) .. controls (189.31,118.81) and (173.49,129.29) .. (153.88,129.44) .. controls (134.27,129.59) and (118.3,119.35) .. (118.2,106.57) .. controls (118.1,93.8) and (133.92,83.32) .. (153.53,83.17) .. controls (173.13,83.02) and (189.11,93.25) .. (189.21,106.03) -- cycle ;
\draw  [color={rgb, 255:red, 31; green, 119; blue, 180 }  ,draw opacity=0.4 ][fill={rgb, 255:red, 255; green, 255; blue, 255 }  ,fill opacity=1 ] (202.04,104.2) .. controls (202.14,116.98) and (186.32,127.46) .. (166.71,127.61) .. controls (147.1,127.76) and (131.13,117.52) .. (131.03,104.74) .. controls (130.93,91.97) and (146.75,81.49) .. (166.36,81.34) .. controls (185.97,81.19) and (201.94,91.42) .. (202.04,104.2) -- cycle ;
\draw  [color={rgb, 255:red, 31; green, 119; blue, 180 }  ,draw opacity=0.58 ][fill={rgb, 255:red, 255; green, 255; blue, 255 }  ,fill opacity=1 ] (213.04,104.2) .. controls (213.14,116.98) and (197.32,127.46) .. (177.71,127.61) .. controls (158.1,127.76) and (142.13,117.52) .. (142.03,104.74) .. controls (141.93,91.97) and (157.75,81.49) .. (177.36,81.34) .. controls (196.97,81.19) and (212.94,91.42) .. (213.04,104.2) -- cycle ;
\draw  [color={rgb, 255:red, 31; green, 119; blue, 180 }  ,draw opacity=0.84 ][fill={rgb, 255:red, 255; green, 255; blue, 255 }  ,fill opacity=1 ] (224.71,105.76) .. controls (224.81,118.53) and (209,129.01) .. (189.39,129.16) .. controls (169.78,129.31) and (153.8,119.08) .. (153.7,106.3) .. controls (153.61,93.53) and (169.42,83.05) .. (189.03,82.89) .. controls (208.64,82.74) and (224.62,92.98) .. (224.71,105.76) -- cycle ;
\draw  [color={rgb, 255:red, 31; green, 119; blue, 180 }  ,draw opacity=1 ][fill={rgb, 255:red, 255; green, 255; blue, 255 }  ,fill opacity=1 ] (237.13,109.03) .. controls (237.23,121.8) and (221.41,132.28) .. (201.81,132.43) .. controls (182.2,132.59) and (166.22,122.35) .. (166.12,109.57) .. controls (166.02,96.8) and (181.84,86.32) .. (201.45,86.17) .. controls (221.06,86.01) and (237.03,96.25) .. (237.13,109.03) -- cycle ;
\draw  [color={rgb, 255:red, 0; green, 0; blue, 0 }  ,draw opacity=1 ][dash pattern={on 0.84pt off 2.51pt}] (230.42,137.62) .. controls (233.51,160.21) and (204.62,182.82) .. (165.9,188.11) .. controls (127.17,193.41) and (93.27,179.39) .. (90.18,156.8) .. controls (87.09,134.21) and (115.98,111.6) .. (154.7,106.3) .. controls (193.43,101) and (227.33,115.02) .. (230.42,137.62) -- cycle ;
\draw  [draw opacity=0][fill={rgb, 255:red, 255; green, 255; blue, 255 }  ,fill opacity=1 ] (179.54,96.77) -- (221,96.77) -- (221,121) -- (179.54,121) -- cycle ;

\draw (158.3,154.21) node  [font=\LARGE] [align=left] {$\displaystyle F_{n|t} P_{n|t} F_{n|t}^{\top }$};
\draw (167.3,250.71) node  [font=\LARGE,color={rgb, 255:red, 0; green, 0; blue, 0 }  ,opacity=1 ] [align=left] {$\displaystyle P_{n+1|t}$};
\draw (50.31,164.26) node  [font=\LARGE,color={rgb, 255:red, 209; green, 97; blue, 93 }  ,opacity=1 ] [align=left] {$\displaystyle e_{n|t}^{\text{max}}$};
\draw (201.63,109.3) node  [font=\LARGE,color={rgb, 255:red, 31; green, 119; blue, 180 }  ,opacity=1 ] [align=left] {$\displaystyle \epsilon \overline{\Sigma }_{n|t}$};

\end{tikzpicture}}
        \vspace{-1em}
        \caption{Reachable set construction.}
        \label{fig:upsi_elli}
    \end{subfigure}
    \vspace{-1em}
    \caption{Uncertainty-Aware Predictive Safety Filter (\algo): \emph{(a) \algo~leverages the same PE dynamics model used in state-of-the-art MBRL. (b) \algo~enforces safety by ensuring that predicted reachable sets satisfy constraints $\stateset$ and remain within accurately modeled regions $\accset$.\\ (c) \algo~accounts for uncertainty accumulation and modeling errors $e$ in reachable sets. }}
    \label{fig:fig1}
\vspace{-1em}
\end{figure}

Building on recent results in uncertainty propagation for PE models~\citep{frauenknecht_rollouts_2025}, we present the \emph{\textbf{U}ncertainty-Aware \textbf{P}redictive \textbf{S}afety F\textbf{i}lter}~(\algo)\footnote{pronounced: oop-see}, a PSF that makes
rigorous predictions of future outcomes using a PE dynamics model.
As illustrated in Figure~\ref{fig:upsi_block}, \algo~seamlessly integrates into Dyna-style
MBRL frameworks like model-based policy optimization (MBPO)~\citep{janner_when_2019-1}. Its formulation of future outcomes as
reachable sets, depicted in Figures~\ref{fig:upsi_pred} and~\ref{fig:upsi_elli}, makes it compatible with a broad range of MPC schemes.
\algo~aims to combine the scalability of MBRL methods with the rigor of MPC approaches to enforce exploration safety. In particular,\\
\mbox{}\hspace{4mm}$\bullet$ we obtain a rigorous notion of reachable sets from a probabilistic neural network model;\\
\mbox{}\hspace{4mm}$\bullet$ we develop an explicit certainty constraint to avoid model exploitation; \\
\mbox{}\hspace{4mm}$\bullet$ we empirically validate the reachable set overapproximation and evaluate a practical simplifi- \\ 
\mbox{}\hspace{6mm} cation of \algo integrated into MBPO on common safe exploration benchmarks.~\algo yields \\ 
\mbox{}\hspace{6mm} substantial improvements in exploration safety compared to prior neural network PSFs while\\
\mbox{}\hspace{6mm} achieving performance comparable to standard MBPO. 

\section{Background}
\subsection{Reinforcement Learning}
Reinforcement learning addresses sequential decision-making by having an agent learn through interaction with an environment modeled as a Markov decision process (MDP), defined by the tuple $\mdp = \{ \statesp, \actsp, \rewsp, \dyn, \initdist, \discount \}$. At each step $\globid$, the agent applies action $\rvact_\globid \in \actsp \subseteq \mathbb{R}^{n_\actsp}$ in state $\rvstate_\globid \in \statesp \subseteq \mathbb{R}^{\dimS}$, transitions according to $\rvstate_{\globid+1} \sim \dyn(\cdot \mid \rvstate_\globid, \rvact_\globid)$, and receives reward $\rvrew_{\globid+1} = \rewfcn(\rvstate_\globid, \rvact_\globid, \rvstate_{\globid+1})$.
Starting from a set of initial states $\rvstate_0 \in \initdist$, the objective is to find a policy $\rvact_\globid \sim \pol(\cdot | \rvstate_\globid)$ maximizing the expected return discounted by $\discount$, i.e.,
    $\max_\pol \; \E_\pol \!\left[ \sum_{\globid=0}^\infty \discount^\globid \rvrew_{\globid+1} \right].$
Where required, we use $\rvstate_\globid$, $\rvact_\globid$, $\rvrew_\globid$ for random variables and $\state_\globid$, $\act_\globid$, $\rew_\globid$ for their realizations. An episode index $\epiid$ is incremented at each environment reset, such that $\buffer^\epiid$ denotes the replay buffer containing all transitions up to episode $\epiid$.

\subsection{Dynamics Modeling} 
We consider stochastic nonlinear systems with additive heteroscedastic process noise of the form
\begin{equation}
\label{eq:env_dyn}
\begin{aligned}
    & \rvstate_{\globid + 1} = \meanenv(\rvstate_\globid, \rvact_\globid) +\cholenv(\rvstate_\globid, \rvact_\globid)\rvalea_\globid \text{ with } \rvalea_\globid  \in \aleaset \text{ and } \E[\rvalea_\globid] = 0,\\
\end{aligned}
\end{equation}
where $\aleaset = \{ w \in \mathbb{R}^{\dimS } \mid \| w \|_2^2 \leq \ellscale\}$ for $\ellscale \geq 0$ with nominal dynamics $\meanenv(\rvstate_\globid, \rvact_\globid)$. The process noise $\rvalea_\globid$ is bounded within the hyperball $\aleaset$ and scaled by a diagonal covariance matrix  $\varenv(\rvstate_\globid, \rvact_\globid) = \cholenv(\rvstate_\globid, \rvact_\globid) \cholenv(\rvstate_\globid, \rvact_\globid)^\top$. 
PE models~\citep{lakshminarayanan_simple_2017} approximate the dynamics in~\eqref{eq:env_dyn} via $\nummods$ probabilistic neural networks with parameters $\params_\modid$, where $\modid \in \{1, \dots, \nummods\}$. 
The model state is
\begin{equation}
\label{eq:mod_dyn}
    \hat{\rvstate}_{\globid + 1} =  \meanmod_{\params_\modid}(\rvstate_\globid, \rvact_\globid) + \cholmod_{\params_\modid}(\rvstate_\globid, \rvact_\globid) \rvmodalea_\globid \text{ with } \rvmodalea_\globid \sim \gauss(0, I),
\end{equation}
which comprises nominal dynamics $\meanmod_{\params_\modid}(\rvstate_\globid, \rvact_\globid)$ and the diagonal covariance $\varmod_{\params_\modid}(\rvstate_\globid, \rvact_\globid) = \cholmod_{\params_\modid}(\rvstate_\globid, \rvact_\globid) \cholmod_{\params_\modid}(\rvstate_\globid, \rvact_\globid)^\top$ that scales normally distributed process noise $\rvmodalea_\globid$.
The PE separates aleatoric uncertainty, representing irreducible process noise, captured by $\varmod_{\params_\modid}$, from epistemic uncertainty due to lack of data, estimated via ensemble disagreement. Assuming sufficient model capacity, the absence of epistemic uncertainty indicates predictive accuracy~\citep{frauenknecht_trust_2024}, enabling the definition of a certain set $\accset \subseteq \statesp \times \actsp$ within which predictions closely capture the environment dynamics \eqref{eq:env_dyn}. We denote the certain set of a PE model trained on data $\buffer^\epiid$ as $\accset^\epiid$.

\subsection{Predictive Safety Filtering and Model Predictive Control}
\label{subsec:safe_exploration}
PSFs are closely related to MPC~\citep{rawlings_model_2024}, where a constrained optimal control problem is solved by planning with a dynamics model over a finite horizon $\predhorizon$ in a receding fashion. At each time step $\globid$, MPC computes a control sequence $\mathcal{U}_\globid = [\predact_{0|\globid}, \dots, \predact_{\predhorizon-1|\globid}]$ and corresponding predicted state sequence $\mathcal{Z}_\globid = [\state_{0|\globid}, \dots, \state_{\predhorizon|\globid}]$ by solving\looseness=-1
\begin{equation}
\label{eq:nominal_mpc}
    \mathcal{U}^*_\globid = \arg\min_{\mathcal{U}_\globid} J(\mathcal{U}_\globid, \mathcal{Z}_\globid)
    \quad \text{s.t.} \quad \state_{\predid|\globid} \in \stateset,\ \predact_{\predid|\globid} \in \actset,\ \state_{\predhorizon|\globid} \in \termset,
\end{equation}
where $\state_{\predid|\globid}$ denotes the predicted state $\predid$ steps ahead from time $\globid$, $\stateset \subseteq \statesp$ and $\actset \subseteq \actsp$ are the state and input constraint sets, and $\termset \subseteq \stateset$ is a terminal set within which the system can be stabilized indefinitely. 
PSFs are a special case of MPC where the objective is to minimize the distance to the RL action, such that $ J(\mathcal{U}_\globid, \mathcal{Z}_\globid) = \| \act_\globid - \predact_{0|\globid}\|_2^2$ and $\predact_{0|\globid}$ only deviates from $\act_\globid$ if necessary to satisfy the constraints in \eqref{eq:nominal_mpc}. The first PSF input $\act^{\mathrm{F}}_\globid = \predact^*_{0|\globid}$ is applied as a filtered action, and problem \eqref{eq:nominal_mpc} is re-solved upon observing $\state_{\globid+1}$. 
Robust MPC addresses systems such as \eqref{eq:env_dyn}, where the bounded process noise introduces ambiguity. Thus, a control sequence yields a sequence of robustly reachable sets (RRS) as per Definition \ref{def:robust_reachable_set}.
\begin{definition}[Robust Reachable Set]
\label{def:robust_reachable_set}
   Let $\prs_{\predid|\globid}(\state_\globid, \predact_{0|\globid}, \dots, \predact_{\predid-1|\globid}): \statesp \times \actsp^\predid \rightarrow \statesp$ for $\predid \in [1, \dots, \predhorizon]$ be the robust reachable set, such that $\state_{\globid + \predid} \in \prs_{\predid|\globid}(\state_\globid, \predact_{0|\globid}, \dots, \predact_{\predid-1|\globid}):=\prs_{\predid|\globid}$ under dynamics \eqref{eq:env_dyn} for the control sequence $\predact_{0|\globid}, \dots, \predact_{\predid-1|\globid}$ and all possible $\alea_\globid, \dots, \alea_{\globid+\predid - 1} \in \aleaset^{\predid}$.
\end{definition}
Planning through RRSs is typically performed along a nominal trajectory $\mathcal{Z}_\globid = [\predmean_{0|\globid}, \dots, \predmean_{\predhorizon|\globid}]$ with $\predmean_{\predid|\globid} \in \prs_{\predid|\globid}$. Growth of the RRS is commonly limited by augmenting the nominal control sequence with an ancillary controller such that $v_{\predid|\globid} = \predact_{\predid|\globid} + \contgain_{\predid|\globid}(\state_{\predid|\globid} - \predmean_{\predid|\globid})$ for all $\state_{\predid|\globid} \in \prs_{\predid|\globid}$. The ancillary feedback controller $\contgain_{\predid|\globid}$ is typically chosen based on $\predmean_{\predid|\globid}$.
To stabilize the stochastic system indefinitely, $\termset$ needs to be robustly positive invariant as per Definition \ref{def:robst_pos_inv}.
\begin{definition}[Robust Positive Invariance]
\label{def:robst_pos_inv}
    A set $\mathbb{X}\subseteq \statesp$ is robustly positive invariant under a control law $\act_\globid = \initpol(\state_\globid)$ for the dynamics in \eqref{eq:env_dyn}, i.e., $\state_{\globid + 1} = \meanenv(\state_\globid, \kappa(\state_\globid)) + \cholenv(\state_\globid, \initpol(\state_\globid))\alea_\globid$ and $\alea_\globid \in \mathbb{W}$, if for every $\state_\globid \in \mathbb{X}$, $\meanenv(\state_\globid, \kappa(\state_\globid)) + \cholenv(\state_\globid,  \initpol(\state_\globid))\mathbb{W} \subseteq \mathbb{X}$ (see \cite{rawlings_model_2024}, Definition 3.7).
\end{definition}

\section{Related Work}
\label{sec:related_work}

Various approaches address exploration safety in RL. Constrained MDP formulations~\citep{achiam_constrained_2017, ray2019benchmarking} impose constraints on expected cumulative constraint violations. This, however, encourages safe behavior rather than enforcing it, which is insufficient for many applications. 
Safety filters provide stronger guarantees by ensuring constraint satisfaction with high probability. At every time step, the RL action is evaluated and, if necessary, projected onto the closest safe action. Existing approaches can be clustered into Hamilton-Jacobi reachability (HJR) methods~\citep{bansal_hamilton-jacobi_2017}, control barrier functions (CBFs)~\citep{ames_control_2019}, and predictive safety filters (PSFs)~\citep{wabersich_linear_2018} with a comprehensive overview provided in \citet{wabersich_data-driven_2023} and \citet{brunke_safe_2022}. While HJR methods require substantial domain knowledge and scale poorly with dimension, CBFs become increasingly difficult to find for high-dimensional systems. We focus on PSFs, which generally scale more favorably~\citep{wabersich_data-driven_2023}.
PSFs are closely related to MPC and typically rely on first-principles models~\citep{wabersich_linear_2018, bejarano_safety_2025} or data-driven models such as Bayesian linear regression~\citep{wabersich_predictive_2021, wabersich_probabilistic_2022} or GPs~\citep{koller_learning-based_2018, goodall2026safe}. While these enable rigorous reachable set computation, they either require substantial domain knowledge or scale poorly with large datasets. \citet{gronauer_reinforcement_2024} integrate PE dynamics models into PSFs, with minor extensions presented in~\citet{guzelkaya_ensemble_2024}. However, both works neglect potential model exploitation under high epistemic uncertainty, resulting in heuristic reachable set computation. The consequences are frequent failures of the resulting PSF as presented in Section \ref{sec:experiments}. \algo provides a rigorous reachable set computation comparable to common MPC model classes while leveraging the strong representational capacity of PE models. 

\section{Problem Setting}
Similar to \citet{koller_learning-based_2018}, we assume access to an initial data-generating controller as stated in Assumption \ref{assumption:init} that induces an initial certain set $\accset^0$ within which \algo~can operate.

\begin{assumption}[Initial controller]
\label{assumption:init}
Let $\initpol$ be a controller with $\initpol(s) \in \actset$ for all $s\in\stateset$, inducing a terminal set $\termset^0 \subseteq \stateset$ robustly positively invariant around $\initdist$. Training the PE model on data $\buffer^0$ generated by $\initpol$ yields an initial certain set $\accset^0$ such that $\termset^0 \times \mathbb{A}^0 \subseteq \accset^0$ and $\initpol(s)\in\mathbb{A}^0$ for all $s\in\termset^0$.
\end{assumption}

Starting from this initialization, \algo guides exploration by ensuring the system remains in the constraint-satisfying subsets $\stateset \subseteq \statesp$ and $\actset \subseteq \actsp$.
This is achieved by solving a robust PSF problem 
\begin{subequations}\label{eq:mpc_problem_statement}
\begin{align}
    \min_{\predact_{0|\globid}, \dots, \predact_{ \predhorizon-1|\globid}} & \|\act_\globid - \predact_{0|\globid} \|_2^2 \label{eq:mpc_cost_prob} \\
    \text{s.t. } & \prs_{\predid|\globid}(\state_\globid, \predact_{0|\globid}, \dots, \predact_{ \predid-1|\globid}) \times v_{ \predid|\globid} \subseteq \stateset \times \actset \cap \accset^\epiid \label{eq:mpc_state_const} \quad \forall \predid \in [1, \dots, \predhorizon-1] \\
                & \prs_{\predhorizon|\globid}(\state_\globid, \predact_{0|\globid}, \dots, \predact_{ \predhorizon-1|\globid}) \subseteq \termset^\epiid ,\label{eq:mpc_term_const}
\end{align}
\end{subequations}
that differs from standard formulations solely by the certainty constraint in $\accset^\epiid$. Throughout training, $\accset^\epiid$ gradually grows as data is observed.
Given $\prs_{\predid|\globid}$, $\accset^\epiid$, and $\termset^\epiid$, several robust MPC schemes \citep{cannon2011robust, houska2018robust, kohler_computationally_2021} are well suited to solve \eqref{eq:mpc_problem_statement}.

The key contribution of this work is obtaining these ingredients from a learned neural network model. Leveraging properties of the PE architecture, we specifically derive: \\
\mbox{}\hspace{7mm} (i) the certain set $\accset^\epiid$ for episode $\epiid$; \\
\mbox{}\hspace{7mm} (ii) an overapproximation of the reachable set sequence $\prs_\predid(\state_\globid, \predact_{0|\globid}, \dots, \predact_{\predid-1|\globid})$; and \\
\mbox{}\hspace{7mm} (iii) a data-driven expansion of the terminal set $\termset^\epiid$.

\section{Uncertainty-Aware Predictive Safety Filtering}
\label{sec:method}
In the following, we introduce the notion of the certain set $\accset^\epiid$ in Section \ref{subsec:cert_set}. Within $\accset^\epiid$, we develop the overapproximation to reachable sets $\prs_\predid(\state_\globid, \predact_{0|\globid}, \dots, \predact_{\predid-1|\globid})$ in Section \ref{subsec:reach_set}. We present \algo in Section \ref{subsec:mpc_setting} and introduce the expansion of $\termset^\epiid$ in Section \ref{subsec:term_set}.

We require several assumptions for our reachable sets. Assumptions~\ref{assumption:consistent_estimator} and~\ref{assumption:unbiased_estimator} address the approximation behavior of the PE model within $\accset^\epiid$, which were originally formulated and validated by \citet{frauenknecht_rollouts_2025}. The former can be enforced via suitable variance initialization, while the latter typically holds given sufficient representational capacity. Assumption~\ref{assumption:improve} states monotonic improvement of the model for ${(\state_\globid, \act_\globid) \in \accset^\epiid}$ as more data is gathered, and is common in episodic methods \citep{didier_adaptive_2021}. Finally, Assumption~\ref{assumption:lipschitz_bounds} states regularity of the environment and model.
\begin{assumption}
    [Consistent estimator] The model is a consistent estimator of process noise, i.e.,
            $( \varmod_{\params^\epiid_\modid}(\state_\globid, \act_\globid) - \varenv(\state_\globid, \act_\globid)) \succcurlyeq 0 \text{ for all } (\state_\globid, \act_\globid) \in \accset^\epiid \text{ and } \modid \in \{ 1, \dots, \nummods \}$
        \label{assumption:consistent_estimator}
    within the certain set $\accset^\epiid \subseteq \statesp \times \actsp$, following the definition of \citet{Julier01}.
\end{assumption}
\begin{assumption}
    [Unbiased estimator] Within the certain set $\accset^\epiid \subseteq \statesp \times \actsp$, model bias is negligible, i.e.,
        $\E [ \rvmodstate_{t+1} \mid \state_\globid, \act_\globid ] = \E [ \rvstate_{t+1} \mid \state_\globid, \act_\globid ] \text{ for all } (\state_\globid, \act_\globid) \in \accset^\epiid$.
    \label{assumption:unbiased_estimator}
\end{assumption}
\begin{assumption}[Monotonic improvement]
\label{assumption:improve} The process noise estimation monotonically improves, such that
    $\varmod_{\params^\epiid_\modid}(\state_\globid, \act_\globid) \succcurlyeq \varmod_{\params^{\epiid+1}_\modid}(\state_\globid, \act_\globid) \succcurlyeq \varenv(\state_\globid, \act_\globid) \text{ for all } (\state_\globid, \act_\globid) \in \accset^\epiid, \text{ and } \modid \in \{ 1, \dots, \nummods \}$
and the certain set monotonically grows over the episode iterations $\epiid$, that is,
    $\accset^{\epiid} \subseteq \accset^{\epiid + 1} \subseteq \statesp \times \actsp.$
\end{assumption}

\begin{assumption}[Regularity of dynamics, noise scale, and certainty measure]
\label{assumption:lipschitz_bounds}
The nominal dynamics $\meanenv$ are twice continuously differentiable. Further, the nominal dynamics $\meanenv$, process noise scale $\cholenv$, and certainty measure 
$g(\state,\act):=\tfrac{1}{\dimS}\mathrm{tr}(\kalgain(\state,\act))$, as will be introduced in Section~\ref{subsec:cert_set},
are Lipschitz continuous.
That is, there exist constants $\lipschitz,\lipschitzvarchol,\lipschitzmod$ such that for all $(\state_1,\act_1),(\state_2,\act_2)\in\statesp\times\actsp$,
$\tfrac{\|\grad \meanenv(\state_1,\act_1)-\grad \meanenv(\state_2,\act_2)\|_F}{\|(\state_1,\act_1)-(\state_2,\act_2)\|_2}\le\lipschitz$,
$\tfrac{\|\cholenv(\state_1,\act_1)-\cholenv(\state_2,\act_2)\|_F}{\|(\state_1,\act_1)-(\state_2,\act_2)\|_2}\le\lipschitzvarchol$,
and $\tfrac{|g(\state_1,\act_1)-g(\state_2,\act_2)|}{\|(\state_1,\act_1)-(\state_2,\act_2)\|_2}\le\lipschitzmod$.

\end{assumption}

\subsection{Uncertainty Quantification and Certain Set}
\label{subsec:cert_set}

Under Assumptions \ref{assumption:consistent_estimator} and \ref{assumption:unbiased_estimator}, \citet{frauenknecht_rollouts_2025} propose to leverage ideas from sensor fusion \citep{Julier01} to separate epistemic stochasticity due to model error from aleatoric stochasticity describing the environment, and to solely propagate the latter to closely capture the real system. We adopt their formulation of the aleatoric variance $\mlvarmod$ as $ \mlvarmod(\predmean_{\predid \mid \globid}, \predact_{\predid \mid \globid}) = ( \frac{1}{\nummods} \sum_{\modid=1}^{\nummods} ( \varmod_{\params_\modid}(\predmean_{\predid \mid \globid}, \predact_{\predid \mid \globid}))^{-1})^{-1} $, the nominal dynamics $\mlmeanmod$ as $\mlmeanmod(\predmean_{\predid \mid \globid}, \predact_{\predid \mid \globid}) = \mlvarmod(\predmean_{\predid \mid \globid}, \predact_{\predid \mid \globid}) ( \frac{1}{\nummods} \sum_{\modid=1}^{\nummods} ( \varmod_{\params_\modid}(\predmean_{\predid \mid \globid}, \predact_{\predid \mid \globid}))^{-1} \meanmod_{\params_\modid}(\predmean_{\predid \mid \globid}, \predact_{\predid \mid \globid}) )$ and the epistemic variance $\epivarmod$ as $\epivarmod(\predmean_{\predid \mid \globid}, \predact_{\predid \mid \globid}) = \frac{1}{\nummods} \sum_{\modid=1}^{\nummods} ( \meanmod_{\params_\modid}(\predmean_{\predid \mid \globid}, \predact_{\predid \mid \globid}) - \mlmeanmod(\predmean_{\predid \mid \globid}, \predact_{\predid \mid \globid}))( \meanmod_{\params_\modid}(\predmean_{\predid \mid \globid}, \predact_{\predid \mid \globid}) - \mlmeanmod(\predmean_{\predid \mid \globid}, \predact_{\predid \mid \globid}))^\top$ of the PE model.
Thus, the aleatoric variance estimate $\mlvarmod$ consistently estimates the real process noise variance~$\varenv~$~in~\eqref{eq:env_dyn} \citep{julier_consistent_1997} under Assumption~\ref{assumption:consistent_estimator}, and the nominal dynamics $\meanenv$ are closely approximated by~$\mlmeanmod$ under Assumption~\ref{assumption:unbiased_estimator}.

The Kalman gain
$\kalgain(\predmean_{\predid \mid \globid}, \predact_{\predid \mid \globid}) = \mlvarmod(\predmean_{\predid \mid \globid}, \predact_{\predid \mid \globid}) ( \mlvarmod(\predmean_{\predid \mid \globid}, \predact_{\predid \mid \globid}) + \epivarmod(\predmean_{\predid \mid \globid}, \predact_{\predid \mid \globid}) )^{-1}$
yields the ratio of aleatoric variance to total predictive variance. We define the sufficiently certain set
\begin{equation}
    \accset := \{ (\state, \act) \in \statesp \times \actsp | \dimS^{-1} \mathrm{tr}(\kalgain(\state, \act)) \geq \xi \}
    \label{eq:certain_set}
\end{equation}
as the set of state-action pairs with sufficiently high Kalman gain. The average dimension-wise Kalman gain $1/\dimS\cdot\mathrm{tr}(\kalgain(\state, \act)) \in [0, 1]$ approaches $1$ when epistemic uncertainty is negligible and $0$ under high model uncertainty, providing a bounded, easy-to-tune certainty measure for the PSF.

\subsection{Reachable Sets}
\label{subsec:reach_set}
To predict future environment behavior, we overapproximate the robustly reachable sets by separating the nominal trajectory $\predmean$ from an error region $\errball$ resulting from accumulated process noise
\begin{equation}
\label{eq:standard_reachable_set}
    \prs_{\predid|\globid}(\state_\globid, \predact_{0|\globid}, \dots, \predact_{\predid-1|\globid}) \subseteq \predmean_{\predid|\globid} \oplus \errball_{\predid|\globid}.
\end{equation}

This standard approach from robust MPC literature \citep{soloperto2018learning, koller_learning-based_2018, kohler_computationally_2021} makes our formulation compatible with existing robust MPC schemes. Our key contribution is providing a reachable set formulation for \eqref{eq:standard_reachable_set} that provably overapproximates the environment \eqref{eq:env_dyn} by propagating ellipsoidal tubes $\tube_{n|t}=\{ s\in\statesp \ | \ (s-\predmean_{n|t})^\top (\predvar_{n|t})^{-1}(s-\predmean_{n|t})\le 1\}$ with shape matrix $\predvar_{n|t}$ along the nominal trajectory $\predmean_{n|t}$, such that $\prs_{\predid|\globid}\subseteq \tube_{n|t}$. 
We construct the tube from the estimates introduced in Section~\ref{subsec:cert_set}, where truncating the Gaussian process noise estimate $\mlvarmod$ yields the ellipsoid with shape $\epsilon\mlvarmod$ containing the system's noise from~\eqref{eq:env_dyn} (see Lemma~\ref{lemma_app:gauss_lvl_set}).

Starting from $\predmean_{0|\globid} = \state_\globid$ with $\predvar_{0|\globid} = 0$, where the zero shape matrix results in $\errball_{0|\globid} = \{0\}$ and $\tube_{0|\globid} = \{\state_\globid\}$ by the convention in Lemma~\ref{lemma_app:outer_ellipsoid}, we propagate the ellipsoidal tube forward in time.
The mean is propagated through the nominal dynamics
\begin{equation}
    \label{eq:mean_pred}
    \predmean_{\predid+1 \mid \globid} = \mlmeanmod(\predmean_{\predid \mid \globid}, \predact_{\predid \mid \globid}).
\end{equation}
The shape $\predvar_{\predid \mid \globid}$ is propagated via a first-order Taylor approximation of the nominal dynamics~\citep{hewing2018cautious, koller_learning-based_2018}.
The required state and action Jacobians are the model's average partial derivatives evaluated on the nominal trajectory pair $(\predmean_{\predid \mid \globid}, \predact_{\predid \mid \globid})$, which leads to $\sysmat_{\predid \mid \globid} = \frac{1}{\nummods}\sum_{\modid=1}^\nummods \grad_\state \meanmod_{\params_\modid} ( \predmean_{\predid \mid \globid}, \predact_{\predid \mid \globid})$ and $\inmat_{\predid \mid \globid} = \frac{1}{\nummods}\sum_{\modid=1}^\nummods \grad_\act \meanmod_{\params_\modid} ( \predmean_{\predid \mid \globid}, \predact_{\predid \mid \globid})$. 
Considering the ancillary controller $\contgain_{\predid \mid \globid}$ discussed in Section \ref{subsec:safe_exploration} yields linearized dynamics $\cloopmat_{\predid \mid \globid} := \sysmat_{\predid \mid \globid} + \inmat_{\predid \mid \globid} \contgain_{\predid \mid \globid}$, as shown in Lemma \ref{lemma_app:ancilliary_cont} of Supp. \ref{supp:proofs}. Consequently, naive tube propagation yields
 $\prepredvar_{n+1|\globid} = (1+\beta^{-1})\cloopmat_{n|\globid}\predvar_{n|\globid}\cloopmat_{n|\globid}^\top + (1+\beta)\,\ellscale\mlvarmod(\predmean_{n|\globid},\predact_{n|\globid})$ with $\beta = \sqrt{\mathrm{tr}(\cloopmat_{n|\globid}\predvar_{n|\globid}\cloopmat_{n|\globid}^\top)/\mathrm{tr}(\ellscale\mlvarmod(\predmean_{n|\globid},\predact_{n|\globid}))}$, describing an ellipsoidal outer approximation of the Minkowski sum of the propagated uncertainty $\cloopmat_{n|\globid}\predvar_{n|\globid}\cloopmat_{n|\globid}^\top$ and the scaled single-step uncertainty $\ellscale\mlvarmod(\predmean_{n|\globid},\predact_{n|\globid})$ obtained from the PE model. The construction of the ellipsoidal outer approximation is discussed in Lemma~\ref{lemma_app:outer_ellipsoid} of Appendix  \ref{app:proofs}.
However, propagating the tube shape through $\cloopmat_{\predid \mid \globid}$ introduces a linearization error that leads to a systematic underestimation of $\predvar_{\predid \mid \globid}$ such that the environment stochasticity is no longer overapproximated.  The corresponding linearization error at $(\predmean_{\predid \mid \globid}, \predact_{\predid \mid \globid})$ is upper bounded by {\small $ e^{\mathrm{max}}_{\predid \mid \globid} = \frac{\lipschitz}{2} (  \sqrt{ \lambda_\mathrm{max}(\predvar_{\predid \mid \globid})} + \sqrt{ \lambda_{\mathrm{max}}( \contgain_{\predid \mid \globid}  \predvar_{\predid \mid \globid} \contgain_{\predid \mid \globid}^\top )} )^2 + \lipschitzvarchol \sqrt{\ellscale} \sqrt{\lambda_{\mathrm{max}}( \predvar_{\predid \mid \globid} ) + \lambda_{\mathrm{max}}( \contgain_{\predid \mid \globid}  \predvar_{\predid \mid \globid} \contgain_{\predid \mid \globid}^\top )}$}, as presented in Lemmas \ref{lemma_app:lin_err_bound}, \ref{lemma_app:lin_err_var}, and \ref{lemma_app:total_lin_err} of Appendix \ref{app:proofs}.

Accounting for $e^{\mathrm{max}}_{\predid \mid \globid}$ in the shape matrix propagation yields the ellipsoidal outer approximation
\begin{equation}
\label{eq:var_pred}
    \predvar_{n+1|\globid} = (1+\tau^{-1})\,\prepredvar_{n+1|\globid} + (1+\tau)\,(e^{\max}_{n|\globid})^2 I:= \Phi ( \predvar_{\predid \mid \globid},\contgain_{\predid \mid \globid},\predmean_{\predid \mid \globid}, \predact_{\predid \mid \globid}),
\end{equation}
with $\tau = \sqrt{\mathrm{tr}(\prepredvar_{n+1|\globid})/(\dimS(e^{\max}_{n|\globid})^2)}$, that overapproximates environment behavior for $n>0$ as presented in Lemma \ref{lemma_app:onestep_ellipsoid} of Appendix \ref{app:proofs}. Consequently, propagating $\tube_{n|t}$ via \eqref{eq:mean_pred} and \eqref{eq:var_pred} yields the desired overapproximation of reachable sets formulated in \eqref{eq:standard_reachable_set}, as stated in Theorem \ref{theo:overapproximation_tube}.

\begin{theorem}[Overapproximation of robust reachable sets]
\label{theo:overapproximation_tube}
Under Assumptions \ref{assumption:consistent_estimator}, \ref{assumption:unbiased_estimator}, and  \ref{assumption:lipschitz_bounds}, the predictive tube $\tube_{n|t}$, propagated according to \eqref{eq:mean_pred} and \eqref{eq:var_pred}, overapproximates the robust reachable sets of the environment with dynamics \eqref{eq:env_dyn} such that
\begin{equation}
\begin{aligned}
    &\prs_{\predid|\globid}(\state_\globid, \predact_{0|\globid}, \dots, \predact_{\predid-1|\globid}) \subseteq \tube_{\predid|\globid}\\
    &\text{with }\tube_{\predid|\globid} := \left\{ \state \in \statesp \mid (\state-\predmean_{\predid|\globid})^\top (\predvar_{\predid|\globid})^{-1} (\state-\predmean_{\predid|\globid}) \leq 1\right\} = \predmean_{\predid|\globid} \oplus \predchol_{\predid|\globid}^\top \unitball_{\statesp} \label{eq:pred_tube}
\end{aligned}
\end{equation}
and the Cholesky decomposition $\predchol_{\predid|\globid}^\top \predchol_{\predid|\globid} = \predvar_{\predid|\globid}$ and the unit ball $\unitball_{\statesp} = \{ \unitvec \in \mathbb{R}^{\dimS} \mid \| \unitvec \|_2 \leq 1 \}$.
\begin{proof}
    See Theorem \ref{theo:app_overapproximation_tube} of Appendix \ref{app:proofs}.
\end{proof}
\end{theorem}

\subsection{Predictive Safety Filter Formulation}
\label{subsec:mpc_setting}
Since the reachable set construction in Theorem \ref{theo:overapproximation_tube} relies on Assumptions \ref{assumption:consistent_estimator} and \ref{assumption:unbiased_estimator} that are only valid within the certain set $\accset^\epiid$, we need to enforce certainty along predicted trajectories. That is, we require that all pairs $(\state_{\predid|\globid}, v_{\predid|\globid}) \in \accset^\epiid$ for all $\state_{\predid|\globid} \in \tube_{\predid|\globid}$ and $v_{\predid|\globid} = \predact_{\predid|\globid} + \contgain_{\predid|\globid}(\state_{\predid|\globid} - \predmean_{\predid|\globid})$. Formulating the definition of $\accset^\epiid$, i.e., $\frac{1}{\dimS} \mathrm{tr}(\kalgain(\state, \act)) \geq \xi$, as a general nonlinear constraint \citep{kohler_computationally_2021} allows us to maintain certainty throughout the reachable set by enforcing 
\begin{equation} 
\label{eq:certainty_const}
(\predmean_{\predid|\globid}, \predact_{\predid|\globid}) \oplus g_{\predid \mid \globid} \unitball_{\statesp\times \actsp} \subseteq \accset^\epiid \text{ with } g_{\predid \mid \globid} =\lipschitzmod \sqrt{\left(\lambda_{\mathrm{max}}(  \predvar_{\predid \mid \globid} \right)  +\lambda_{\mathrm{max}}( \contgain_{\predid \mid \globid}  \predvar_{\predid \mid \globid} \contgain_{\predid \mid \globid}^\top )) }
\end{equation}
and $\unitball_{\statesp \times \actsp } = \{ \unitvec \in \mathbb{R}^{\dimS + n_{\actsp}} \mid \| \unitvec \|_2 \leq 1 \}$, as is shown in Lemma \ref{lemma_app:uncertainty_const_tightening} of Appendix \ref{app:proofs}.
Integrating the reachable set from \eqref{eq:mean_pred} and \eqref{eq:var_pred}, and the certainty constraint \eqref{eq:certainty_const} into \eqref{eq:mpc_problem_statement}, yields the \algo~problem:
\begin{subequations}\label{eq:upsi_psf}
\begin{align}
    \min_{\predact_{0|\globid}, \dots, \predact_{ \predhorizon-1|\globid}} & \|\act_\globid - \predact_{0|\globid} \|_2^2 \label{eq:mpc_cost}\\
    \text{s.t. } & \predmean_{0|\globid} = \state_\globid, \quad \predvar_{0|\globid} = 0 \label{eq:upsi_init}\\
    &\predmean_{\predid+1 \mid \globid} = \mlmeanmod(\predmean_{\predid \mid \globid}, \predact_{\predid \mid \globid}) \text{ for all }  \predid \in \{0, \dots, \predhorizon -1 \}\label{eq:upsi_mean_pred}\\
    & \predvar_{\predid+1 \mid \globid} =                 \Phi ( \predvar_{\predid \mid \globid},\contgain_{\predid \mid \globid},\predmean_{\predid \mid \globid}, \predact_{\predid \mid \globid}) \text{ for all }  \predid \in \{0, \dots, \predhorizon -1 \}, \text{ see \eqref{eq:var_pred}} \label{eq:upsi_var_pred}\\
    & \predmean_{\predid \mid \globid} \in \stateset \ominus \predchol_{\predid|\globid}^\top \unitball_{\statesp} \text{ for all }  \predid \in \{0, \dots, \predhorizon -1 \}\label{eq:upsi_state_const}\\
    & \predact_{\predid \mid \globid} \in \actset \ominus \contgain_{\predid \mid \globid} \predchol_{\predid|\globid}^\top \unitball_{\statesp}   \text{ for all }  \predid \in \{0, \dots, \predhorizon -1 \} \label{eq:upsi_act_const}\\
    & \predmean_{\predhorizon \mid \globid} \in \termset^\epiid \ominus \predchol_{\predhorizon|\globid}^\top \unitball_{\statesp}\label{eq:upsi_term_const}\\
    & (\predmean_{\predid \mid \globid}, \predact_{\predid \mid \globid}) \in \accset^\epiid \ominus g_{\predid \mid \globid} \unitball_{\statesp \times \actsp} \text{ for all } \predid \in \{0, \dots, \predhorizon - 1 \} \label{eq:upsi_cert_const}
\end{align}
\end{subequations}
Here, \eqref{eq:upsi_init}-\eqref{eq:upsi_var_pred} describe the reachable set computation developed in Section \ref{subsec:reach_set}, \eqref{eq:upsi_state_const} ensures all $\state_{\predid|\globid} \in \tube_{\predid|\globid}$ comply with state constraints $\stateset$, \eqref{eq:upsi_act_const} ensures $v_{\predid|\globid} = \predact_{\predid|\globid} + \contgain_{\predid|\globid} (\state_{\predid|\globid} - \predmean_{\predid|\globid}) \in \actset$ for all $\state_{\predid|\globid} \in \tube_{\predid|\globid}$, and  \eqref{eq:upsi_term_const} ensures the final reachable set lies within the terminal set. Finally, constraint \eqref{eq:upsi_cert_const} ensures all pairs $(\state_{\predid|\globid}, v_{\predid|\globid}) \in \accset^\epiid$ for all $\state_{\predid|\globid} \in \tube_{\predid|\globid}$. 
This general formulation can be solved with different robust MPC schemes \citep{cannon2011robust, kohler_computationally_2021}. The construction of the reachable sets and the certainty constraint allows for rigorous safety guarantees via persistent feasibility. That is, the combination of \eqref{eq:mean_pred}, \eqref{eq:var_pred}, and \eqref{eq:certainty_const} allows us to prove that there always exists a constraint-satisfying solution. However, since the focus of this work lies on rigorous reachable set computation from PE models stated in Theorem \ref{theo:overapproximation_tube}, we do not further focus on this aspect but showcase one such approach in Section Supp. \ref{supp:persistent_feasibility}.

\subsection{Terminal set expansion}
\label{subsec:term_set}
To reduce conservatism of the \algo scheme, we expand the terminal set $\termset^0$.
Similar to \citet{rosolia2017learning}, we define the one-step controllable set of an arbitrary set $\mathbb{X} \subseteq \stateset$ as $\contrset_1(\mathbb{X}) := \pre(\mathbb{X}) \cap \stateset$ with $\pre(\mathbb{X}) := \{ \state \in \statesp \mid \exists \act \in \actset \text{ such that } \prs(\state, \act) \subseteq \mathbb{X} \text{ and } (s,a) \in \accset^\epiid \}$, i.e.,
all state constraint satisfying states from which $\mathbb{X}$ can be reached without action constraint violation while remaining certain.
Consequently, the $\predid$-step controllable set of $\mathbb{X}$ is defined as $\contrset_\predid(\mathbb{X}) := \pre(\contrset_{\predid-1}(\mathbb{X})) \cap \stateset$ for $\predid \in \{0, \dots, \predhorizon \}$ with $\contrset_0(\mathbb{X}) = \mathbb{X}$.
In each episode $\epiid$, we expand the terminal set with its $\predhorizon$-controllable set, such that $\termset^{\epiid+1} = \bigcup_{\predid=0}^{\predhorizon} \contrset_\predid(\termset^\epiid)$ through the sampled expansion\looseness=-1
\begin{equation}
\label{eq:termset_expansion}
    \termset^{\epiid+1} = \termset^{\epiid} \cup \Big\{ \bigcup_{\globid \in \mathcal{I}^\epiid} \bigcup_{\predid=0}^\predhorizon \predmean_{\predid|\globid} \oplus \predchol_{\predid|\globid}^\top\unitball_\statesp \Big\}
\end{equation}
with $\mathcal{I}^\epiid := \{ \globid \in \{1, \dots, T^\epiid\} \text{ where } \eqref{eq:upsi_state_const}, \eqref{eq:upsi_act_const}, \eqref{eq:upsi_term_const} \text{, and }\eqref{eq:upsi_cert_const} \text{ hold} \}$. Feasible paths at episode $\epiid$ are also feasible paths at $\epiid+1$ by Assumption \ref{assumption:improve}, making \eqref{eq:termset_expansion} a valid expansion scheme.
The expansion \eqref{eq:termset_expansion} faces the issue that there exists no closed form of $\termset^\epiid$ that can be easily optimized over when solving \eqref{eq:upsi_psf}. In practice, we therefore approximate \eqref{eq:termset_expansion} via a sampling-based scheme inspired by \citet{rosolia2017learning}, yielding a convex approximation of $\termset^\epiid$  as described in Section Supp. \ref{supp:termset_expansion}. This measure trades theoretical rigor for a less restrictive exploration behavior.

\section{Empirical Evaluation}
\label{sec:experiments}
Below, we evaluate the reachable set computation of \algo in Section \ref{subsec:toy_example} and benchmark its filtering behavior with MBPO \citep{janner_when_2019-1} using practical simplifications in Section \ref{subsec:mbrl_integration}. 
We report a consistent overapproximation of the true reachable sets and a substantial reduction in constraint violations of a prior PE-based PSF \citep{gronauer_reinforcement_2024} with performance on par with MBPO.

\subsection{Reachable Set Computation} 
\label{subsec:toy_example}
\begin{figure}[htbp]
    \centering
    \vspace{-0.5em}
    \begin{subfigure}[c]{0.285\textwidth}
        \centering        
        \includegraphics[width=\textwidth]{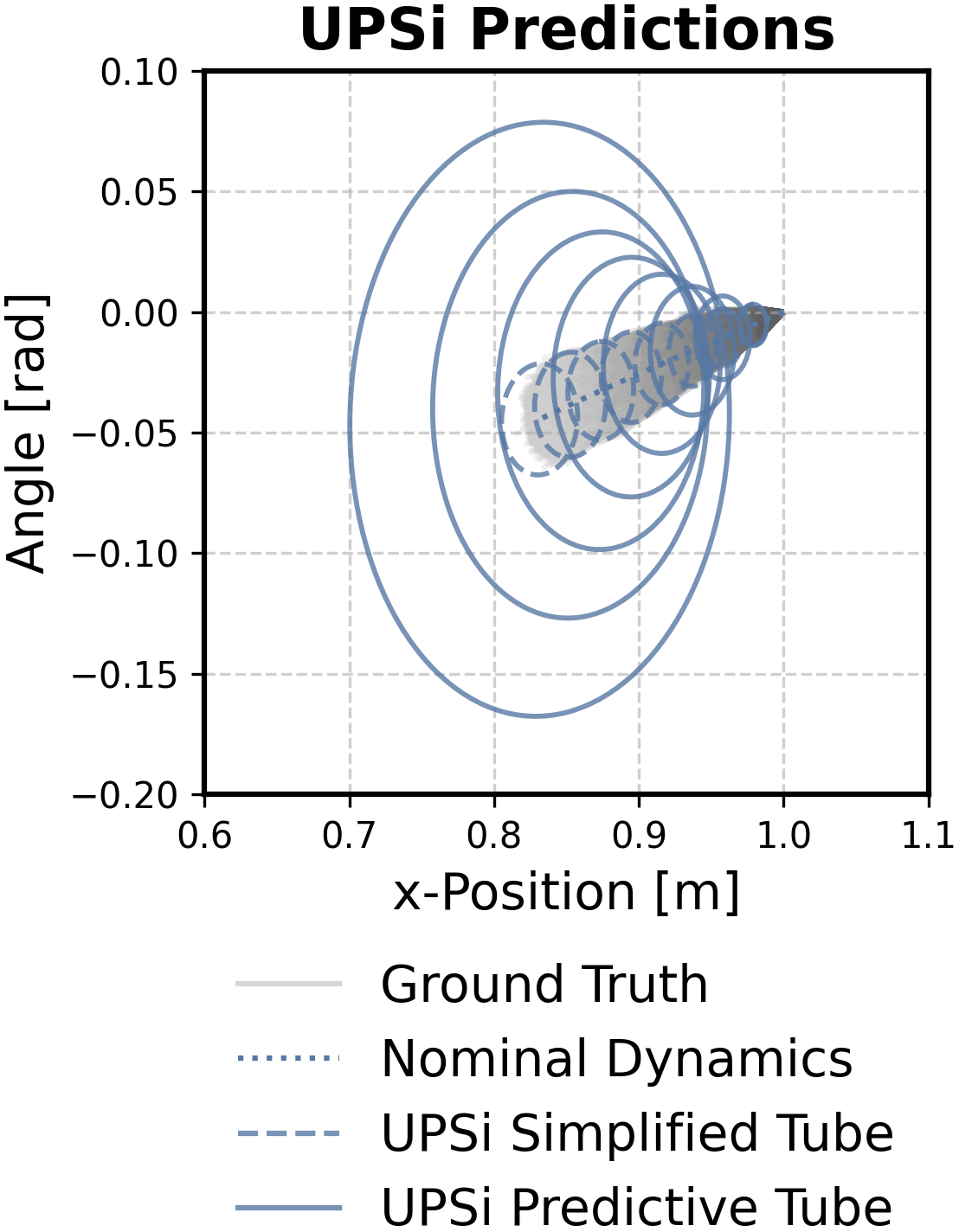}
        \caption{\algo over ground truth.}
        \label{fig:mpc_sequence}
    \end{subfigure}
    \hfill
    \begin{subfigure}[c]{0.7\textwidth}
        \centering
        \includegraphics[width=\textwidth]{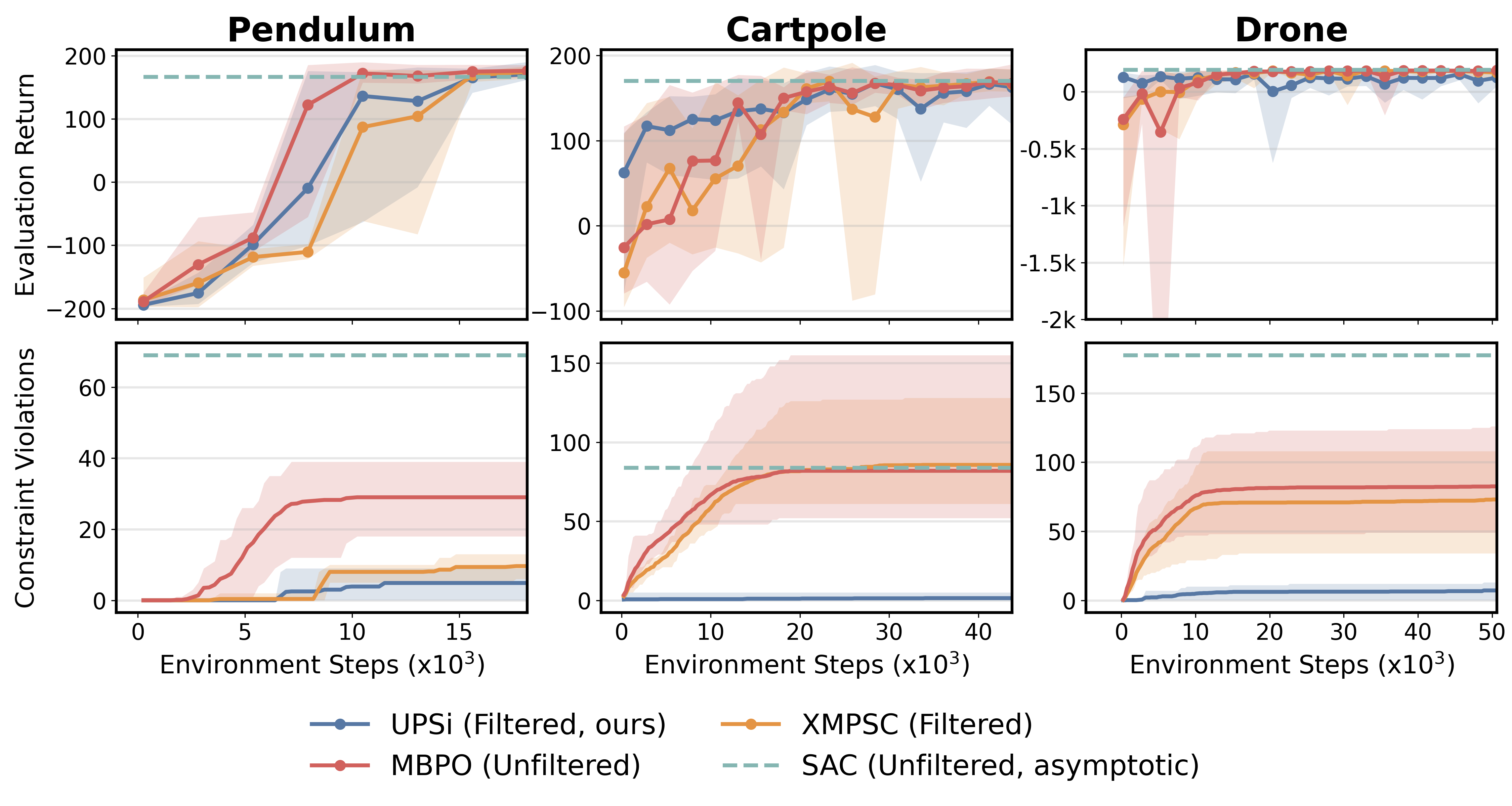}
        \caption{Evaluation return and accumulated failures over environment steps; solid lines show the mean and shaded regions the min–max range across $8$ seeds.}
        \label{fig:experiments}
    \end{subfigure}
    \vspace{-1em}
\caption{Experimental results: \emph{(a) \algo~yields a valid overapproximation of reachable sets.\\ (b) \algo~with practical simplifications yields a substantial reduction in constraint violations.}}
    \label{fig:results}
    \vspace{-1em}
\end{figure}

We evaluate the overapproximation of reachable sets propagated according to \eqref{eq:mean_pred} and \eqref{eq:var_pred} as stated in Theorem \ref{theo:overapproximation_tube} on Cartpole ($\statesp\subseteq\mathbb{R}^4, \actsp\subseteq\mathbb{R}^1$) with an ancillary controller $\contgain$, where we assume $\ellscale$, $\lipschitz$, and $\lipschitzvarchol$ to be known and the model to be evaluated within $\accset$. Further details are provided in Section Supp.~\ref{supp:reachable_sets}. Figure \ref{fig:mpc_sequence} shows the predicted sets $\predmean_{\predid|\globid} \oplus \predchol_{\predid|\globid}^\top \unitball_\statesp$ in blue, as well as an approximation of the corresponding reachable sets $\prs_{\predid|\globid}$, obtained by simulating the system 100,000 times with trajectories depicted in gray. We observe that $\prs_{\predid|\globid} \subseteq \predmean_{\predid|\globid} \oplus \predchol_{\predid|\globid}^\top \unitball_\statesp$ consistently holds, verifying Theorem \ref{theo:overapproximation_tube}. Since $\ellscale$, $\lipschitz$, and $\lipschitzvarchol$ are frequently unknown in practice, we further present a simplified naive tube prediction that propagates uncertainty as a Gaussian random variable $\tilde{\prepredvar}_{\predid +1 \mid \globid} = \cloopmat_{\predid \mid \globid} \tilde{\prepredvar}_{\predid \mid \globid}\cloopmat_{\predid \mid \globid}^\top + \mlvarmod(\predmean_{\predid \mid \globid}, \predact_{\predid \mid \globid})$ and obtains a tube shape matrix by clipping its support at a certain probability level. While this approach lacks theoretical rigor, it empirically approximates the reachable sets of the environment $\prs_{\predid|\globid}$ sufficiently well.

\subsection{Integration of \algo with MBRL}
\label{subsec:mbrl_integration}
We integrate \algo into MBPO\footnote{Code and videos can be accessed online: \hyperlink{https://github.com/Data-Science-in-Mechanical-Engineering/upsi}{https://github.com/Data-Science-in-Mechanical-Engineering/upsi}.}, making several practical simplifications. 
In particular, we leverage the simplified naive propagation scheme introduced in Section \ref{subsec:toy_example}, assume $\lipschitzmod=0$, and employ no ancillary controller, i.e.,\ $\contgain=0$.
While these simplifications break the formal guarantees of Section~\ref{sec:method}, they demonstrate the practical effectiveness of \algo with minimal domain knowledge. Resulting infeasibilities are handled via the backup control scheme presented by \cite{wabersich_linear_2018} and soft constraints on \eqref{eq:upsi_psf}. Implementation details are provided in Section Supp.~\ref{supp:upsi_implementation}.

We evaluate on three environments from~\cite{gronauer_reinforcement_2024}: Pendulum ($\statesp\subseteq\mathbb{R}^2, \actsp\subseteq\mathbb{R}^1$), Cartpole ($\statesp\subseteq\mathbb{R}^4, \actsp\subseteq\mathbb{R}^1$), and Drone ($\statesp\subseteq\mathbb{R}^{12}, \actsp\subseteq\mathbb{R}^4$). We compare against MBPO with XMPSC\footnote{We do not compare \algo to~\citet{guzelkaya_ensemble_2024} due to similarities with XMPSC and the absence of published code.}~\citep{gronauer_reinforcement_2024} and the baselines MBPO~\citep{janner_when_2019-1} and SAC~\citep{haarnoja_soft_2018}, with a detailed description of the experimental setup provided in Section Supp. \ref{supp:experimental_setup}.
As shown in Figure \ref{fig:experiments}, \algo achieves control returns on par with all baselines while reducing constraint violations during exploration compared to XMPSC. The improvement is more pronounced in Cartpole and Drone, likely because XMPSC lacks a mechanism to prevent model exploitation, which becomes more problematic in higher-dimensional spaces. Additional results in Section Supp.~\ref{supp:additional_results} show that \algo~both filters and finds feasible solutions of the PSF problem more often than XMPSC.

\section{Conclusion}

Exploration safety is a fundamental challenge in RL that can be efficiently addressed via PSFs. However, standard PSF methods rely on common MPC models that either require substantial domain knowledge or scale poorly with data. In contrast, PE models have proven their capability to model high-dimensional systems with nonlinear dynamics, problems in which deep RL methods excel. \algo~addresses the gap between these fields. We present a novel construction of robust reachable sets from PE models that seamlessly integrates into robust PSF formulations and introduce model certainty as a PSF constraint to avoid model exploitation.
Further, we empirically validate the reachable set overapproximation and report the effectiveness of a formulation of \algo under practical simplifications when integrated into MBRL.
These simplifications, together with the slow convergence of the PSF problem, are the main limitations of the presented work. Future work should address the limitations through improvements toward real-time feasibility, which should enable to target more challenging control problems and hardware deployment.

\subsection*{Acknowledgements}
We thank Manuel Dahmen, Artur Eisele, Johannes Berger, Devdutt Subhasish, Alexander Gräfe, and Sven Gronauer for their support of this work. This work was supported in part by the Helmholtz School for Data Science in Life, Earth and Energy (HDS-LEE), the Helmholtz Association of German Research Centres, and the German Research Foundation (DFG): RTG 2236/2 (UnRAVeL). The authors acknowledge the computing time provided at the NHR Center NHR4CES at RWTH Aachen University (project numbers p0022301,~p0021919).

\appendix
\section{Proofs}
\label{app:proofs}
Extended versions of all lemmas and theorems are provided in Section~\ref{supp:proofs} of the supplementary material.

\begin{lemma}[Linearization error bound on nominal dynamics]
    \label{lemma_app:lin_err_bound}
    Let Assumption~\ref{assumption:lipschitz_bounds} hold.
    The linearization error $e_{\predid \mid \globid}$ for the propagation of predictive shapes for all $P\succ0$ is upper-bounded by $\|e_{\predid \mid \globid}\|_2 \leq e^{\mathrm{max}}_{\predid \mid \globid} = \tfrac{\lipschitz}{2} (  \sqrt{ \lambda_\mathrm{max}(\predvar)} + \sqrt{ \lambda_{\mathrm{max}}( \contgain  \predvar \contgain^\top )} )^2.$
\begin{proof}
    The linearization yields a first-order Taylor approximation of the nominal dynamics
    \begin{equation}
    \begin{aligned}
        \meanmod(\state, \contgain(\state-\predmean) + \predact) = \meanmod(\predmean, \predact) + \sysmat(\state-\predmean) + \inmat\contgain(\state - \predmean) + e = \meanmod(\predmean, \predact) + \grad_\state \meanmod \diff + \grad_\act \meanmod \contgain \diff + e
       \end{aligned} 
    \end{equation}
    with deviation $\diff := (\state-\predmean)$ and the linearization error $e$ depending on the Hessian $\mathcal{H}$ with the $i$-th component $\mathcal{H}_i$
    \begin{equation}
        e = \tfrac{1}{2} \left[ \diff^\top (\contgain\diff)^\top \right] \hess \begin{bmatrix} \diff \\ \contgain\diff \end{bmatrix} \text{ with }
        \hess_i = \begin{bmatrix}
            \grad_\state \grad_\state \meanmod_i  & \grad_\state \grad_\act \meanmod_i \\\grad_\act \grad_\state \meanmod_i & \grad_\act \grad_\act \meanmod_i
        \end{bmatrix}.
    \end{equation}
    Thus, the $i$-th component of $e$ is
    $
            e_i = \tfrac{1}{2} \left( \diff^\top \grad_\state \grad_\state \meanmod_i \diff + 2 \diff^\top \grad_\state \grad_\act \meanmod_i \contgain \diff +  (\contgain \diff)^\top \grad_\act \grad_\act \meanmod_i \contgain \diff \right).
        $
     Combining the output dimensions $i$, we have a linearization error
    $
            e = \left[ e_1, e_2, \dots, e_{\dimS} \right]^\top
            = \tfrac{1}{2} \left( \diff^\top \grad_\state \grad_\state \meanmod \diff + 2 \diff^\top \grad_\state \grad_\act \meanmod \contgain \diff +  (\contgain \diff)^\top \grad_\act \grad_\act \meanmod \contgain \diff \right).
       $
    Using Assumption~\ref{assumption:lipschitz_bounds} allows us to formulate an upper bound on the L2 norm of the linearization error
    \begin{equation}
    \begin{aligned}
    \label{eq:app_lin_err_bound}
        \|e\|_2 &= \tfrac{1}{2} \| \diff^\top \grad_\state \grad_\state \meanmod \diff + 2 \diff^\top \grad_\state \grad_\act \meanmod \contgain \diff +  (\contgain \diff)^\top \grad_\act \grad_\act \meanmod \contgain \diff \|_2\\
        & \leq \tfrac{\lipschitz}{2} \left( \| \diff \|_2^2 + 2 \| \diff \|_2 \| \contgain \diff \|_2 +  \|\contgain\diff \|_2^2\right).
    \end{aligned}
    \end{equation}

    We  upper bound the components starting with $\| \diff \|_2^2 = \| \state - \predmean \|_2^2$. Since $\state\in\mathcal{P}$, the deviation satisfies $\diff^\top (\predvar)^{-1}\diff \le 1$. Further, since $\predvar$ is a symmetric positive definite shape matrix, there exist matrices $Q$ and $\Lambda$ such that
    $\predvar = Q \Lambda Q^\top$  with  $Q^\top Q = I \text{ and } \Lambda = \mathrm{diag}(\lambda_1(\predvar), \dots, \lambda_{\dimS}(\predvar)) $
    with $\lambda_i(\predvar)$ being the $i$-th eigenvalue of $\predvar$. Thus, we have
    \begin{equation}
    \begin{aligned}
        \diff^\top  (\predvar)^{-1} \diff = \diff^\top  (Q \Lambda Q^\top)^{-1} \diff
        = y^\top \Lambda^{-1} y = \sum_{i=1}^{\dimS} \tfrac{y_i^2}{\lambda_i(\predvar)} \leq 1 \text{ with } y = Q^\top \diff. \\
            1 \geq \sum_{i=1}^{\dimS} \tfrac{y_i^2}{\lambda_i(\predvar)} \geq \tfrac{1}{\lambda_\mathrm{max}(\predvar)} \sum_{i=1}^{\dimS} y_i^2 = \tfrac{\|y\|_2^2}{\lambda_\mathrm{max}(\predvar)} = \tfrac{(Q^\top \diff)^\top (Q^\top \diff)}{\lambda_\mathrm{max}(\predvar)} = \tfrac{\|\diff\|_2^2}{\lambda_\mathrm{max}(\predvar)} 
        \end{aligned}
    \end{equation}
    yields the upper bound $\lambda_\mathrm{max}(\predvar) \geq \|\diff\|_2^2$.
    Next, we compute an upper bound for $\|\contgain \diff \|_2^2$. Since $\predvar$ is symmetric positive definite, there exists the square root $(\predvar)^{-\tfrac{1}{2}}$. We define $\Tilde{\diff}:=(\predvar)^{-\tfrac{1}{2}}\diff$
    \begin{equation}
    \begin{aligned}
        &\| \contgain \diff\|_2^2 = \| \contgain (\predvar)^{\tfrac{1}{2}} \Tilde{\diff}\|_2^2 = (\contgain (\predvar)^{\tfrac{1}{2}} \Tilde{\diff})^\top (\contgain (\predvar)^{\tfrac{1}{2}} \Tilde{\diff}) = \Tilde{\diff}^\top (\predvar)^{\tfrac{1}{2}} \contgain^\top \contgain (\predvar)^{\tfrac{1}{2}}  \Tilde{\diff}\\
        & \leq \lambda_{\mathrm{max}}\left( ( \predvar)^{\tfrac{1}{2}} \contgain^\top \contgain ( \predvar)^{\tfrac{1}{2}}\right)\| \Tilde\diff\|_2^2 
        \leq \lambda_{\mathrm{max}}\left( (\predvar)^{\tfrac{1}{2}} \contgain^\top \contgain (\predvar)^{\tfrac{1}{2}} \right) 
         = \lambda_{\mathrm{max}}\left( \contgain  \predvar \contgain^\top \right).
    \end{aligned}
    \end{equation}
    Since both $\predvar$ and $\contgain \predvar \contgain^\top$ are symmetric positive definite, we can further bound $\|\diff\|_2 \leq \sqrt{\lambda_\mathrm{max}(\predvar)}$ and $\| \contgain \diff \|_2 \leq \sqrt{\lambda_{\mathrm{max}}\left( \contgain  \predvar \contgain^\top \right)}$. Substituting in \eqref{eq:app_lin_err_bound} yields the desired result.
\end{proof}
\end{lemma}

\begin{lemma}[Linearization error for aleatoric variance]
\label{lemma_app:lin_err_var}
Let Assumption~\ref{assumption:lipschitz_bounds} hold. 
The state-space error induced by approximating the noise injection
$\cholenv(\state,v)$ by $\cholenv(\predmean,\predact)$, that is, $e_{\varenv} := \Big(\cholenv(\state,v)-\cholenv(\predmean,\predact)\Big)w$ with $w\in\aleaset,\ \|w\|_2^2\le \ellscale$, is bounded by $\|e_{\varenv}\|_2
\le
\lipschitzvarchol\sqrt{\ellscale}\sqrt{\lambda_{\max}(\predvar)+\lambda_{\max}(\contgain\predvar\contgain^\top)}$.

\begin{proof}
By definition, we have $\|e_{\varenv}\|_2 \le \|\cholenv(\state,v)-\cholenv(\predmean,\predact)\|\,\|w\|_2.$
Since $w\in\aleaset$, we also have $\|w\|_2\le \sqrt{\ellscale}$.
Moreover, by Assumption~\ref{assumption:lipschitz_bounds}, it holds $\|\cholenv(\state,v)-\cholenv(\predmean,\predact)\|
    \le
    \lipschitzvarchol \|(\state,v)-(\predmean,\predact)\|_2
    =
    \lipschitzvarchol \sqrt{\|\diff\|_2^2+\|\contgain\diff\|_2^2}.$
Combining these two yields
\begin{equation}
\label{eqn:912d94601}
    \|e_{\varenv}\|_2
    \le
    \lipschitzvarchol\sqrt{\ellscale}\,\sqrt{\|\diff\|_2^2+\|\contgain\diff\|_2^2}.
\end{equation}
It remains to bound $\|\diff\|_2^2$ and $\|\contgain\diff\|_2^2$ for $\state\in\mathcal P$.
From $\state\in\mathcal P$ we have $\diff^\top(\predvar)^{-1}\diff \le 1$,
which gives $\|\diff\|_2^2 \le\lambda_{\max}(\predvar)$.
Similarly, $\|\contgain\diff\|_2^2
    =
    \diff^\top \contgain^\top\contgain \diff
    \le\lambda_{\max}(\contgain\predvar\contgain^\top)$.
Substituting into \eqref{eqn:912d94601} gives the bound.
\end{proof}
\end{lemma}

\begin{lemma}[Total linearization error]
    \label{lemma_app:total_lin_err}
    Let Assumption~\ref{assumption:lipschitz_bounds} hold. The overall linearization error $e$ of the \algo set propagation is upper bounded by
    $\|e\|_2 \leq e^{\mathrm{max}} = \tfrac{\lipschitz}{2} (  \sqrt{ \lambda_\mathrm{max}(\predvar)} + \sqrt{ \lambda_{\mathrm{max}}( \contgain  \predvar \contgain^\top )} )^2 + \lipschitzvarchol \sqrt{\ellscale}\sqrt{ \lambda_{\mathrm{max}}( \predvar ) + \lambda_{\mathrm{max}}( \contgain  \predvar \contgain^\top )}.$
    \begin{proof}
        The overall linearization error comprises the linearization error of the nominal dynamics and the process noise variance $e = e_{\grad\meanenv} + e_{\varenv}$. Using Lemmas \ref{lemma_app:lin_err_bound} and \ref{lemma_app:lin_err_var}, we can define the upper bound on linearization error norm $\| e \|_2 = \| e_{\grad\meanenv} + e_{\varenv} \|_2 \leq \| e_{\grad\meanenv} \|_2 + \| e_{\varenv} \|_2
             \leq \tfrac{\lipschitz}{2} (  \sqrt{ \lambda_\mathrm{max}(\predvar)} + \sqrt{ \lambda_{\mathrm{max}}( \contgain  \predvar \contgain^\top )} )^2 + \lipschitzvarchol \sqrt{\ellscale} \sqrt{ \lambda_{\mathrm{max}}( \predvar ) + \lambda_{\mathrm{max}}( \contgain  \predvar \contgain^\top )}
             = : e^{\mathrm{max}}$ 
        that yields the desired result.
    \end{proof}
\end{lemma}

\begin{lemma}[Outer ellipsoid of a Minkowski sum]
    \label{lemma_app:outer_ellipsoid}
    Let $\mathcal{E}(A):=\{x\mid x^\top A^{-1}x\le1\}$, as well as $\mathcal{E}(B):=\{x\mid x^\top B^{-1}x\le1\}$ be two ellipsoids with $A,B\succ0$. Then a valid outer approximation of the Minkowski sum $\mathcal{E}(A)\oplus\mathcal{E}(B)\subseteq\mathcal{E}(C)$ is achieved with
    \begin{equation}
    \label{eq:app_c_min_sum}
        C = \left(1+\tfrac{1}{\beta}\right)A + \left(1+\beta\right)B,
    \end{equation}
    for every $\beta>0$. Furthermore, the minimal valid outer approximation of the Minkowski sum has
    \begin{equation}
        \beta^* = \sqrt{\tfrac{\mathrm{tr}(A)}{\mathrm{tr}(B)}}.
    \end{equation}
    For cases with $A = 0$, we adopt $\mathcal{E}(A) = \{0\}$, so that \eqref{eq:app_c_min_sum} reduces to $C = B$ (similarly for $B = 0$).
    \begin{proof}
    To be a valid overapproximation, we need to find $C$ such that the support of $\mathcal{E}(A)\oplus\mathcal{E}(B)$ lies within the support of $\mathcal{E}(C)$ for any arbitrary direction vector $v$. That is
    \begin{equation}
        \begin{aligned}
            h_{\mathcal{E}(C)}(v) \overset{!}{\geq} h_{\mathcal{E}(A)\oplus\mathcal{E}(B)}(v) = h_{\mathcal{E}(A)}(v) + h_{\mathcal{E}(B)}(v) \; &\Rightarrow \;
            \sqrt{v^\top C\, v} \overset{!}{\geq} \sqrt{v^\top A v} + \sqrt{v^\top B v} \\
            v^\top C\, v &\overset{!}{\geq} \left( \sqrt{v^\top A v} + \sqrt{v^\top B v} \right)^2.
        \end{aligned}
    \end{equation}
    We use the outer approximation from \cite[Chp. 2.2 or 2.4]{kurzhanski1997ellipsoidal}, similarly used in~\cite{halder_parameterized_2018}: $ ( \sqrt{v^\top A v} + \sqrt{v^\top B v} )^2 \leq (1+\tfrac{1}{\beta}) v^\top A v + (1+\beta ) v^\top B v $ for $\beta > 0$. Consequently, for any $\beta > 0$
    \begin{equation}
    \label{eq:app_beta_approx}
        C = \left(1+\tfrac{1}{\beta}\right)A + \left(1+\beta\right)B
    \end{equation}
    yields a valid overapproximation of $\mathcal{E}(A)\oplus\mathcal{E}(B)$. To find the minimal valid overapproximation, we determine $\beta$ to minimize $\mathrm{tr}(C) = \left(1+\tfrac{1}{\beta}\right)\mathrm{tr}(A) + \left(1+\beta\right)\mathrm{tr}(B)$. Thus, we get
    \begin{equation}
    \label{eq:app_opt_beta}
        \grad_\beta \left[ \left(1+\tfrac{1}{\beta}\right)\mathrm{tr}(A) + \left(1+\beta\right)\mathrm{tr}(B) \right] \overset{!}{=} 0, \; \Rightarrow \;
        -\tfrac{1}{\beta^2}\mathrm{tr}(A) + \mathrm{tr}(B) = 0,
    \end{equation}
    which yields $\beta^* = \sqrt{\tfrac{\mathrm{tr}(A)}{\mathrm{tr}(B)}}$. Inserting into \eqref{eq:app_beta_approx} yields the desired result.
    \end{proof}
\end{lemma}

\begin{lemma}[Outer approximation of process noise]
    \label{lemma_app:gauss_lvl_set}
    Fix $\globid$ and a nominal pair $(\predmean_{n|t},\predact_{n|t})$, and let Assumption~\ref{assumption:consistent_estimator} hold.
    For the process noise $L_{n|t}\alea$ with $L_{n|t}:=\cholenv(\predmean_{n|t},\predact_{n|t})$,
    $\|\alea\|_2^2\le\ellscale$, and $L_{n|t} L_{n|t}^\top=\varenv(\predmean_{n|t},\predact_{n|t})$, it holds that
    \begin{equation}
        \label{eq:app_noise_ellips}
        L_{n|t}\alea
        \in
        \Big\{ x \ \big| \ x^\top\big(\ellscale\,\mlvarmod(\predmean_{n|t},\predact_{n|t})\big)^{-1}x \le 1\Big\}.
    \end{equation}
    \begin{proof}
    The Gaussian aleatoric covariance estimate $\mlvarmod$ (see Section~\ref{subsec:cert_set}) can be used to outer approximate $L_{n|t}\alea$. Since $\|\alea\|_2^2\le\ellscale$, the noise is bounded as
        $L_{n|t}\alea\in\{x\mid x^\top(\ellscale\,\varenv(\predmean_{n|t},\predact_{n|t}))^{-1}x\le1\}$ with $\varenv(\predmean_{n|t},\predact_{n|t})=L_{n|t}L_{n|t}^\top$.
        By Assumption~\ref{assumption:consistent_estimator} $\mlvarmod(\predmean_{n|t},\predact_{n|t})\succcurlyeq\varenv(\predmean_{n|t},\predact_{n|t})$, so the above ellipsoid is contained in the one with shape matrix $\ellscale\,\mlvarmod(\predmean_{n|t},\predact_{n|t})$, which yields~\eqref{eq:app_noise_ellips}.
    \end{proof}
\end{lemma}

\begin{lemma}[Reformulation of an ellipsoid as a Minkowski sum]
    \label{lemma_app:ellipsoid_to_minkowski}
    Let $\predvar$ be a symmetric positive definite matrix with Cholesky decomposition 
    $\predvar = \predchol^\top \predchol$. Then the ellipsoidal set
    \begin{equation}
    \label{eq:app_ellipsoid}
        \mathcal{P} = \left\{ \state \in \statesp \mid (\state - \predmean)^\top (\predvar)^{-1} (\state - \predmean) \leq 1 \right\}
    \end{equation}
    is equivalent to the set $\mathcal{Q} = \left\{ \predmean \oplus \predchol^\top q \mid q \in \mathbb{R}^{\dimS},\, \| q \|_2 \leq 1 \right\}.$
    \begin{proof}
    We show set equivalence by proving both inclusions $\mathcal{Q} \subseteq \mathcal{P}$ 
    and $\mathcal{P} \subseteq \mathcal{Q}$. By the definition of the Minkowski sum, $\mathcal{Q}$ can be written as
    \begin{equation}
    \label{eq:app_reform_mink}
        \mathcal{Q} = \left\{ \predmean \oplus \predchol^\top q \mid \|q\|_2 \leq 1 \right\} 
        = \left\{ \predmean + \predchol^\top q \mid \|q\|_2 \leq 1 \right\},
    \end{equation}
    i.e., $\mathcal{Q}$ consists of all points of the form $\state = \predmean + \predchol^\top q$ 
    where $q$ lies in the unit ball.

    Part 1: $\mathcal{Q} \subseteq \mathcal{P}$.
    Let $\state \in \mathcal{Q}$ be arbitrary. By \eqref{eq:app_reform_mink}, there exists a 
    $q \in \mathbb{R}^{\dimS}$ with $\|q\|_2 \leq 1$ such that $\state = \predmean + \predchol^\top q$.
    Substituting this expression for $\state$ into the quadratic form in \eqref{eq:app_ellipsoid} 
    and using the Cholesky decomposition $\predvar = \predchol^\top \predchol$ yields $(\state - \predmean)^\top (\predvar)^{-1} (\state - \predmean) 
            = (\predchol^\top q)^\top (\predvar)^{-1} (\predchol^\top q) 
            = (\predchol^\top q)^\top (\predchol^\top \predchol)^{-1} (\predchol^\top q) 
            = q^\top q = \|q\|_2^2.$
    Since $\|q\|_2 \leq 1$ by assumption, it follows $\|q\|_2^2 \leq 1$, and therefore $(\state - \predmean)^\top (\predvar)^{-1} (\state - \predmean) = \|q\|_2^2 \leq 1,$
    which means $\state \in \mathcal{P}$. Since $\state \in \mathcal{Q}$ was arbitrary, 
    we have $\mathcal{Q} \subseteq \mathcal{P}$.

    Part 2: $\mathcal{P} \subseteq \mathcal{Q}$.
    Let $\state \in \mathcal{P}$ be arbitrary, so that 
    $(\state - \predmean)^\top (\predvar)^{-1} (\state - \predmean) \leq 1$.
    Since $\predvar$ is symmetric positive definite, its Cholesky factor $\predchol$ 
    is invertible. We can therefore define $q := (\predchol^\top)^{-1}(\state - \predmean),$
    which by construction satisfies $\state = \predmean + \predchol^\top q$, i.e., $\state$ is of 
    the form required by \eqref{eq:app_reform_mink}. It remains to show that $\|q\|_2 \leq 1$.
    Expanding $\|q\|_2^2$ and using the Cholesky decomposition gives $\|q\|_2^2 = q^\top q 
            = ((\predchol^\top)^{-1}(\state-\predmean))^\top ((\predchol^\top)^{-1}(\state-\predmean)) 
            = (\state - \predmean)^\top (\predchol^\top \predchol)^{-1} (\state - \predmean) 
            = (\state - \predmean)^\top (\predvar)^{-1} (\state - \predmean) \leq 1,$
    where we used $((\predchol^\top)^{-1})^\top = \predchol^{-1}$ in the third line,
    and the last inequality holds since $\state \in \mathcal{P}$. Therefore $\|q\|_2^2 \leq 1$, 
    which implies $\|q\|_2 \leq 1$. Since $\state = \predmean + \predchol^\top q$ with $\|q\|_2 \leq 1$, 
    we conclude $\state \in \mathcal{Q}$. Since $\state \in \mathcal{P}$ was arbitrary, 
    we have $\mathcal{P} \subseteq \mathcal{Q}$.
    
    Since $\mathcal{Q} \subseteq \mathcal{P}$ and $\mathcal{P} \subseteq \mathcal{Q}$, 
    we conclude $\mathcal{P} = \mathcal{Q}$, which yields the desired result.
    \end{proof}
\end{lemma}

\begin{lemma}[One-step ellipsoidal propagation]
\label{lemma_app:onestep_ellipsoid}
Fix $\globid$ and a nominal pair $(\predmean_{n|t},\predact_{n|t})$.
Let Assumptions~\ref{assumption:consistent_estimator},~\ref{assumption:unbiased_estimator}, and~\ref{assumption:lipschitz_bounds} hold.
Let $\predvar_{n+1|t}$ be obtained by repeated outer-ellipsoid inflation following Lemma~\ref{lemma_app:outer_ellipsoid}.
Define the tube
\begin{equation}
\label{eq:onestep_tube_np1}
    \tube_{n+1|t} :=
    \Big\{ s\in\statesp \ \Big|\ (s-\predmean_{n+1|t})^\top (\predvar_{n+1|t})^{-1}(s-\predmean_{n+1|t})\le 1\Big\}.
\end{equation}
Then for every $s\in\tube_{n|t}$ and every $w\in\aleaset$, the successor $s^+$
satisfies $s^+ \in \tube_{n+1|t}$.
\begin{proof}
Take any $s\in\tube_{n|t}$ and let $\diff=s-\predmean_{n|t}$.
By a first-order Taylor expansion of the nominal dynamics around $(\predmean_{n|t},\predact_{n|t})$ under the ancillary law
(and collecting higher-order terms), $\meanenv(s,v)=\meanenv(\predmean_{n|t},\predact_{n|t})+\cloopmat_{n|t}\diff+e_{\grad\meanenv}.$
Similarly, write the noise-injection term as nominal injection plus mismatch-induced state error, $\cholenv(s,v)\,w=\cholenv(\predmean_{n|t},\predact_{n|t})\,w+e_{\varenv}.$
Using the definition $s^+ = \meanenv(s,v) + \cholenv(s,v)\,w$ and $\predmean_{n+1|t}=\meanenv(\predmean_{n|t},\predact_{n|t})$ yields $s^+ - \predmean_{n+1|t}=\cloopmat_{n|t}\diff+\cholenv(\predmean_{n|t},\predact_{n|t})\,w+e,$ where $e:=e_{\grad\meanenv}+e_{\varenv}$.
Since $s\in\tube_{n|t}$, the deviation $\diff$ lies in the ellipsoid with shape matrix $\predvar_{n|t}$, hence
$\cloopmat_{n|t}\diff$ lies in the ellipsoid with shape matrix $\cloopmat_{n|t}\predvar_{n|t}\cloopmat_{n|t}^\top$.
Moreover, since $w\in\aleaset$, using Lemma~\ref{lemma_app:gauss_lvl_set}, the term $\cholenv(\predmean_{n|t},\predact_{n|t})\,w$ lies in the ellipsoid with
shape matrix $\ellscale\,\mlvarmod(\predmean_{n|t},\predact_{n|t})$.
Finally, by Lemma~\ref{lemma_app:total_lin_err} the linearization error satisfies $\|e\|_2\le e^{\mathrm{max}}_{n|t}$, i.e.,
$e\in\linerr_{n|t}:=\{e\in\mathbb R^{\dimS}\mid \|e\|_2\le e^{\mathrm{max}}_{n|t}\}$.
Therefore, the set of possible deviations $s^+-\predmean_{n+1|t}$ is contained in
\begin{equation}
\label{eq:onestep_mink_sum}
    \left\{x\,|\,x^\top(\cloopmat_{n|t}\predvar_{n|t}\cloopmat_{n|t}^\top)^{-1}x\le1\right\}
    \ \oplus\
    \left\{x\,|\,x^\top(\ellscale\,\mlvarmod(\predmean_{n|t},\predact_{n|t}))^{-1}x\le1\right\}
    \ \oplus\
    \linerr_{n|t}.
\end{equation}
We resolve~\eqref{eq:onestep_mink_sum} by applying the outer-ellipsoid inflation (Lemma~\ref{lemma_app:outer_ellipsoid}) twice.
Applied to the first part, it gives the ellipsoid with shape matrix
$Q_{n+1|t}=(1+\beta^{-1})\,\cloopmat_{n|t}\predvar_{n|t}\cloopmat_{n|t}^\top+(1+\beta)\,\ellscale\,\mlvarmod(\predmean_{n|t},\predact_{n|t})$, with $\beta \in \mathbb{R}$.
Applied once more with $\linerr_{n|t}$, we have
\begin{equation}
\label{eq:onestep_final}
    \predvar_{n+1|t}
    =
    (1+\tau^{-1})\left[(1+\beta^{-1})\,\cloopmat_{n|t}\predvar_{n|t}\cloopmat_{n|t}^\top
    +(1+\beta)\,\ellscale\,\mlvarmod(\predmean_{n|t},\predact_{n|t})\right]
    +(1+\tau)\,(e^{\mathrm{max}}_{n|t})^2 I
\end{equation}
for $\tau \in \mathbb{R}$ and $\beta,\tau>0$, which results in \eqref{eq:var_pred} for the minimizing $\beta,\tau$ from Lemma~\ref{lemma_app:outer_ellipsoid}. This yields scaling factors $\beta^\star = \sqrt{\mathrm{tr}(\cloopmat_{n|\globid}\predvar_{n|\globid}\cloopmat_{n|\globid}^\top)/\mathrm{tr}(\ellscale\mlvarmod(\predmean_{n|\globid},\predact_{n|\globid}))}$ and $\tau^\star = \sqrt{\mathrm{tr}(\prepredvar_{n+1|\globid})/(\dimS(e^{\max}_{n|\globid})^2)}$, where $\prepredvar_{n+1|\globid} = (1+\beta^{-1})\cloopmat_{n|\globid}\predvar_{n|\globid}\cloopmat_{n|\globid}^\top + (1+\beta)\,\ellscale\mlvarmod(\predmean_{n|\globid},\predact_{n|\globid})$.
Thus, it holds for the proposed tube propagation that $s^+ - \predmean_{n+1|t}\in\{x\mid x^\top \predvar_{n+1|t}^{-1}x\le1\}$. Equivalently, $s^+\in\tube_{n+1|t}$.
\end{proof}
\end{lemma}

\begin{theorem}[Overapproximation of robustly reachable sets]
\label{theo:app_overapproximation_tube}
\begin{proof}[Proof of Theorem~\ref{theo:overapproximation_tube}]
Fix $\globid$ and an input sequence $\predact_{0|t},\dots,\predact_{N-1|t}$.
Define the robust reachable sets $\prs_{n|t}$ as in Definition~\ref{def:robust_reachable_set}.

Support containment: Since $\rvalea_k \in \aleaset$ for all $k$, any realization of $\rvstate_{t+n}$ corresponds to a disturbance sequence
$(w_t,\dots,w_{t+n-1})\in\aleaset^n$. Hence, 
$\mathrm{supp}\!\left(
        \rvstate_{t+n}\mid \rvstate_t=\state_t,\predact_{0:n-1|t}
    \right)
    \subseteq
    \prs_{n|t}$.
Thus, it suffices to show $\prs_{n|t}\subseteq \tube_{n|t}$.

Induction on $n$:
For $n=0$, $\prs_{0|t}=\{\state_t\}=\{\predmean_{0|t}\}\subseteq \tube_{0|t}$, and by convention from Lemma~\ref{lemma_app:outer_ellipsoid} $\predvar_{0|t}=0 \Rightarrow \tube_{0|t}=\{\predmean_{0|t}\}$. 
Take any $s\in\prs_{n|t}$ and any $w\in\aleaset$. By the definition of $\prs_{n+1|t}$, it holds that
$s^+ := \meanenv(s,v) + \cholenv(s,v)w \in \prs_{n+1|t}$,
where $v=\predact_{n|t}+\contgain_{n|t}(s-\predmean_{n|t})$.
Since $s\in\prs_{n|t}\subseteq \tube_{n|t}$, Lemma~\ref{lemma_app:onestep_ellipsoid} implies $s^+\in\tube_{n+1|t}$.
This shows $\prs_{n+1|t}\subseteq \tube_{n+1|t}$. By induction, $\prs_{n|t}\subseteq \tube_{n|t}$ for all $n\in\{0,\dots,N\}$.
Combining the induction with the support containment yields 
$\mathrm{supp}(\rvstate_{t+n}\mid \rvstate_t=\state_t,\predact_{0:n-1|t})\subseteq \prs_{n|t} \subseteq \tube_{n|t}$.
Finally, the equality $\tube_{n|t}=\predmean_{n|t}\oplus \predchol_{n|t}^\top\unitball_{\statesp}$ follows from
$\predchol_{n|t}^\top \predchol_{n|t}= \predvar_{n|t}$ and Lemma~\ref{lemma_app:ellipsoid_to_minkowski}.
\end{proof}
\end{theorem}

\begin{lemma}[Constraint tightening for model uncertainty]
    \label{lemma_app:uncertainty_const_tightening}   
    Let Assumption~\ref{assumption:lipschitz_bounds} hold. Define the certain set \looseness=-1 
    $
        \accset := \{ (\state, \act) \in \statesp \times \actsp \;|\; \tfrac{1}{\dimS}\, \mathrm{tr}\!(\kalgain(\state, \act)) \geq \xi \},
    $
    and let $\mathcal{P} := \{ \state \in \statesp \;\Big|\; (\state - \predmean)^\top (\predvar)^{-1} (\state - \predmean) \leq 1 \}$. Then for $\state \in \mathcal{P}$, $v = \predact + \contgain(\state - \predmean)$, and $\unitball_{\statesp \times \actsp} = \{ \unitvec \in \mathbb{R}^{\dimS + n_{\actsp}} \mid \| \unitvec \|_2 \leq 1 \}$,
    the pair $(\state, v) \in \accset$, provided that
        $(\predmean,\, \predact) \oplus \unitball_{\statesp \times \actsp} \, \lipschitzmod \sqrt{ \left(\lambda_{\mathrm{max}}\left(  \predvar \right)  +\lambda_{\mathrm{max}}\left( \contgain  \predvar \contgain^\top \right)\right) }  \subseteq \accset$.
    
    \begin{proof}
    We can upper bound the increase in uncertainty, indicated by the reduction of the normalized trace of the Kalman gain matrix, as 
    \begin{equation}
    \begin{aligned}
        &\tfrac{1}{\dimS} \mathrm{tr}(\kalgain(\predmean ,\predact)) - \tfrac{1}{\dimS} \mathrm{tr}(\kalgain(\state ,v)) 
         \leq \lipschitzmod \| (\predmean ,\predact) - (\state ,v) \|_2
        = \lipschitzmod \sqrt{\|\state - \predmean \|_2^2 + \|v - \predact \|_2^2}\\
        &= \lipschitzmod \sqrt{\|\state - \predmean \|_2^2 + \|\predact + \contgain (\state - \predmean )- \predact \|_2^2}
        \leq \lipschitzmod \sqrt{ \left(\lambda_{\mathrm{max}}\left(  \predvar \right)  +\lambda_{\mathrm{max}}\left( \contgain  \predvar \contgain^\top \right)\right) } 
    \end{aligned}.
    \end{equation}
    Thus, we can lower bound 
    $\tfrac{1}{\dimS} \mathrm{tr}(\kalgain(\state ,v)) \geq \tfrac{1}{\dimS} \mathrm{tr}(\kalgain(\predmean ,\predact)) - \lipschitzmod \sqrt{ \left(\lambda_{\mathrm{max}}\left(  \predvar \right)  +\lambda_{\mathrm{max}}\left( \contgain  \predvar \contgain^\top \right)\right) }$.
    The Minkowski sum $(\predmean, \predact) \oplus \unitball_{\statesp \times \actsp}\, \lipschitzmod \sqrt{ (\lambda_{\mathrm{max}}(  \predvar )  +\lambda_{\mathrm{max}}( \contgain  \predvar \contgain^\top )) }  \subseteq \accset$
    ensures that for all $\unitvec \in \unitball_{\statesp \times \actsp}$, we have $\tfrac{1}{\dimS}\mathrm{tr}\!(\kalgain\!(\predmean+\lipschitzmod \sqrt{ (\lambda_{\mathrm{max}}(  \predvar ) +\lambda_{\mathrm{max}}( \contgain  \predvar \contgain^\top )) } \,\unitvec_{\state},\, \predact + \lipschitzmod \sqrt{ (\lambda_{\mathrm{max}}(  \predvar )  +\lambda_{\mathrm{max}}( \contgain  \predvar \contgain^\top )) } \,\unitvec_{\act})) \geq \xi.$
    Combining this with the bound above, we have that $\xi \overset{!}{\leq} \tfrac{1}{\dimS} \mathrm{tr}(\kalgain(\predmean ,\predact)) - \lipschitzmod \sqrt{ (\lambda_{\mathrm{max}}(  \predvar )  +\lambda_{\mathrm{max}}( \contgain  \predvar \contgain^\top )) } \leq \tfrac{1}{\dimS} \mathrm{tr}(\kalgain(\state ,v)),$
    leading to $(\state, v) \in \accset$.
    \end{proof}
\end{lemma}

\bibliography{umpsc}
\bibliographystyle{rlj}

\beginSupplementaryMaterials
\section{Further Theoretical Results}
\label{supp:theo}
\subsection{Persistent Feasibility}
\label{supp:persistent_feasibility}
In the following, we showcase one way to show persistent feasibility as mentioned in Section \ref{subsec:mpc_setting}.
To do so, we introduce two additional assumptions:
\begin{assumption}[Contractive behavior within the terminal set]
\label{assumption:terminal_contraction_supp}
    There exists a robust terminal controller $\initpol$, robust positive invariant set $\termset^\epiid \subseteq \stateset$ and a maximum predictive uncertainty $\bar{\lambda}$, such that for all $\predmean_{\predid|\globid} \oplus \predchol_{\predid|\globid}^\top \unitball_\statesp \subseteq \termset^\epiid$ with $\lambda_{\mathrm{max}}(\predvar_{\predid|\globid}) \leq \bar{\lambda}$ it holds that\\ 
    \emph{(a)} the terminal controller satisfies input constraints $\initpol(\predmean_{\predid|\globid}) \oplus \contgain_{\predid|\globid}\predchol_{\predid|\globid}^\top \unitball_\statesp \subseteq \actset$; and \\\emph{(b)} $\predmean_{\predid+1|\globid}, \predvar_{\predid+1|\globid}$  from propagating $\predmean_{\predid|\globid}, \predvar_{\predid|\globid}$ with control $v_{\predid|\globid} = \initpol(\predmean_{\predid|\globid}) + \contgain_{\predid|\globid}(\state - \predmean_{\predid|\globid})$ according to \eqref{eq:mean_pred} and \eqref{eq:var_pred}, satisfy $\lambda_{\mathrm{max}}(\predvar_{\predid+1|\globid}) \leq \bar{\lambda}$ and $\predmean_{\predid+1|\globid} \oplus \predchol_{\predid+1|\globid}^\top \unitball_\statesp \subseteq \termset^\epiid$.
\end{assumption}

\begin{assumption}[PSD dominance of tube propagation]
\label{ass:psd_dominance_supp}
At time $t+1$, consider the shifted candidate sequence $\mathcal U_{0:N-1|t+1}=[\mathcal U_{1:N-1|t},\kappa]+K_{1:N-1|t}\big(z_{0:N|t+1}-z_{1:N|t}\big)$,
with $z_{0|t+1}=s_{t+1}\in\mathcal{P}_{1|t}$ and shape update $P_{0|t+1}=0$, $P_{n+1|t+1}=\Phi(P_{n|t+1},K_{n|t+1},z_{n|t+1},u_{n|t+1})$ defined by \eqref{eq:var_pred}.  
Assume that the resulting shape sequence satisfies $P_{n|t+1} \preceq P_{n+1|t}$ for all $n\in\{0,\dots,N-1\}$.
\end{assumption}
Introducing the additional constraint \eqref{eq:upsi_eig_const_supp}, to account for the condition $\lambda_{\mathrm{max}}(\predvar_{\predhorizon|\globid}) \leq \bar{\lambda}$ required for Assumption \ref{assumption:terminal_contraction_supp}, we have a slightly extended \algo~scheme:
\begin{subequations}\label{eq:upsi_psf_supp}
\begin{align}
    \min_{\predact_{0|\globid}, \dots, \predact_{ \predhorizon-1|\globid}} & \|\act_\globid - \predact_{0|\globid} \|_2^2 \label{eq:mpc_cost_supp}\\
    \text{s.t. } & \predmean_{0|\globid} = \state_\globid, \quad \predvar_{0|\globid} = 0 \label{eq:upsi_init_supp}\\
    &\predmean_{\predid+1 \mid \globid} = \mlmeanmod(\predmean_{\predid \mid \globid}, \predact_{\predid \mid \globid}) \text{ for all }  \predid \in \{0, \dots, \predhorizon -1 \}\label{eq:upsi_mean_pred_supp}\\
    & \predvar_{\predid+1 \mid \globid} =                 \Phi ( \predvar_{\predid \mid \globid},\contgain_{\predid \mid \globid},\predmean_{\predid \mid \globid}, \predact_{\predid \mid \globid}) \text{ for all }  \predid \in \{0, \dots, \predhorizon -1 \}, \text{ see \eqref{eq:var_pred}} \label{eq:upsi_var_pred_supp}\\
    & \predmean_{\predid \mid \globid} \in \stateset \ominus \predchol_{\predid|\globid}^\top \unitball_{\statesp} \text{ for all }  \predid \in \{0, \dots, \predhorizon -1 \}\label{eq:upsi_state_const_supp}\\
    & \predact_{\predid \mid \globid} \in \actset \ominus \contgain_{\predid \mid \globid} \predchol_{\predid|\globid}^\top \unitball_{\statesp}   \text{ for all }  \predid \in \{0, \dots, \predhorizon -1 \} \label{eq:upsi_act_const_supp}\\
    & \predmean_{\predhorizon \mid \globid} \in \termset^\epiid \ominus \predchol_{\predhorizon|\globid}^\top \unitball_{\statesp}\label{eq:upsi_term_const_supp}\\
    & \lambda_{\mathrm{max}}(\predvar_{\predhorizon|\globid}) \leq \bar{\lambda}\label{eq:upsi_eig_const_supp}\\ 
    & (\predmean_{\predid \mid \globid}, \predact_{\predid \mid \globid}) \in \accset^\epiid \ominus g_{\predid \mid \globid} \unitball_{\statesp \times \actsp} \text{ for all } \predid \in \{0, \dots, \predhorizon - 1 \} \label{eq:upsi_cert_const_supp}
\end{align}
\end{subequations}
For which we can show persistent feasibility as presented in Theorem \ref{theo:recusive_feasibility_supp}.
\begin{theorem}[Persistent Feasibility]
\label{theo:recusive_feasibility_supp}

Let Assumptions \ref{assumption:init}-\ref{ass:psd_dominance_supp} hold, and assume that the \algo scheme in \eqref{eq:upsi_psf} is feasible at time $t=0$. Then, \eqref{eq:upsi_psf_supp} is feasible for all times $t>0$ when applying $\predact_{0|\globid}$ to a system \eqref{eq:env_dyn} in closed loop. In particular, we have $\state_\globid\in\stateset$ and $\act_\globid \in \actset$ for all $t>0$.
\begin{proof}
\textbf{Induction start:}
We assume feasibility at time $\globid=0$. In particular, if we initialize in $\termset^0$, rolling out $\initpol$ provides a trivial feasible initial solution.
\textbf{Induction step:}
Assume that \eqref{eq:upsi_psf_supp} is feasible at time $\globid$ with optimal control sequence
$\mathcal{U}^{*,\epiid}_{0:\predhorizon-1|\globid}=\left[
\predact^{*,\epiid}_{0|\globid},\predact^{*,\epiid}_{1|\globid},\dots,\predact^{*,\epiid}_{\predhorizon-1|\globid}\right]$
and the corresponding nominal trajectory
$\mathcal{Z}^{*,\epiid}_{0:\predhorizon|\globid}=\left[\predmean^{*,\epiid}_{0|\globid},\predmean^{*,\epiid}_{1|\globid},\dots,\predmean^{*,\epiid}_{\predhorizon|\globid}\right]$.
After applying $\predact^{*,\epiid}_{0|\globid}$ to the system \eqref{eq:env_dyn}, the successor state satisfies
$\state_{\globid+1}\in\predmean^{*,\epiid}_{1|\globid}\oplus\predchol^{*,\epiid \top}_{1|\globid}\unitball_{\statesp}=\mathcal{P}_{1|\globid}$, by Theorem~\ref{theo:overapproximation_tube}.
We construct a candidate sequence at time $\globid+1$ as
$\mathcal{U}^{\epiid}_{0:\predhorizon-1|\globid+1}=[\mathcal{U}^{*,\epiid}_{1:\predhorizon-1|\globid},\,
\initpol(\predmean^{*,\epiid}_{\predhorizon|\globid})]+K_{1:\predhorizon-1|\globid}\big(z_{0:\predhorizon|\globid+1}-z_{1:\predhorizon|\globid}\big)$,
where $z_{0|\globid+1}=\state_{\globid+1}$ and the nominal rollout satisfies
$z_{n+1|\globid+1}=\mlmeanmod(z_{n|\globid+1},u_{n|\globid+1})$.
We further reset the shape $P_{0|\globid+1}=0$, and propagate the tube shape using \eqref{eq:var_pred}.
By construction, $z_{n|\globid+1}\in \mathcal{P}_{n+1|\globid}$ for all $n\in\{0,\dots,\predhorizon-1\}$.
Assumption~\ref{ass:psd_dominance_supp} yields the tube containment $\mathcal{P}_{n|\globid+1}\subseteq\mathcal{P}_{n+1|\globid}$ for all $n\in\{0,\dots,\predhorizon-1\}$ by construction.
Since the original solution at time $\globid$ satisfied the tightened state constraint
$\predmean^{*,\epiid}_{n+1|\globid}\in\stateset\ominus\predchol^{*,\epiid \top}_{n+1|\globid}\unitball_{\statesp}$, it follows from Lemma~\ref{lemma_app:ellipsoid_to_minkowski_supp} that
$\mathcal{P}_{n+1|\globid}\subseteq\stateset$.
Using tube containment, we obtain
$\mathcal{P}_{n|\globid+1}\subseteq\mathcal{P}_{n+1|\globid}\subseteq\stateset$,
which establishes \eqref{eq:upsi_state_const_supp}.
Similarly, feasibility at time $\globid$ ensures
$\predact^{*,\epiid}_{n+1|\globid}\in\actset\ominus K_{n+1|\globid}\predchol^{*,\epiid \top}_{n+1|\globid}\unitball_{\statesp}$.
Therefore, for all states in $\mathcal{P}_{n+1|\globid}$ the ancillary control law satisfies the input constraints. Since $\mathcal{P}_{n|\globid+1}\subseteq\mathcal{P}_{n+1|\globid}$, the candidate inputs also satisfy \eqref{eq:upsi_act_const_supp}.
The certainty constraint \eqref{eq:upsi_cert_const_supp} follows analogously from Lemma~\ref{lemma_app:uncertainty_const_tightening_supp} and the tube containment.
Finally, feasibility at time $\globid$ implies
$\mathcal{P}_{\predhorizon|\globid}\subseteq\termset^\epiid$,$\lambda_{\max}(P_{\predhorizon|\globid})\le \bar\lambda$.
Using tube containment for $n=\predhorizon-1$ yields
$\mathcal{P}_{\predhorizon-1|\globid+1} \subseteq \mathcal{P}_{\predhorizon|\globid} \subseteq \termset^\epiid,
 \lambda_{\max}(P_{\predhorizon-1|\globid+1})\le\bar\lambda$.
Applying Assumption~\ref{assumption:terminal_contraction_supp} with the terminal controller $\initpol$ then yields
$\mathcal{P}_{\predhorizon|\globid+1}\subseteq\termset^\epiid, \lambda_{\max}(P_{\predhorizon|\globid+1})\le\bar\lambda,$
which establishes \eqref{eq:upsi_term_const_supp} and \eqref{eq:upsi_eig_const_supp}.
Thus, the candidate sequence satisfies all constraints of \eqref{eq:upsi_psf_supp}, and the problem remains feasible at time $\globid+1$. By induction, feasibility holds for all $t>0$.
\end{proof}
\end{theorem}

\subsection{Extended Proofs}
\label{supp:proofs}
\begin{lemma}[Ancillary controller]
    \label{lemma_app:ancilliary_cont}
    Adding an ancillary control law to the PSF action $v_{\predid|\globid} = \contgain_{\predid|\globid}(\state_{\predid|\globid}-\predmean_{\predid|\globid}) + \predact_{\predid|\globid}$ yields dynamics $\cloopmat_{\predid|\globid} = \sysmat_{\predid|\globid} + \inmat_{\predid|\globid} \contgain_{\predid|\globid}$ regarding the deviation state 
    $\diff_{\predid|\globid} = \state_{\predid|\globid} - \predmean_{\predid|\globid}$ when propagating through the linearized dynamics $\state_{\predid+1|\globid} = \sysmat_{\predid|\globid} \state_{\predid|\globid} + \inmat_{\predid|\globid} v_{\predid|\globid}$.
    Here, the Jacobians $\sysmat_{\predid|\globid} = \grad_\state \mlmeanmod(\state, \act)|_{s=\predmean_{\predid|\globid}, a=\predact_{\predid|\globid}} $ and $\inmat_{\predid|\globid} = \grad_\act \mlmeanmod(\state, \act)|_{s=\predmean_{\predid|\globid}, a=\predact_{\predid|\globid}} $ are obtained from linearizing the nominal model dynamics at the nominal trajectory.
    \begin{proof}
    The result follows trivially as
    \begin{equation}
    \begin{aligned}
        \diff_{\predid+1|\globid} &= \state_{\predid+1|\globid} - \predmean_{\predid+1|\globid} = \sysmat_{\predid|\globid} \state_{\predid|\globid} + \inmat_{\predid|\globid}(\contgain_{\predid|\globid}(\state_{\predid|\globid} - \predmean_{\predid|\globid}) + \predact_{\predid|\globid}) - (\sysmat_{\predid|\globid} \predmean_{\predid|\globid} + \inmat_{\predid|\globid} \predact_{\predid|\globid})\\
        &= (\sysmat_{\predid|\globid} + \inmat_{\predid|\globid} \contgain_{\predid|\globid}) \state_{\predid|\globid} - (\sysmat_{\predid|\globid} + \inmat_{\predid|\globid} \contgain_{\predid|\globid}) \predmean_{\predid|\globid} = \cloopmat_{\predid|\globid} \diff_{\predid|\globid}.
    \end{aligned}
    \end{equation}
    \end{proof}
\end{lemma}

\begin{lemma}[Linearization error bound on the nominal dynamics; extended version of Lemma~\ref{lemma_app:lin_err_bound}]
    \label{lemma_app:lin_err_bound_supp}
    Let Assumption~\ref{assumption:lipschitz_bounds} hold.
    The linearization error $e_{\predid \mid \globid}$ for the propagation of predictive shapes for all $P\succ0$ is upper-bounded by $$\|e_{\predid \mid \globid}\|_2 \leq e^{\mathrm{max}}_{\predid \mid \globid} = \frac{\lipschitz}{2} \left(  \sqrt{ \lambda_\mathrm{max}(\predvar)} + \sqrt{ \lambda_{\mathrm{max}}\left( \contgain  \predvar \contgain^\top \right)} \right)^2.$$
\begin{proof}
    Through the linearization, we propagate through a first-order Taylor approximation of the nonlinear nominal dynamics
    \begin{equation}
    \begin{aligned}
        \meanmod(\state, \contgain(\state-\predmean) + \predact)& = \meanmod(\predmean, \predact) + \sysmat(\state-\predmean) + \inmat\contgain(\state - \predmean) + e\\
        & = \meanmod(\predmean, \predact) + \grad_\state \meanmod \diff + \grad_\act \meanmod \contgain \diff + e
       \end{aligned} 
    \end{equation}
    with deviation $\diff := (\state-\predmean)$ and the linearization error 
    \begin{equation}
        e = \frac{1}{2} \left[ \diff^\top (\contgain\diff)^\top \right] \hess \begin{bmatrix} \diff \\ \contgain\diff \end{bmatrix}
    \end{equation}
    depending on the Hessian $\hess$. For each output dimension $i \in \{1, \dots, \dimS \}$ the Hessian is given as
    \begin{equation}
        \hess_i = \begin{bmatrix}
            \grad_\state \grad_\state \meanmod_i  & \grad_\state \grad_\act \meanmod_i \\\grad_\act \grad_\state \meanmod_i & \grad_\act \grad_\act \meanmod_i
        \end{bmatrix}.
    \end{equation}
    Thus, the $i$-th component of the linearization error is
    \begin{equation}
        \begin{aligned}
            e_i 
            &= \frac{1}{2} \left[ \diff^\top (\contgain\diff)^\top \right] \hess_i \begin{bmatrix} \diff \\ \contgain\diff \end{bmatrix}
            = \frac{1}{2} \left[ \diff^\top (\contgain\diff)^\top \right] \begin{bmatrix}
                \grad_\state \grad_\state \meanmod_i \diff + \grad_\state \grad_\act \meanmod_i \contgain \diff \\ \grad_\act \grad_\state \meanmod_i \diff + \grad_\act \grad_\act \meanmod_i \contgain \diff \end{bmatrix}\\
            &=\frac{1}{2}\left(\diff^\top \left( \grad_\state \grad_\state \meanmod_i \diff + \grad_\state \grad_\act \meanmod_i \contgain \diff \right) + (\contgain \diff)^\top \left( \grad_\act \grad_\state \meanmod_i \diff + \grad_\act \grad_\act \meanmod_i \contgain \diff \right)\right)\\
            &= \frac{1}{2} \left( \diff^\top \grad_\state \grad_\state \meanmod_i \diff + \diff^\top \grad_\state \grad_\act \meanmod_i \contgain \diff + (\contgain \diff)^\top \grad_\act \grad_\state \meanmod_i \diff + (\contgain \diff)^\top \grad_\act \grad_\act \meanmod_i \contgain \diff \right)\\
            &\overset{(a)}{=} \frac{1}{2} \left( \diff^\top \grad_\state \grad_\state \meanmod_i \diff + 2 \diff^\top \grad_\state \grad_\act \meanmod_i \contgain \diff +  (\contgain \diff)^\top \grad_\act \grad_\act \meanmod_i \contgain \diff \right)
        \end{aligned}
    \end{equation}
    where equality $(a)$ holds since $\meanmod_i$ is twice continuously differentiable and therefore by Schwarz's theorem $\grad_\act \grad_\state \meanmod_i = \grad_\state \grad_\act \meanmod_i$ and since $(\contgain \diff)^\top \grad_\act \grad_\state \meanmod_i \diff$ is a scalar. Combining the output dimensions $i$, we have
    \begin{equation}
        \begin{aligned}
            e &= \left[ e_1, e_2, \dots, e_{\dimS} \right]^\top
            = \frac{1}{2}\left[ \diff^\top (\contgain \diff)^\top\right] \hess \begin{bmatrix} \diff \\ \contgain\diff \end{bmatrix}\\
            &= \frac{1}{2} \left( \diff^\top \grad_\state \grad_\state \meanmod \diff + 2 \diff^\top \grad_\state \grad_\act \meanmod \contgain \diff +  (\contgain \diff)^\top \grad_\act \grad_\act \meanmod \contgain \diff \right).
        \end{aligned}
    \end{equation}
  
    Using Assumption~\ref{assumption:lipschitz_bounds} allows us to formulate an upper bound on the L2 norm of the linearization error
    \begin{equation}
    \begin{aligned}
    \label{eq:app_lin_err_bound_1}
        \|e\|_2 &= \frac{1}{2} \| \diff^\top \grad_\state \grad_\state \meanmod \diff + 2 \diff^\top \grad_\state \grad_\act \meanmod \contgain \diff +  (\contgain \diff)^\top \grad_\act \grad_\act \meanmod \contgain \diff \|_2\\
        & \leq \frac{1}{2} \left( \| \grad_\state \grad_\state \meanmod \| \| \diff \|_2^2 + 2 \| \diff \|_2 \| \grad_\state \grad_\act \meanmod \| \| \contgain \diff \|_2 + \| \grad_\act \grad_\act \meanmod \| \| \contgain\diff \|_2^2\right)\\
        & \leq \frac{\lipschitz}{2} \left( \| \diff \|_2^2 + 2 \| \diff \|_2 \| \contgain \diff \|_2 +  \|\contgain\diff \|_2^2\right)\\
    \end{aligned}
    \end{equation}

    We  upper bound the components starting with $\| \diff \|_2^2 = \| \state - \predmean \|_2^2$. Since $\state\in\mathcal{P}$, the deviation satisfies 
    \begin{equation}
        \label{eq:app_delta_max_start_supp}
        \diff^\top (\predvar)^{-1}\diff \le 1.
    \end{equation}
    Further, since $\predvar$ is a symmetric positive definite shape matrix, there exist matrices $Q$ and $\Lambda$ such that
    \begin{equation}
        \predvar = Q \Lambda Q^\top \text{ with } Q^\top Q = I \text{ and } \Lambda = \mathrm{diag}(\lambda_1(\predvar), \dots, \lambda_{\dimS}(\predvar))
    \end{equation}
    with $\lambda_i(\predvar)$ being the $i$-th eigenvalue of $\predvar$. Thus, we have
    \begin{equation}
    \begin{aligned}
        \diff^\top  (\predvar)^{-1} \diff &= \diff^\top  (Q \Lambda Q^\top)^{-1} \diff
        = (Q^\top \diff)^\top \Lambda^{-1} Q^\top \diff = y^\top \Lambda^{-1} y = \sum_{i=1}^{\dimS} \frac{y_i^2}{ \lambda_i(\predvar)} \leq 1
    \end{aligned}
    \end{equation}
    with $y = Q^\top \diff$. Further,
    \begin{equation}
        \begin{aligned}
            1 \geq \sum_{i=1}^{\dimS} \frac{y_i^2}{ \lambda_i(\predvar)} \geq \frac{1}{ \lambda_\mathrm{max}(\predvar)} \sum_{i=1}^{\dimS} y_i^2 = \frac{\|y\|_2^2}{ \lambda_\mathrm{max}(\predvar)} = \frac{(Q^\top \diff)^\top (Q^\top \diff)}{ \lambda_\mathrm{max}(\predvar)} = \frac{\|\diff\|_2^2}{ \lambda_\mathrm{max}(\predvar)} 
        \end{aligned}
    \end{equation}
    yields the upper bound
    \begin{equation}
        \label{eq:app_diff_bound_supp}
         \lambda_\mathrm{max}(\predvar) \geq \|\diff\|_2^2
    \end{equation}

    Next, we compute an upper bound for $\|\contgain \diff \|_2^2$. Since $\predvar$ is symmetric positive definite, there exists the square root $( \predvar)^{-\frac{1}{2}}$. We define $\Tilde{\diff}:=( \predvar)^{-\frac{1}{2}}\diff$ with $\| \Tilde{\diff} \|_2^2 \leq 1$ as can be seen from substituting $\Tilde{\diff}$ into \eqref{eq:app_delta_max_start_supp}.
    \begin{equation}
        \diff^\top  ( \predvar)^{-1} \diff = \Tilde{\diff}^\top \left( ( \predvar)^{\frac{1}{2}} \right)^\top  ( \predvar)^{-1} \left( ( \predvar)^{\frac{1}{2}} \right)\Tilde{\diff} = \Tilde{\diff}^\top \Tilde{\diff} = \| \Tilde{\diff} \|_2^2 \leq 1
    \end{equation}
    Using this fact, we substitute
    \begin{equation}
    \begin{aligned}
        \| \contgain \diff\|_2^2 &= \| \contgain ( \predvar)^{\frac{1}{2}} \Tilde{\diff}\|_2^2 = (\contgain ( \predvar)^{\frac{1}{2}} \Tilde{\diff})^\top (\contgain ( \predvar)^{\frac{1}{2}} \Tilde{\diff}) = \Tilde{\diff}^\top ( \predvar)^{\frac{1}{2}} \contgain^\top \contgain ( \predvar)^{\frac{1}{2}}  \Tilde{\diff}\\
        & \leq \lambda_{\mathrm{max}}\left( ( \predvar)^{\tfrac{1}{2}} \contgain^\top \contgain ( \predvar)^{\tfrac{1}{2}}\right)\| \Tilde\diff\|_2^2 
        \leq \lambda_{\mathrm{max}}\left( ( \predvar)^{\frac{1}{2}} \contgain^\top \contgain ( \predvar)^{\frac{1}{2}} \right) = \\
        & \lambda_{\mathrm{max}}\left( \contgain ( \predvar)^{\frac{1}{2}}  ( \predvar)^{\frac{1}{2}} \contgain^\top \right)
        =  \lambda_{\mathrm{max}}\left( \contgain  \predvar \contgain^\top \right).
    \end{aligned}
    \end{equation}
    Since both $\predvar$ and $\contgain \predvar \contgain^\top$ are symmetric positive definite, we can further bound $\|\diff\|_2 \leq \sqrt{ \lambda_\mathrm{max}(\predvar)}$ and $\| \contgain \diff \|_2 \leq \sqrt{ \lambda_{\mathrm{max}}\left( \contgain  \predvar \contgain^\top \right)}$. Substituting these bounds in \eqref{eq:app_lin_err_bound_1} yields the desired result.
    \begin{equation}
        \begin{aligned}
           \| e \| & \leq \frac{\lipschitz}{2} \left( \| \diff \|_2^2 + 2 \| \diff \|_2 \| \contgain \diff \|_2 + \| \contgain\diff \|_2^2\right)\\
           &\leq \frac{\lipschitz}{2} \left(  \lambda_\mathrm{max}(\predvar) + 2 \sqrt{ \lambda_\mathrm{max}(\predvar)} \sqrt{ \lambda_{\mathrm{max}}\left( \contgain  \predvar \contgain^\top \right)} +  \lambda_{\mathrm{max}}\left( \contgain  \predvar \contgain^\top \right) \right)\\
           &\leq \frac{ \lipschitz}{2} \left(  \sqrt{ \lambda_\mathrm{max}(\predvar)} + \sqrt{ \lambda_{\mathrm{max}}\left( \contgain  \predvar \contgain^\top \right)} \right)^2
        \end{aligned}
    \end{equation}
    \end{proof}
\end{lemma}

\begin{lemma}[Linearization error for aleatoric variance; extended version of Lemma~\ref{lemma_app:lin_err_var}]
\label{lemma_app:lin_err_var_supp}
Let Assumption~\ref{assumption:lipschitz_bounds} hold. 
The state-space error induced by approximating the noise injection
$\cholenv(\state,v)$ by $\cholenv(\predmean,\predact)$, that is, $e_{\varenv} := \Big(\cholenv(\state,v)-\cholenv(\predmean,\predact)\Big)w$ with $w\in\aleaset,\ \|w\|_2^2\le \ellscale$, is bounded by $\|e_{\varenv}\|_2
\le
\lipschitzvarchol\sqrt{\ellscale}\sqrt{\lambda_{\max}(\predvar)+\lambda_{\max}(\contgain\predvar\contgain^\top)}$.

\begin{proof}
By definition,
\begin{equation}
\label{eq:app_eSigma_submult_supp}
    \|e_{\varenv}\|_2
    \le
    \|\cholenv(\state,v)-\cholenv(\predmean,\predact)\|\,\|w\|_2.
\end{equation}
Since $w\in\aleaset$ we have $\|w\|_2\le \sqrt{\ellscale}$.
Moreover, by Assumption~\ref{assumption:lipschitz_bounds},
\begin{equation}
    \|\cholenv(\state,v)-\cholenv(\predmean,\predact)\|
    \le
    \lipschitzvarchol \|(\state,v)-(\predmean,\predact)\|_2
    =
    \lipschitzvarchol \sqrt{\|\diff\|_2^2+\|\contgain\diff\|_2^2}.
\end{equation}
Combining these two yields
\begin{equation}
\label{eqn:912d9460}
    \|e_{\varenv}\|_2
    \le
    \lipschitzvarchol\sqrt{\ellscale}\,\sqrt{\|\diff\|_2^2+\|\contgain\diff\|_2^2}.
\end{equation}
It remains to bound $\|\diff\|_2^2$ and $\|\contgain\diff\|_2^2$ for $\state\in\mathcal P$.
From $\state\in\mathcal P$ we have
\begin{equation}
\label{eq:app_delta_ellipsoid_ineq_supp}
    \diff^\top(\predvar)^{-1}\diff \le 1,
\end{equation}
which gives
\begin{equation}
    \|\diff\|_2^2 \le\lambda_{\max}(\predvar).
\end{equation}
Similarly,
\begin{equation}
    \|\contgain\diff\|_2^2
    =
    \diff^\top \contgain^\top\contgain \diff
    \le\lambda_{\max}(\contgain\predvar\contgain^\top).
\end{equation}
Substituting into \eqref{eqn:912d9460} gives the bound.
\end{proof}
\end{lemma}

\begin{lemma}[Total linearization error; extended version of Lemma~\ref{lemma_app:total_lin_err}]
    \label{lemma_app:total_lin_err_supp}
    Let Assumption~\ref{assumption:lipschitz_bounds} hold. The overall linearization error $e$ of the \algo set propagation is upper bounded by
    \begin{equation}
        \|e\|_2 \leq e^{\mathrm{max}} = \frac{\lipschitz}{2} \left(  \sqrt{ \lambda_\mathrm{max}(\predvar)} + \sqrt{ \lambda_{\mathrm{max}}\left( \contgain  \predvar \contgain^\top \right)} \right)^2 + \lipschitzvarchol \sqrt{\ellscale}\sqrt{ \lambda_{\mathrm{max}}\left( \predvar \right) + \lambda_{\mathrm{max}}\left( \contgain  \predvar \contgain^\top \right)}
    \end{equation}
    \begin{proof}
        The overall linearization error comprises the linearization error of the nominal dynamics and the process noise variance $e = e_{\grad\meanenv} + e_{\varenv}$. Using Lemmas \ref{lemma_app:lin_err_bound_supp} and \ref{lemma_app:lin_err_var_supp}, we can define the upper bound on linearization error norm as
        \begin{equation}
        \begin{aligned}
            \| e \|_2 &= \| e_{\grad\meanenv} + e_{\varenv} \|_2 \leq \| e_{\grad\meanenv} \|_2 + \| e_{\varenv} \|_2\\
            & \leq \frac{\lipschitz}{2} \left(  \sqrt{ \lambda_\mathrm{max}(\predvar)} + \sqrt{ \lambda_{\mathrm{max}}\left( \contgain  \predvar \contgain^\top \right)} \right)^2 + \lipschitzvarchol \sqrt{\ellscale} \sqrt{ \lambda_{\mathrm{max}}\left( \predvar \right) + \lambda_{\mathrm{max}}\left( \contgain  \predvar \contgain^\top \right)}\\
            & = : e^{\mathrm{max}}
        \end{aligned}
        \end{equation}
        that yields the desired result.
    \end{proof}
\end{lemma}

\begin{lemma}[Outer ellipsoid of a Minkowski sum; extended version of Lemma~\ref{lemma_app:outer_ellipsoid}]
    \label{lemma_app:outer_ellipsoid_supp}
    Let $\mathcal{E}(A):=\{x\mid x^\top A^{-1}x\le1\},$ as well as $\mathcal{E}(B):=\{x\mid x^\top B^{-1}x\le1\}$ be two ellipsoids with $A,B\succ0$. Then a valid outer approximation of the Minkowski sum \begin{equation}
        \mathcal{E}(A)\oplus\mathcal{E}(B)\subseteq\mathcal{E}(C)
    \end{equation} is achieved with
    \begin{equation}
        C = \left(1+\tfrac{1}{\beta}\right)A + \left(1+\beta\right)B,
    \end{equation}
    for every $\beta>0$. Furthermore, the minimal valid outer approximation of the Minkowski sum has
    \begin{equation}
        \beta^* = \sqrt{\tfrac{\mathrm{tr}(A)}{\mathrm{tr}(B)}}.
    \end{equation}
    In the degenerate case $A = 0$, we adopt the convention $\mathcal{E}(A) = \{0\}$, so that the sum reduces to $\mathcal{E}(B)$ and $C = B$ (and similarly for $B = 0$).
    \begin{proof}
    For the degenerate case $A = 0$, the ellipsoid collapses to its center, so $\mathcal{E}(0) = \{0\}$. Since $\mathcal{E}(0) \oplus \mathcal{E}(B) = \{0\} \oplus \mathcal{E}(B) = \mathcal{E}(B)$, the sum equals $\mathcal{E}(B)$, and the case $B = 0$ follows similarly. For the non-degenerate case $A, B \succ 0$, to be a valid overapproximation, we need to find $C$ such that the support of $\mathcal{E}(A)\oplus\mathcal{E}(B)$ lies within the support of $\mathcal{E}(C)$ for any arbitrary direction vector $v$. That is
    \begin{equation}
        \begin{aligned}
            h_{\mathcal{E}(C)}(v) \overset{!}{\geq} h_{\mathcal{E}(A)\oplus\mathcal{E}(B)}(v) = h_{\mathcal{E}(A)}(v) + h_{\mathcal{E}(B)}(v) \; &\Rightarrow \;
            \sqrt{v^\top C\, v} \overset{!}{\geq} \sqrt{v^\top A v} + \sqrt{v^\top B v} \\
            v^\top C\, v &\overset{!}{\geq} \left( \sqrt{v^\top A v} + \sqrt{v^\top B v} \right)^2.
        \end{aligned}
    \end{equation}
    We use the following outer approximation from \cite[Chp. 2.2 or 2.4]{kurzhanski1997ellipsoidal}, similarly used in~\cite{halder_parameterized_2018}:
    \begin{equation}
    \label{eq:app_kv_supp}
        \left( \sqrt{v^\top A v} + \sqrt{v^\top B v} \right)^2 \leq \left(1+\tfrac{1}{\beta}\right) v^\top A v + \left(1+\beta\right) v^\top B v
    \end{equation}
    for $\beta > 0$. Consequently, for any $\beta > 0$
    \begin{equation}
    \label{eq:app_beta_approx_supp}
        C = \left(1+\tfrac{1}{\beta}\right)A + \left(1+\beta\right)B
    \end{equation}
    yields a valid overapproximation of $\mathcal{E}(A)\oplus\mathcal{E}(B)$. To find the minimal valid overapproximation, we determine $\beta$ to minimize $\mathrm{tr}(C) = \left(1+\tfrac{1}{\beta}\right)\mathrm{tr}(A) + \left(1+\beta\right)\mathrm{tr}(B)$. Thus, we get
    \begin{equation}
    \label{eq:app_opt_beta_supp}
        \grad_\beta \left[ \left(1+\tfrac{1}{\beta}\right)\mathrm{tr}(A) + \left(1+\beta\right)\mathrm{tr}(B) \right] \overset{!}{=} 0, \; \Rightarrow \;
        -\tfrac{1}{\beta^2}\mathrm{tr}(A) + \mathrm{tr}(B) = 0,
    \end{equation}
    which yields $\beta^* = \sqrt{\tfrac{\mathrm{tr}(A)}{\mathrm{tr}(B)}}$. Inserting into \eqref{eq:app_beta_approx_supp} yields the desired result.
    \end{proof}
\end{lemma}

\begin{lemma}[Outer approximation of process noise; extended version of Lemma~\ref{lemma_app:gauss_lvl_set}]
    \label{lemma_app:gauss_lvl_set_supp}
    Fix $\globid$ and a nominal pair $(\predmean_{n|t},\predact_{n|t})$, and let Assumption~\ref{assumption:consistent_estimator} hold.
    For the process noise $L_{n|t}\alea$ with $L_{n|t}:=\cholenv(\predmean_{n|t},\predact_{n|t})$,
    $\|\alea\|_2^2\le\ellscale$, and $L_{n|t} L_{n|t}^\top=\varenv(\predmean_{n|t},\predact_{n|t})$, it holds that
    \begin{equation}
        \label{eq:app_noise_ellips_supp}
        L_{n|t}\alea
        \in
        \Big\{ x \ \big| \ x^\top\big(\ellscale\,\mlvarmod(\predmean_{n|t},\predact_{n|t})\big)^{-1}x \le 1\Big\}.
    \end{equation}
    \begin{proof}
        Since the process noise $\alea$ is bounded, recall $\|\alea\|_2^2\le\ellscale$, the Gaussian aleatoric covariance estimate $\mlvarmod$ from Section~\ref{subsec:cert_set} can be used to bound $L_{n|t}\alea$, as we show below. Substituting $x = L_{n|t}\alea$ into the quadratic form of the ellipsoid with shape matrix $\ellscale\,\varenv(\predmean_{n|t},\predact_{n|t})=\ellscale\,L_{n|t}L_{n|t}^\top$ and using $\|\alea\|_2^2\le\ellscale$ yields
        \begin{equation}
            \label{eq:app_noise_collapse_supp}
            (L_{n|t}\alea)^\top\big(\ellscale\,L_{n|t}L_{n|t}^\top\big)^{-1}(L_{n|t}\alea)
            = \tfrac{1}{\ellscale}\,\alea^\top \alea
            \le \tfrac{\ellscale}{\ellscale}
            = 1,
        \end{equation}
        so $L_{n|t}\alea$ lies in the ellipsoid with shape matrix $\ellscale\,\varenv(\predmean_{n|t},\predact_{n|t})$.
        By Assumption~\ref{assumption:consistent_estimator}, we have an estimate for which it holds $\mlvarmod(\predmean_{n|t},\predact_{n|t})\succcurlyeq\varenv(\predmean_{n|t},\predact_{n|t})$, so an ellipsoid with shape matrix $\ellscale\,\varenv(\predmean_{n|t},\predact_{n|t})$ is contained in the one with shape matrix $\ellscale\,\mlvarmod(\predmean_{n|t},\predact_{n|t})$, which yields~\eqref{eq:app_noise_ellips_supp}.
    \end{proof}
\end{lemma}

\begin{lemma}[Reformulation of an ellipsoid as a Minkowski sum; extended version of Lemma~\ref{lemma_app:ellipsoid_to_minkowski}]
    \label{lemma_app:ellipsoid_to_minkowski_supp}
    Let $\predvar$ be a symmetric positive definite matrix with Cholesky decomposition 
    $\predvar = \predchol^\top \predchol$. Then the ellipsoidal set
    \begin{equation}
    \label{eq:app_ellipsoid_supp}
        \mathcal{P} = \left\{ \state \in \statesp \mid (\state - \predmean)^\top (\predvar)^{-1} (\state - \predmean) \leq 1 \right\}
    \end{equation}
    is equivalent to the set
    \begin{equation}
    \label{eq:app_minkset_supp}
        \mathcal{Q} = \left\{ \predmean \oplus \predchol^\top q \mid q \in \mathbb{R}^{\dimS},\, \| q \|_2 \leq 1 \right\}.
    \end{equation}
    \begin{proof}
    We show set equivalence by proving both inclusions $\mathcal{Q} \subseteq \mathcal{P}$ 
    and $\mathcal{P} \subseteq \mathcal{Q}$.
    
    By the definition of the Minkowski sum, $\mathcal{Q}$ can be written as
    \begin{equation}
    \label{eq:app_reform_mink_supp}
        \mathcal{Q} = \left\{ \predmean \oplus \predchol^\top q \mid \|q\|_2 \leq 1 \right\} 
        = \left\{ \predmean + \predchol^\top q \mid \|q\|_2 \leq 1 \right\},
    \end{equation}
    i.e., $\mathcal{Q}$ consists of all points of the form $\state = \predmean + \predchol^\top q$ 
    where $q$ lies in the unit ball.

    Part 1: $\mathcal{Q} \subseteq \mathcal{P}$.
    Let $\state \in \mathcal{Q}$ be arbitrary. By \eqref{eq:app_reform_mink_supp}, there exists a 
    $q \in \mathbb{R}^{\dimS}$ with $\|q\|_2 \leq 1$ such that $\state = \predmean + \predchol^\top q$.
    Substituting this expression for $\state$ into the quadratic form in \eqref{eq:app_ellipsoid_supp} 
    and using the Cholesky decomposition $\predvar = \predchol^\top \predchol$ yields
    \begin{equation}
        \begin{aligned}
            (\state - \predmean)^\top (\predvar)^{-1} (\state - \predmean) 
            = (\predchol^\top q)^\top (\predvar)^{-1} (\predchol^\top q) 
            = (\predchol^\top q)^\top (\predchol^\top \predchol)^{-1} (\predchol^\top q) 
            = q^\top q = \|q\|_2^2.
        \end{aligned}
    \end{equation}
    Since $\|q\|_2 \leq 1$ by assumption, it follows that $\|q\|_2^2 \leq 1$, and therefore
    \begin{equation}
        (\state - \predmean)^\top (\predvar)^{-1} (\state - \predmean) = \|q\|_2^2 \leq 1,
    \end{equation}
    which means $\state \in \mathcal{P}$. Since $\state \in \mathcal{Q}$ was arbitrary, 
    we have $\mathcal{Q} \subseteq \mathcal{P}$.

    Part 2: $\mathcal{P} \subseteq \mathcal{Q}$.
    Let $\state \in \mathcal{P}$ be arbitrary, so that 
    $(\state - \predmean)^\top (\predvar)^{-1} (\state - \predmean) \leq 1$.
    Since $\predvar$ is symmetric positive definite, its Cholesky factor $\predchol$ 
    is invertible. We can therefore define
    \begin{equation}
        q := (\predchol^\top)^{-1}(\state - \predmean),
    \end{equation}
    which by construction satisfies $\state = \predmean + \predchol^\top q$, i.e., $\state$ is of 
    the form required by \eqref{eq:app_reform_mink_supp}. It remains to show that $\|q\|_2 \leq 1$.
    Expanding $\|q\|_2^2$ and using the Cholesky decomposition gives
    \begin{equation}
        \begin{aligned}
            \|q\|_2^2 &= q^\top q 
            = \left((\predchol^\top)^{-1}(\state-\predmean)\right)^\top \left((\predchol^\top)^{-1}(\state-\predmean)\right) \\
            &= (\state - \predmean)^\top (\predchol^\top \predchol)^{-1} (\state - \predmean) 
            = (\state - \predmean)^\top (\predvar)^{-1} (\state - \predmean) \leq 1,
        \end{aligned}
    \end{equation}
    where we used $\left((\predchol^\top)^{-1}\right)^\top = \predchol^{-1}$ in the third line,
    and the last inequality holds since $\state \in \mathcal{P}$. Therefore $\|q\|_2^2 \leq 1$, 
    which implies $\|q\|_2 \leq 1$. Since $\state = \predmean + \predchol^\top q$ with $\|q\|_2 \leq 1$, 
    we conclude $\state \in \mathcal{Q}$. Since $\state \in \mathcal{P}$ was arbitrary, 
    we have $\mathcal{P} \subseteq \mathcal{Q}$.

    Since $\mathcal{Q} \subseteq \mathcal{P}$ and $\mathcal{P} \subseteq \mathcal{Q}$, 
    we conclude $\mathcal{P} = \mathcal{Q}$, which yields the desired result.
    \end{proof}
\end{lemma}

\begin{lemma}[One-step ellipsoidal propagation; extended version of Lemma~\ref{lemma_app:onestep_ellipsoid}]
\label{lemma_app:onestep_ellipsoid_supp}
Fix $\globid$ and a nominal pair $(\predmean_{n|t},\predact_{n|t})$.
Let Assumptions~\ref{assumption:consistent_estimator},~\ref{assumption:unbiased_estimator}, and~\ref{assumption:lipschitz_bounds} hold.
Let $\predvar_{n+1|t}$ be obtained by repeated outer-ellipsoid inflation following Lemma~\ref{lemma_app:outer_ellipsoid_supp}.
Define the tube
\begin{equation}
\label{eq:onestep_tube_np1_supp}
    \tube_{n+1|t} :=
    \Big\{ s\in\statesp \ \Big|\ (s-\predmean_{n+1|t})^\top (\predvar_{n+1|t})^{-1}(s-\predmean_{n+1|t})\le 1\Big\}.
\end{equation}
Then for every $s\in\tube_{n|t}$ and every $w\in\aleaset$, the successor $s^+$
satisfies $s^+ \in \tube_{n+1|t}$.
\begin{proof}
Take any $s\in\tube_{n|t}$ and let $\diff=s-\predmean_{n|t}$.
By a first-order Taylor expansion of the nominal dynamics around $(\predmean_{n|t},\predact_{n|t})$ under the ancillary law
(and collecting higher-order terms),
\begin{equation}
\label{eq:onestep_taylor_mu_supp}
    \meanenv(s,v)
    =
    \meanenv(\predmean_{n|t},\predact_{n|t})
    +
    \cloopmat_{n|t}\diff
    +
    e_{\grad\meanenv}.
\end{equation}
Similarly, write the noise-injection term as nominal injection plus mismatch-induced state error,
\begin{equation}
\label{eq:onestep_noise_split_supp}
    \cholenv(s,v)\,w
    =
    \cholenv(\predmean_{n|t},\predact_{n|t})\,w
    +
    e_{\varenv}.
\end{equation}
Using the definition $s^+ = \meanenv(s,v) + \cholenv(s,v)\,w$ and
$\predmean_{n+1|t}=\meanenv(\predmean_{n|t},\predact_{n|t})$ yields
\begin{equation}
\label{eq:onestep_dev_supp}
    s^+ - \predmean_{n+1|t}
    =
    \cloopmat_{n|t}\diff
    +
    \cholenv(\predmean_{n|t},\predact_{n|t})\,w
    +
    e,
\end{equation}
where $e:=e_{\grad\meanenv}+e_{\varenv}$.
Since $s\in\tube_{n|t}$, the deviation $\diff$ lies in the ellipsoid with shape matrix $\predvar_{n|t}$, hence
$\cloopmat_{n|t}\diff$ lies in the ellipsoid with shape matrix $\cloopmat_{n|t}\predvar_{n|t}\cloopmat_{n|t}^\top$.
Moreover, since $w\in\aleaset$, by Lemma~\ref{lemma_app:gauss_lvl_set_supp} the term $\cholenv(\predmean_{n|t},\predact_{n|t})\,w$
lies in the ellipsoid with shape matrix $\ellscale\,\mlvarmod(\predmean_{n|t},\predact_{n|t})$.
Finally, by Lemma~\ref{lemma_app:total_lin_err_supp} the linearization error satisfies $\|e\|_2\le e^{\mathrm{max}}_{n|t}$, i.e.
\begin{equation}
\label{eq:onestep_errball_supp}
    e\in\linerr_{n|t}:=\{e\in\mathbb R^{\dimS}\mid \|e\|_2\le e^{\mathrm{max}}_{n|t}\}.
\end{equation}
Therefore, the set of possible deviations $s^+-\predmean_{n+1|t}$ is contained in the Minkowski sum
\begin{equation}
\label{eq:onestep_mink_sum_supp}
    \Big\{x\,\big|\,x^\top(\cloopmat_{n|t}\predvar_{n|t}\cloopmat_{n|t}^\top)^{-1}x\le1\Big\}
    \ \oplus\
    \Big\{x\,\big|\,x^\top(\ellscale\,\mlvarmod(\predmean_{n|t},\predact_{n|t}))^{-1}x\le1\Big\}
    \ \oplus\
    \linerr_{n|t}.
\end{equation}
We resolve~\eqref{eq:onestep_mink_sum_supp} by applying the outer-ellipsoid inflation
(Lemma~\ref{lemma_app:outer_ellipsoid_supp}) twice.
Applied to the first two terms, it gives the outer ellipsoid with shape matrix
\begin{equation}
\label{eq:onestep_Qbeta_supp}
    \prepredvar_{n+1|t}
    =
    (1+\beta^{-1})\,\cloopmat_{n|t}\predvar_{n|t}\cloopmat_{n|t}^\top
    +(1+\beta)\,\ellscale\,\mlvarmod(\predmean_{n|t},\predact_{n|t})
\end{equation}
for any $\beta \in \mathbb{R}$ and $\beta>0$, with the minimizing choice from Lemma~\ref{lemma_app:outer_ellipsoid_supp} giving
\begin{equation}
\label{eq:onestep_betastar_supp}
    \beta^\star
    =
    \sqrt{\frac{\mathrm{tr}\left(\cloopmat_{n|t}\predvar_{n|t}\cloopmat_{n|t}^\top\right)}
    {\mathrm{tr}\left(\ellscale\,\mlvarmod(\predmean_{n|t},\predact_{n|t})\right)}}.
\end{equation}
Applied once more to $\prepredvar_{n+1|t}$ and the error ball $\linerr_{n|t}$ of shape matrix
$(e^{\mathrm{max}}_{n|t})^2 I$, it gives
\begin{equation}
\label{eq:onestep_final_supp}
    \predvar_{n+1|t}
    =
    (1+\tau^{-1})\,\prepredvar_{n+1|t}
    +(1+\tau)\,(e^{\mathrm{max}}_{n|t})^2 I,
\end{equation}
valid for any $\tau \in \mathbb{R}$ and $\tau>0$, with the minimizing choice
\begin{equation}
\label{eq:onestep_taustar_supp}
    \tau^\star
    =
    \sqrt{\frac{\mathrm{tr}\big(\prepredvar_{n+1|t}\big)}
    {\dimS\,(e^{\mathrm{max}}_{n|t})^2}}.
\end{equation}
Substituting $\prepredvar_{n+1|t}$ from~\eqref{eq:onestep_Qbeta_supp} into~\eqref{eq:onestep_final_supp}
recovers the propagated shape matrix~\eqref{eq:var_pred}. Note that if $e^{\mathrm{max}}_{\predid|\globid} = 0$, e.g., in the linear case, the error ball reduces to $\{0\}$ and, by the convention of Lemma~\ref{lemma_app:outer_ellipsoid_supp}, the second Minkowski sum in \eqref{eq:onestep_mink_sum_supp} leaves $\prepredvar_{\predid+1|\globid}$ unchanged. This holds similarly for cases where $\predvar_{\predid|t}=0$, e.g., in the initial step for $\predid=0$.
The sets in~\eqref{eq:onestep_mink_sum_supp}, and hence the outer ellipsoid~\eqref{eq:onestep_final_supp},
are centered at the origin, as they collect the possible deviations $s^+-\predmean_{n+1|t}$ rather than states.
Shifting this origin-centered ellipsoid by the nominal successor $\predmean_{n+1|t}$ yields the tube~\eqref{eq:onestep_tube_np1_supp},
such that $s^+\in\tube_{n+1|t}$.
\end{proof}
\end{lemma}

\begin{theorem}[Overapproximation of robustly reachable sets; extended version of Theorem~\ref{theo:app_overapproximation_tube}] \leavevmode\\
\label{theo:app_overapproximation_tube_supp}
\begin{proof}[Proof of Theorem~\ref{theo:overapproximation_tube}]
Fix $\globid$ and an input sequence $\predact_{0|t},\dots,\predact_{N-1|t}$.
Define the robust reachable sets $\prs_{n|t}$ as in Definition~\ref{def:robust_reachable_set}.

Support containment: Since $\rvalea_k \in \aleaset$ for all $k$, any realization of $\rvstate_{t+n}$ corresponds to a disturbance sequence
$(w_t,\dots,w_{t+n-1})\in\aleaset^n$. Hence, \begin{equation}
\label{eq:support_in_R_supp}
    \mathrm{support}\!\left(
        \rvstate_{t+n}\mid \rvstate_t=\state_t,\predact_{0:n-1|t}
    \right)
    \subseteq
    \prs_{n|t}.
\end{equation}
Thus, it suffices to show $\prs_{n|t}\subseteq \tube_{n|t}$.

Induction on $n$:
For $n=0$, $\prs_{0|t}=\{\state_t\}=\{\predmean_{0|t}\}\subseteq \tube_{0|t}$, and by convention from Lemma~\ref{lemma_app:outer_ellipsoid_supp} $\predvar_{0|t}=0 \Rightarrow \tube_{0|t}=\{\predmean_{0|t}\}$.
Take any $s\in\prs_{n|t}$ and any $w\in\aleaset$. By the definition of $\prs_{n+1|t}$, it holds that
$s^+ := \meanenv(s,v) + \cholenv(s,v)w \in \prs_{n+1|t}$,
where $v=\predact_{n|t}+\contgain_{n|t}(s-\predmean_{n|t})$.
Since $s\in\prs_{n|t}\subseteq \tube_{n|t}$, Lemma~\ref{lemma_app:onestep_ellipsoid_supp} implies $s^+\in\tube_{n+1|t}$.
This shows $\prs_{n+1|t}\subseteq \tube_{n+1|t}$. By induction, $\prs_{n|t}\subseteq \tube_{n|t}$ for all $n\in\{0,\dots,N\}$.

Combining the induction with \eqref{eq:support_in_R_supp} yields 
$\mathrm{support}(
        \rvstate_{t+n}\mid \rvstate_t=\state_t,\predact_{0:n-1|t})
    \subseteq
    \prs_{n|t}
    \subseteq
    \tube_{n|t}$.
Finally, the equality $\tube_{n|t}=\predmean_{n|t}\oplus \predchol_{n|t}^\top \unitball_{\statesp}$ follows from
$\predchol_{n|t}^\top \predchol_{n|t}= \predvar_{n|t}$ and Lemma~\ref{lemma_app:ellipsoid_to_minkowski_supp}.
\end{proof}
\end{theorem}

\begin{lemma}[Constraint tightening for model uncertainty; extended version of Lemma~\ref{lemma_app:uncertainty_const_tightening}]
    \label{lemma_app:uncertainty_const_tightening_supp}   
    Given a certain set 
    \begin{equation}
        \accset := \left\{ (\state, \act) \in \statesp \times \actsp \;\Big|\; \frac{1}{\dimS}\, \mathrm{tr}\!\left(\kalgain(\state, \act)\right) \geq \xi \right\},
    \end{equation}
    and the Lipschitz constant of the model uncertainty
    \begin{equation}
        \lipschitzmod := \sup_{(\state_1, \act_1) \neq (\state_2, \act_2)} \frac{1}{\dimS} \frac{|\mathrm{tr}(\kalgain(\state_1, \act_1))- \mathrm{tr}(\kalgain(\state_2, \act_2))|}{\| (\state_1, \act_1) - (\state_2, \act_2)\|_2},
    \end{equation}
    let $\mathcal{P} := \left\{ \state \in \statesp \;\Big|\; (\state - \predmean)^\top (\predvar)^{-1} (\state - \predmean) \leq 1 \right\}$. Then for $\state \in \mathcal{P}$, $v = \predact + \contgain(\state - \predmean)$, and $\unitball_{\statesp \times \actsp} = \{ \unitvec \in \mathbb{R}^{\dimS + n_{\actsp}} \mid \| \unitvec \|_2 \leq 1 \}$,
    the pair $(\state, v) \in \accset$, provided that
    \begin{equation}
        (\predmean,\, \predact) \oplus \unitball_{\statesp \times \actsp} \, \lipschitzmod \sqrt{ \left(\lambda_{\mathrm{max}}\left(  \predvar \right)  +\lambda_{\mathrm{max}}\left( \contgain  \predvar \contgain^\top \right)\right) }  \subseteq \accset.
    \end{equation}
    
    \begin{proof}
    We can upper bound the increase in uncertainty, indicated by the reduction of the normalized trace of the Kalman gain matrix, as 
    \begin{equation}
    \begin{aligned}
        &\frac{1}{\dimS} \mathrm{tr}(\kalgain(\predmean ,\predact)) - \frac{1}{\dimS} \mathrm{tr}(\kalgain(\state ,v))
         \leq \lipschitzmod \| (\predmean ,\predact) - (\state ,v) \|_2\\
        & = \lipschitzmod \sqrt{\|\state - \predmean \|_2^2 + \|v - \predact \|_2^2} = \lipschitzmod \sqrt{\|\state - \predmean \|_2^2 + \|\predact + \contgain (\state - \predmean )- \predact \|_2^2}\\
        & \leq \lipschitzmod \sqrt{ \left(\lambda_{\mathrm{max}}\left(  \predvar \right)  +\lambda_{\mathrm{max}}\left( \contgain  \predvar \contgain^\top \right)\right) } .
    \end{aligned}
    \end{equation}
    Consequently, we can lower bound $\frac{1}{\dimS} \mathrm{tr}(\kalgain(\state ,v))$ as
    \begin{equation}
        \frac{1}{\dimS} \mathrm{tr}(\kalgain(\state ,v)) \geq \frac{1}{\dimS} \mathrm{tr}(\kalgain(\predmean ,\predact)) - \lipschitzmod \sqrt{ \left(\lambda_{\mathrm{max}}\left(  \predvar \right)  +\lambda_{\mathrm{max}}\left( \contgain  \predvar \contgain^\top \right)\right) } .
    \end{equation}
    The Minkowski sum $(\predmean, \predact) \oplus \unitball_{\statesp \times \actsp}\, \lipschitzmod \sqrt{ \left(\lambda_{\mathrm{max}}\left(  \predvar \right)  +\lambda_{\mathrm{max}}\left( \contgain  \predvar \contgain^\top \right)\right) }  \subseteq \accset$
    ensures that for all $\unitvec \in \unitball_{\statesp \times \actsp}$,
    \begin{equation}
        \frac{1}{\dimS}\mathrm{tr}\!\left(\kalgain\!\left(\predmean + \lipschitzmod \sqrt{\tilde{P}} \,\unitvec_{\state},\, \predact + \lipschitzmod \sqrt{\tilde{P}} \,\unitvec_{\act}\right)\right)
        \geq \xi,
    \end{equation}
    with $\tilde{P}=\left(\lambda_{\mathrm{max}}\left(  \predvar \right)  +\lambda_{\mathrm{max}}\left( \contgain  \predvar \contgain^\top \right)\right).$
    Combining this with the bound above, we conclude
    \begin{equation}
        \xi \overset{!}{\leq} \frac{1}{\dimS} \mathrm{tr}(\kalgain(\predmean ,\predact)) - \lipschitzmod \sqrt{ \left(\lambda_{\mathrm{max}}\left(  \predvar \right)  +\lambda_{\mathrm{max}}\left( \contgain  \predvar \contgain^\top \right)\right) } \leq \frac{1}{\dimS} \mathrm{tr}(\kalgain(\state ,v)),
    \end{equation}
    which establishes $(\state, v) \in \accset$.
    \end{proof}
\end{lemma}

\section{Extended Experimental Setup}
\label{supp:extended_setup}

\subsection{Implementation of Terminal Set Expansion}
\label{supp:termset_expansion}
As highlighted in Section~\ref{subsec:term_set}, we approximate \eqref{eq:termset_expansion} via a sampling-based scheme inspired by \citet{rosolia2017learning}. To this end, for our practical implementation of \algo, we store initial points $\predmean_{0|\globid}^*$ at which \eqref{eq:upsi_simple} is feasible, and construct a terminal set $\termset^\epiid$ based on the stored solutions. However, not all solutions to \eqref{eq:upsi_simple} are eligible for the terminal set expansion. Because we use soft constraints and truncated episodes, we must verify that solution points $\predmean_{0|\globid}^*$ actually yield accurate predictions according to \eqref{eq:upsi_simple}. By introducing a parameter \verb|slack threshold|, we set a limit on the maximum values of slack variables for $\predmean_{0|\globid}^*$ to be stored, where low slack variables indicate that state, action, and certainty constraints were satisfied and the terminal set was reached. In addition, using a \verb|proximity threshold|, we remove all solutions close to a truncation from the estimation.

\begin{figure}
    \centering
    \includegraphics[width=0.6\linewidth]{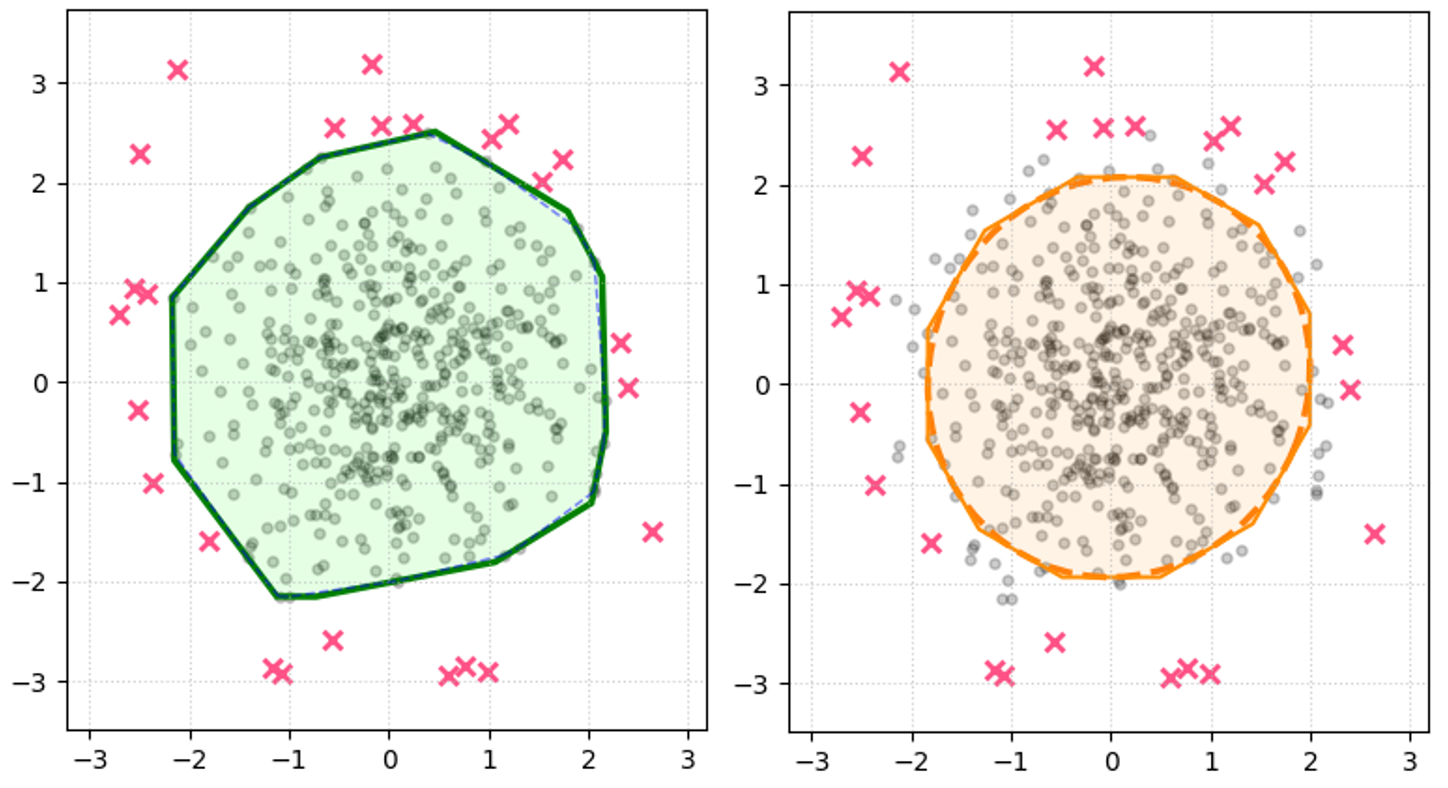}
    \caption{Sampling-based terminal set expansion. The methods \texttt{polytope} (left, green) and \texttt{ellipsoid} (right, orange) are applied to data from a normal distribution. Gray dots represent existing solutions. Crosses indicate data points removed using k-Nearest-Neighbors. Dashed lines show the exact contour, while solid lines indicate the reduced facets employed as $\termset^\epiid$ in \eqref{eq:upsi_simple_term_const}.}
    \label{fig:termset_methods}
\end{figure}

In practice, the expansion of $\termset^\epiid$ must remain highly conservative. Thus, we use the following heuristic: via scikit-learn's k-Nearest-Neighbors ($k=10, p=0.95$), we remove outliers from $\predmean_{0|\globid}^*$, followed by set estimation based on the reduced number of solution points. 
Two methods are available for set estimation: \verb|polytope| uses the convex hull of all data, and \verb|ellipsoid| computes the variance of the data, which enables the scaling of a terminal ellipsoid (probability level $0.7$). Finally, for the integration of the resulting sets as $\termset^\epiid$ in \eqref{eq:upsi_simple}, we convert their outer contour into linear facets, for which \eqref{eq:upsi_simple_term_const} can be verified using numerical optimization. A visualization of both methods is shown in Figure~\ref{fig:termset_methods}. 
We note that the methods presented above are approximate, and their conservatism needs to be manually verified in practice.

\subsection{Reachable Set Computation in Cartpole}
\label{supp:reachable_sets}

For the empirical validation of Theorem \ref{theo:overapproximation_tube} in Figure~\ref{fig:mpc_sequence}, we train a PE model with $E=5$ members, each member consisting of two layers with $64$ neurons and \verb|tanh| activation. We train this model on sufficient data (30,000 environment rollouts) to obtain a highly accurate representation. Data is sampled from a Cartpole environment that follows the general dynamics in~\eqref{eq:env_dyn}. For the process noise, we sample from a multivariate Gaussian, truncated by setting $\ellscale = \mathcal{F}^{-1}_{\chi^2(\dimS)}(0.99)$. Here, $\cholenv(\rvstate_\globid, \rvact_\globid)=(C_0+C_1|\theta|)I$ induces heteroscedasticity, with parameters $C_0=3\times 10^{-3}$ and $C_1=5\times 10^{-4}$. Based on $\cholenv(\rvstate_\globid, \rvact_\globid)$, we have $\lipschitzvarchol =C_1 \sqrt{\dimS} $. We use a starting state $\predmean_{0|\globid} = \state_\globid = [1.0, -1.0,0.0,-0.25]$ and define a nominal trajectory via zeroed actions for simplicity.
Similar to \eqref{eq:upsi_init}, we set $\predvar_{0|\globid} = 0$. The following ellipsoids are determined using \eqref{eq:mean_pred} and \eqref{eq:var_pred}. An ancillary feedback controller $\contgain$ is computed at $\state_\globid$ via a minimal norm stabilizing gain selected from LQR. Finally, we compare the resulting tubes with the ground truth of 100,000 trajectories sampled from the environment under ancillary feedback. For more details, we refer the reader to the implementation in \verb|reachable_sets.py|.

\subsection{UPSi Implementation Details}
\label{supp:upsi_implementation}

As stated in Section~\ref{sec:experiments}, for the experiments in Figure~\ref{fig:results}, we integrate \algo into MBPO by propagating the simplified naive tube $\tilde{\prepredvar}_{n+1|\globid}$ followed by scaling with $\ellscale$, instead of using shape $\prepredvar_{n+1|\globid}$. Propagating the simplified naive tube is motivated by achieving a simple implementation that does not require extensive systems analysis, while also reducing conservatism, which could inhibit exploration. However, these choices risk decreased robustness.
In practice, we observe that the simplified naive tube empirically captures the ground truth well, given a conservative choice of $\epsilon$ (see Figure~\ref{fig:mpc_sequence}). Furthermore, propagating the naive tube reduces the numerical effort involved in computing \eqref{eq:var_pred}. Also motivated by the simplicity of implementation, we do not enforce the use of a series of stabilizing $\contgain_{\predid|\globid}$ for ancillary feedback. However, it is advisable to consider ancillary feedback if long prediction horizons $\predhorizon$ are required, which we do not observe in our experiments. Lastly, we only require the nominal state and action to remain in $\accset ^j$ (see \eqref{eq:upsi_simple_cert_const}), based on the finding that a conservatively chosen value of $\xi$ can compensate for this relaxation. 

To mitigate potential infeasibility arising from our practical assumptions, we use the backup scheme from~\citet{wabersich_linear_2018}, whereby previous PSF solutions are used if \algo cannot find a solution at time step $t$. Note that due to the terminal set expansion, the local control law $\initpol$ may not be valid everywhere in $\termset^\epiid$; therefore, we do not employ $\initpol$ in the last step of the backup algorithm and instead resort to the policy input $\act_\globid$. Furthermore, we use soft constraints with severely penalized slack variables. Below, we present the practical \algo problem~\eqref{eq:upsi_simple}, resulting from the above assumptions, along with the backup sequence detailed in Algorithm~\ref{alg:backup_controls}:
\begin{subequations}\label{eq:upsi_simple}
\begin{align}
    \min_{\predact_{0|\globid}, \dots, \predact_{ \predhorizon-1|\globid}} & \|\act_\globid - \predact_{0|\globid} \|_2 \label{eq:upsi_simple_cost}\\ %
    \text{s.t. } & \predmean_{0|\globid} = \state_\globid, \quad \tilde{Q}_{0|\globid} = 0 \label{eq:upsi_simple_init}\\
    &\predmean_{\predid+1 \mid \globid} = \mlmeanmod(\predmean_{\predid \mid \globid}, \predact_{\predid \mid \globid}) \text{ for all }  \predid \in \{0, \dots, \predhorizon -1 \}\label{eq:upsi_simple_mean_pred}\\
    & \tilde{Q}_{\predid+1 \mid \globid} = \cloopmat_{\predid \mid \globid} \tilde{Q}_{\predid \mid \globid}\cloopmat_{\predid \mid \globid}^\top + \mlvarmod(\predmean_{\predid \mid \globid}, \predact_{\predid \mid \globid})  \text{ for all }  \predid \in \{0, \dots, \predhorizon -1 \} \\
    & \predmean_{\predid \mid \globid} \in \stateset \ominus \tilde\predchol_{\predid|\globid}^\top \unitball_{\statesp} \text{ for all }  \predid \in \{0, \dots, \predhorizon -1 \}\label{eq:upsi_simple_state_const}\\
    & \predact_{\predid \mid \globid} \in \actset \ominus \contgain_{\predid \mid \globid} \tilde\predchol_{\predid|\globid}^\top \unitball_{\statesp}   \text{ for all }  \predid \in \{0, \dots, \predhorizon -1 \} \label{eq:upsi_simple_act_const}\\
    & \predmean_{\predhorizon \mid \globid} \in \termset^\epiid \ominus \tilde\predchol_{\predhorizon|\globid}^\top \unitball_{\statesp}\label{eq:upsi_simple_term_const}\\
    & (\predmean_{\predid \mid \globid}, \predact_{\predid \mid \globid}) \in \accset^\epiid \text{ for all } \predid \in \{0, \dots, \predhorizon - 1 \}. \label{eq:upsi_simple_cert_const}
\end{align}
\end{subequations}
where $\tilde M_{n|t}$ denotes the Cholesky factor $\tilde M_{n|t}^\top \tilde M_{n|t} = \ellscale \tilde{Q}_{n|t}$ of the naive tube scaled by $\ellscale$, such that $z_{n|t} \oplus \tilde M_{n|t}^\top \unitball$ is its $\ellscale$-level set. Despite the simplifications, we achieve substantially improved safety during exploration compared to baselines (see Figure~\ref{fig:results}). 

\begin{algorithm}
\caption{Backup scheme based on \citet{wabersich_linear_2018}.}
\label{alg:backup_controls}
\begin{algorithmic}[1]
\State $n_{\text{inf}} := N - 1$
\For{$t = 0, 1, 2, \dots$}
    \If{\eqref{eq:upsi_simple} feasible}
        \State $n_{\text{inf}} := 0$
        \State Apply $\act^{\mathrm{F}}_\globid = \predact^*_{0|\globid}$ to \eqref{eq:env_dyn}
        \State \text{Store solution of \eqref{eq:upsi_simple}} 
    \ElsIf{$n_{\text{inf}} < N - 1$}
        \State $n_{\text{inf}} := n_{\text{inf}} + 1$
        \State Apply $\act^{\mathrm{F}}_\globid = u_{n_{\text{inf}}|t-n_{\text{inf}}} + K_{n_{\text{inf}}|t-n_{\text{inf}}}(s_t - z_{n_{\text{inf}}|t-n_{\text{inf}}})$ to \eqref{eq:env_dyn}
    \Else
        \State Apply $\act^{\mathrm{F}}_\globid=\act_\globid$ to \eqref{eq:env_dyn}
    \EndIf
\EndFor
\end{algorithmic}
\end{algorithm}

Empirical tuning of our practical implementation of \algo involves defining suitable values for $\ellscale$ and $\xi$. Values for $\ellscale$ can be found by assuming a sufficiently high probability level for the naive tube and computing $\ellscale$ accordingly, e.g., $\ellscale = \mathcal{F}^{-1}_{\chi^2(\dimS)}(0.99)$. For $\xi$, we advise choosing values close to $1.0$. Furthermore, the chosen prediction horizon $N$ must allow flexibility of the RL agent while balancing computing time and tube growth. The parameters used for \algo during our experiments are displayed in Table~\ref{tab:filter_params}. We employ the parameters reported as optimal for XMPSC in \citet{gronauer_reinforcement_2024}, while the parameters for \algo were determined via grid search.

\begin{table}
    \centering
    \caption{Filtering parameters for \algo experiments. When noting $\ellscale$, we also note the equivalent probability level on the propagated Gaussian in brackets.}
    \begin{tabular}{|c|c|c|c|}\hline
         &  Pendulum&  Cartpole& Drone\\\hline
         Certainty Threshold $\xi$&  0.9&  0.7& 0.9\\\hline
         Noise Bound $\ellscale$ (According Probability Level) & 2.41 (70\%)&  9.59 (95\%)& 10.2 (40\%)\\\hline
         Horizon $\predhorizon$&  10&  15& 9\\\hline
 Terminal Set Method& \verb|polytope|& \verb|polytope|&\verb|ellipsoid|\\\hline
 \verb|proximity threshold|& 25& 50&50\\\hline
 \verb|slack threshold|& 0.1& 0.1&0.1\\ \hline
    \end{tabular}
    \label{tab:filter_params}
\end{table}

\subsection{Experimental Setup}
\label{supp:experimental_setup}
\begin{figure}[htbp]
    \centering
    \begin{subfigure}[b]{0.29\textwidth}
        \centering
        \includegraphics[width=\textwidth]{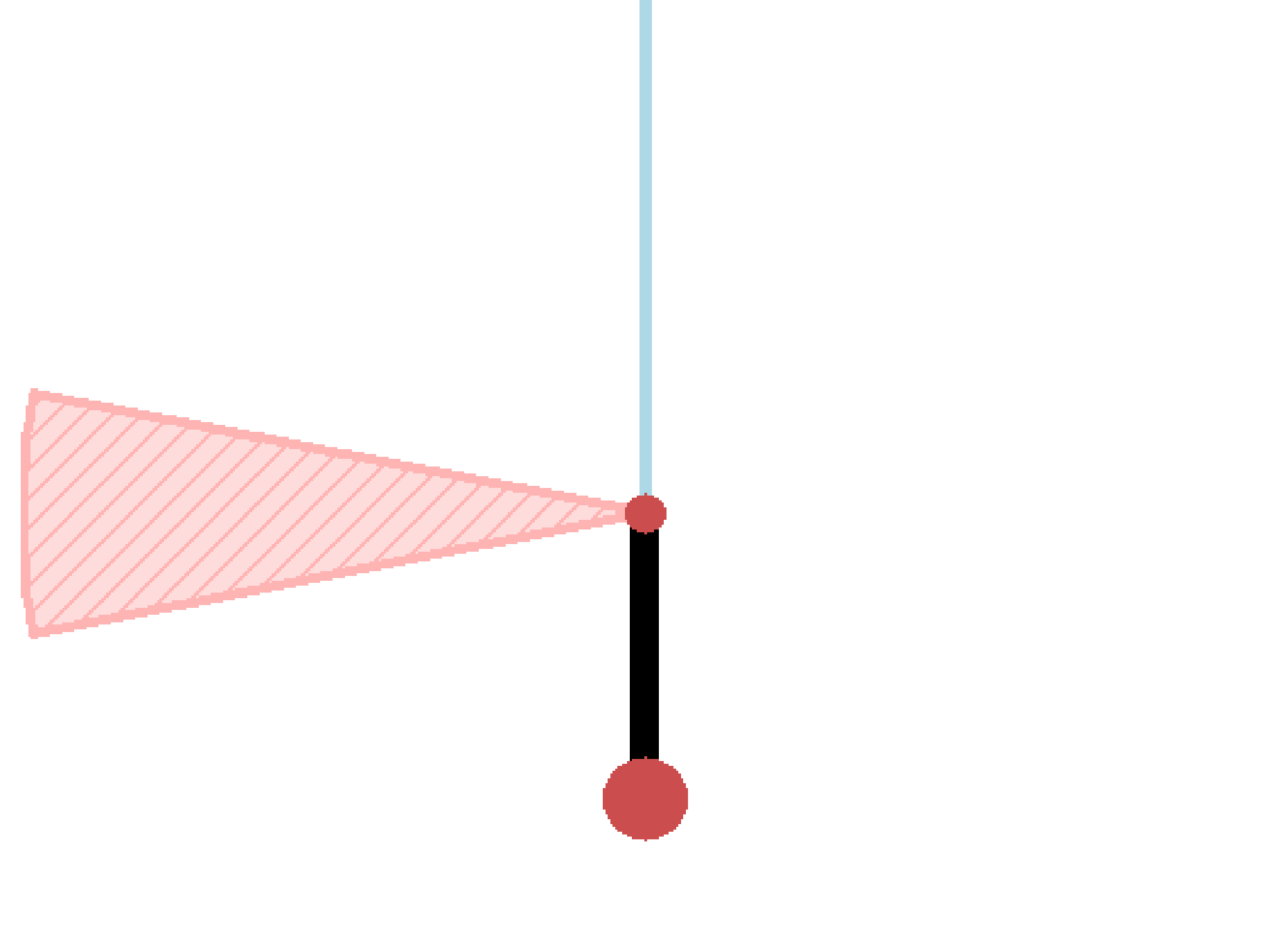}
        \caption{Pendulum}
        \label{fig:sub1}
    \end{subfigure}
    \hfill
    \begin{subfigure}[b]{0.34\textwidth}
        \centering
        \includegraphics[width=\textwidth]{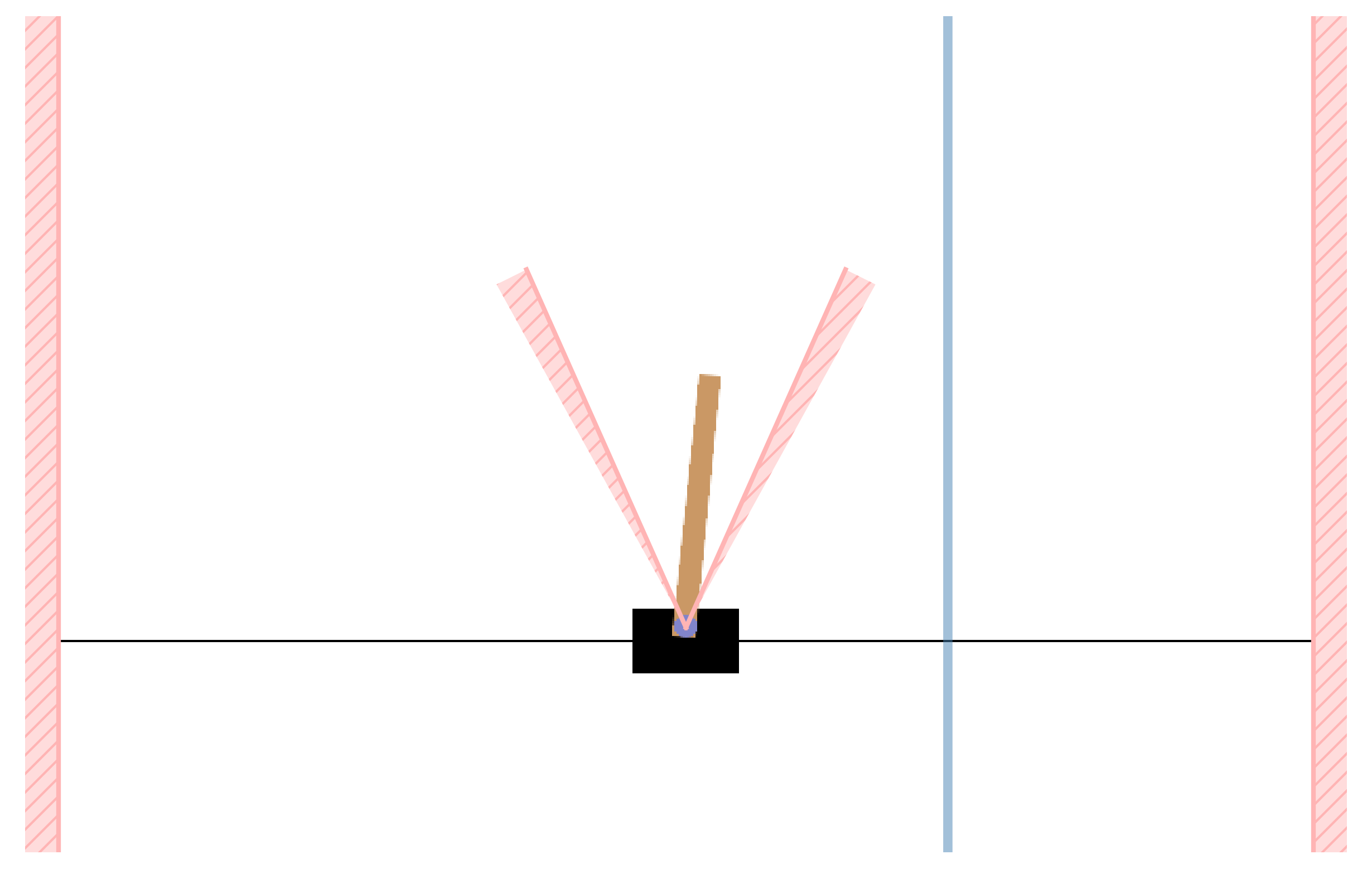}
        \caption{Cartpole}
        \label{fig:sub2}
    \end{subfigure}
    \hfill
    \begin{subfigure}[b]{0.34\textwidth}
        \centering
        \includegraphics[width=\textwidth]{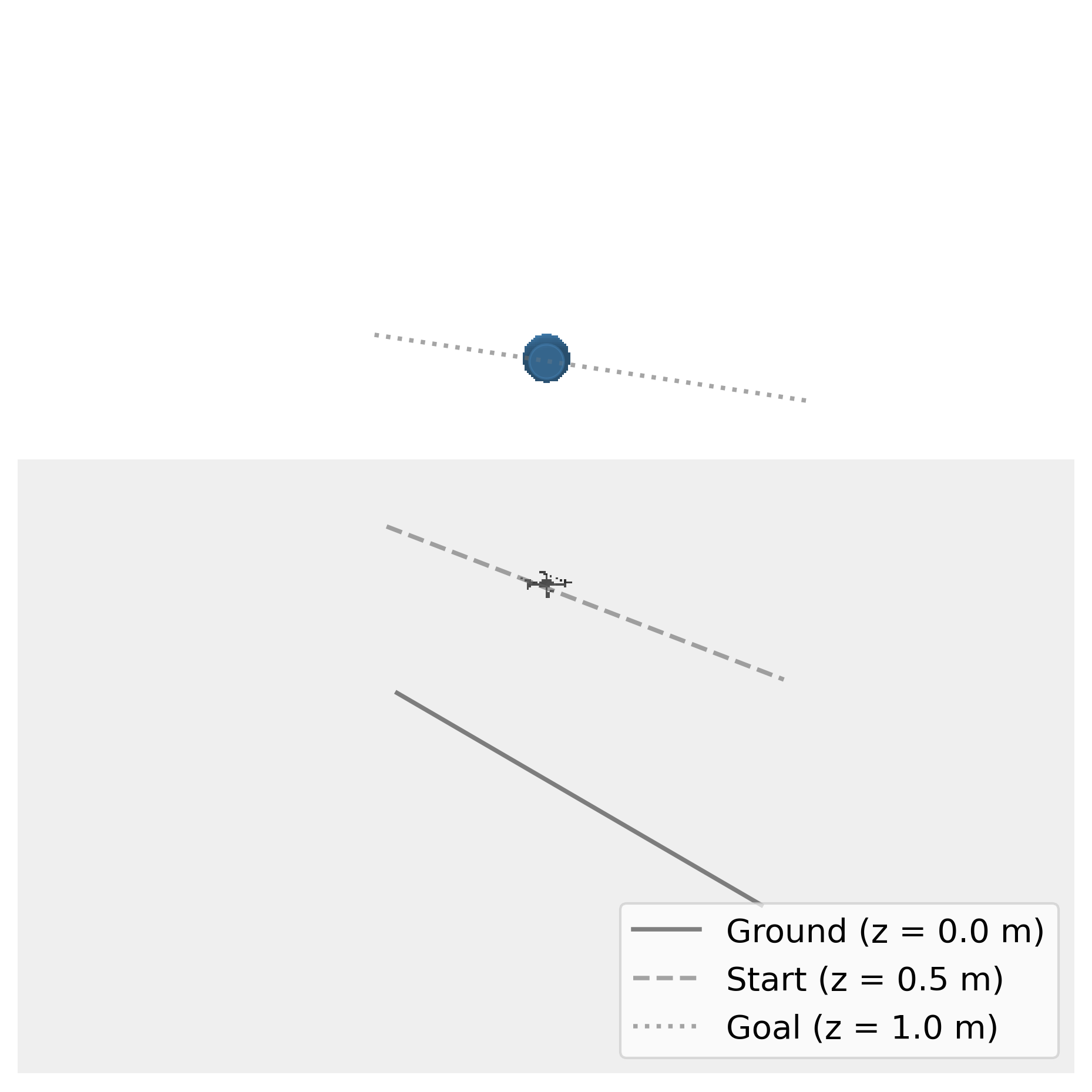}
        \caption{Drone}
        \label{fig:sub3}
    \end{subfigure}
    
    \caption{Simulated environments for experiments. Areas highlighted in red indicate constrained parts of the state space. The goal state of each environment is indicated in blue.}
    \label{fig:environments}
\end{figure}

We show conceptual renders of our simulated environments in Figure~\ref{fig:environments}.
The goal in our Pendulum is to swing up from the bottom ($\theta=\pi$) toward the upright position ($\theta=2\pi$), while we restrict the angle $\theta$ to avoid an area in the left semicircle. The constraints in Pendulum are $\theta \in \left[\tfrac{\pi}{2} + \tfrac{\pi}{16},\; \tfrac{5\pi}{2} - \tfrac{\pi}{16}\right]$, $\dot{\theta} \in \left[-8,\; 8\right]$ rad/s, and $a \in \left[-2,\; 2\right]$ Nm. For increased exploration compared to standard Cartpole, we locate the goal position ($x=1.0$m) to the right of the initialization point ($x=0.0$m) on the $x$-axis. The constraints in Cartpole are $\theta \in \left[-\frac{24 \cdot 2\pi}{360}, \frac{24 \cdot 2\pi}{360}\right]$, $x\in [-2.4, 2.4]$, and $a \in [-2, 2]$ N. In Drone, the objective is to fly to a target height ($z=1$ m) from half of the target height ($z=0.5$ m). The constraints concern the roll angle $\phi \in \left[-\tfrac{\pi}{18},\, \tfrac{\pi}{18}\right]$, as well as the pitch angle $\quad \vartheta \in \left[-\tfrac{\pi}{18},\, \tfrac{\pi}{18}\right]$, with actions $a \in [-1, 1]^4 $. Cartpole and Drone include an initial control law $\initpol$ based on a noisy LQR controller, which is used for safely gathering data at the origin and estimating the initial terminal set. In contrast, Pendulum uses a purely random controller that is incapable of leaving the bottom of the pendulum. All environments run for an episode length of $200$ steps until truncation and incorporate bounded process noise as defined in \eqref{eq:env_dyn}.

We use Gym for modeling our benchmark environments. Neural networks for the actors, critics, and PE models are trained using PyTorch. Optimization problems are solved using CasADi, with UPSi employing sqpmethod with qpOASES. In each experiment, $8$ random seeds are used per algorithm. No performance or safety outliers are discarded\footnote{No full run is excluded from the results; the removal of solution points in Supp.~\ref{supp:termset_expansion} concerns the terminal set only.}. To parallelize training for speedup, MPI is used with $8$ to $12$ processes and one solver instance per process. Each MPI process is instantiated with a seed based on a deterministic permutation of the base seed. Across MPI processes, agents are synchronized while gathering data from environments with varying seeds. For more implementation details, we refer to our code implementation\footnote{Code can be accessed online: \hyperlink{https://github.com/Data-Science-in-Mechanical-Engineering/upsi}{https://github.com/Data-Science-in-Mechanical-Engineering/upsi}.}, specifically \verb|mpsc.py| and \verb|mbpo.py|. The most relevant training hyperparameters for all MBPO agents are displayed in Table~\ref{tab:hyperparams}. Note that our implementation of the MBPO algorithm is similar to that of~\citet{gronauer_reinforcement_2024}, as are the evaluation environments.

\begin{table}
    \centering
    \caption{Relevant hyperparameters for MBPO as used in Figure~\ref{fig:experiments}.
    }
    \begin{tabular}{|c|c|c|c|}\hline
         &  Pendulum&  Cartpole& Drone\\\hline
 Epochs& 100& 200&400\\\hline
 Epoch Steps& 256& 256&256\\\hline
 Model Rollouts per Step& 400& 200&200\\\hline
 Policy Updates per Step& 20& 20&20\\\hline 
         Initial Data Size&  8,192&  8,192& 65,536\\ \hline
         Real Ratio&  0.1&  0.1& 0.1\\ \hline
         Ensemble Size $E$ &  5&  5& 5\\ \hline
 Ensemble Members& (16, 16, \verb|tanh|)& (16, 16, \verb|tanh|)&(16, 16, \verb|tanh|)\\\hline
 Actor Networks& (100, 100, \verb|relu|)& (200, 200, \verb|relu|)&(100, 100, \verb|relu|)\\\hline
 Critic Networks& (100, 100, \verb|relu|)& (200, 200, \verb|relu|)&(100, 100, \verb|relu|)\\ \hline
 MPI Processes& 8& 8&12\\\hline
    \end{tabular}
    \label{tab:hyperparams}
\end{table}

\subsection{Additional Results}
\label{supp:additional_results}
\begin{table}[ht]
    \centering
        \caption{Key metrics of safe exploration. First, the average is computed across MPI processes for each seed, followed by computing the maximum and minimum across seeds. The infeasibility rate depicts the total number of infeasibilities divided by the total number of time steps $\globid$.}
    \begin{tabular}{|l|c|l|c|l|c|l|}\hline
          &\multicolumn{2}{|c|}{Pendulum}&\multicolumn{2}{|c|}{Cartpole}&  \multicolumn{2}{|c|}{Drone}\\\hline
          &UPSi&   XMPSC& UPSi&XMPSC&   UPSi&XMPSC\\\hline
          
          Total Infeasibilities (Train)&11&   142& 382&3,660&   430&664\\\hline
 Infeasibility Rate (Train)& 0.04\%& 0.56\%& 0.75\%& 7.15\%& 0.42\%&0.65\%\\\hline
          Min. Filter Rate (Eval)&10.9\%&   0.00\%& 46.4\%&0.00\%&   80.9\%&0.00\%\\\hline
 Max. Filter Rate (Eval)& 52.3\%& 49.9\%& 95.8\%& 79.9\%& 100\%&38.6\%\\\hline
 Min. Opt. Time (Train)& 0.0405 s& 0.248 s& 0.272 s& 0.405 s& 10.0 s&4.68 s\\\hline
 Max. Opt. Time (Train)& 0.100 s& 0.958 s& 1.68 s& 9.05 s& 25.8 s&37.2 s\\ \hline 
    \end{tabular}
    \label{tab:additional_results}
\end{table}

\begin{figure}[ht]
    \centering
    \includegraphics[width=0.99\linewidth]{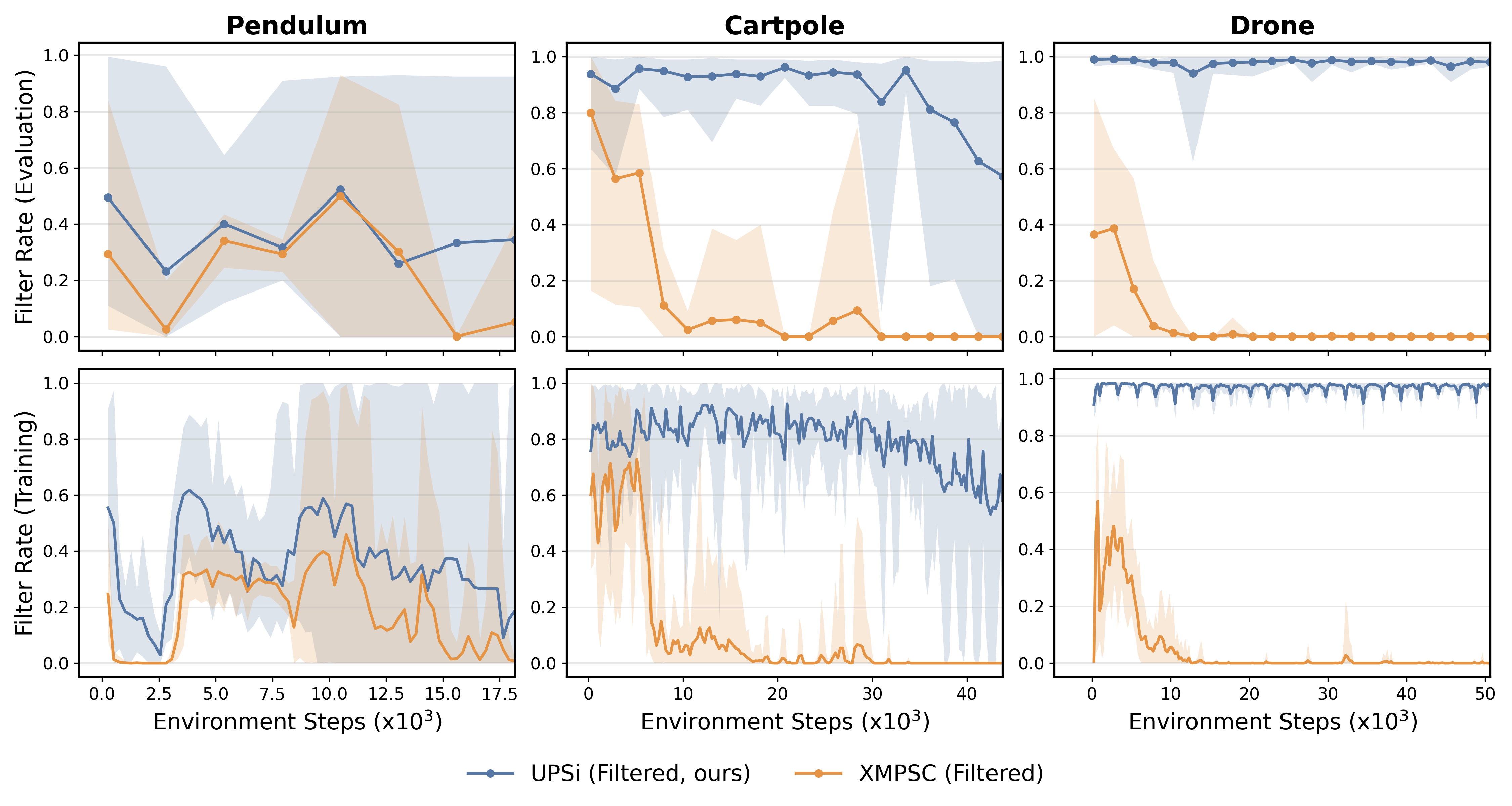}
    \caption{Filter rates over environment steps during training and evaluation; solid lines show the mean and shaded regions the min–max range across $8$ seeds. \algo generally intervenes more frequently than XMPSC. Note that corrections to the input from RL agents are minimal according to \eqref{eq:upsi_simple_cost}, i.e., even at high filter rates, the agent potentially achieves its desired behavior.}
    \label{fig:filter_rates}
\end{figure}

In addition to the results in Figure~\ref{fig:results}, Table~\ref{tab:additional_results} shows key metrics during training and evaluation. Furthermore, Figure~\ref{fig:filter_rates} shows the filter rate, which represents the ratio of filtered steps to total steps in the episode. We observe that \algo generally encounters fewer infeasibility events while filtering actions more frequently than XMPSC across all environments. This behavior provides a basis for understanding the increased safety during exploration when using \algo in Figure~\ref{fig:results}, since \algo intervenes more frequently and still finds solutions more often than XMPSC. 

\begin{figure}[ht]
    \centering
    \includegraphics[width=0.99\linewidth]{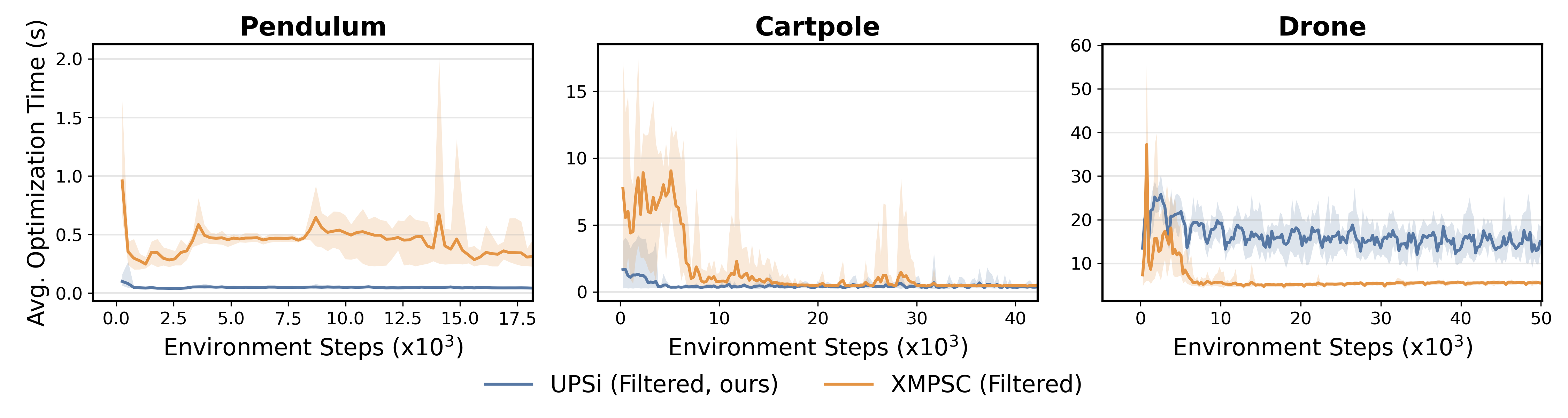}
    \caption{Optimization times over environment steps during training; solid lines show the mean and shaded regions the min–max range across $8$ seeds. }
    \label{fig:opt_times}
\end{figure}

As depicted in Table~\ref{tab:additional_results}, in low-dimensional environments such as Pendulum, the current implementation of \algo allows for a control refresh rate of up to 25 Hz, while real-time feasibility has not yet been achieved for higher-dimensional environments such as Drone. It is generally difficult to compare the numerical performance of XMPSC and \algo because they employ different solver architectures -- \algo uses SQP, while XMPSC employs Ipopt. Nevertheless, we show the optimization times over environment steps during training in Figure~\ref{fig:opt_times}.

\end{document}